\RequirePackage{fix-cm}
\documentclass[twocolumn]{svjour3}          % twocolumn

\makeatletter
\expandafter\let\csname opt@amsmath.sty\endcsname\relax% Remove options passed to amsmath
\makeatother
\smartqed  % flush right qed marks, e.g. at end of proof
\usepackage{graphicx}
\usepackage{amsmath}
\usepackage{hyperref}
\usepackage{multirow}
\usepackage{amssymb}
\usepackage{algorithmic, algorithm}
\usepackage{natbib}
\usepackage{xcolor}
\usepackage{appendix}
\graphicspath{{images/}}
\makeatletter
\newcommand\notsotiny{\@setfontsize\notsotiny{7}{8}}
\makeatother
\mathchardef\mhyphen="2D

\journalname{International Journal of Computer Vision}

\begin{document}

\title{VPR-Bench}

\subtitle{An Open-Source Visual Place Recognition Evaluation Framework with Quantifiable Viewpoint and Appearance Change.}

\author{Mubariz Zaffar         \and
        Sourav Garg \and
        Michael Milford \and
        Julian Kooij \and
        David Flynn \and
        Klaus McDonald-Maier \and
        Shoaib Ehsan 
}

\institute{Mubariz Zaffar, Klaus McDonald-Maier and Shoaib Ehsan \at
              School of Computer Science and Electronic Engineering,
        University of Essex, CO4 3SQ, United Kingdom \\
              \email{mubariz.zaffar, kdm, sehsan@essex.ac.uk}           
           \and
           Mubariz Zaffar and Julian Kooij \at
              Cognitive Robotics, TU Delft, 2628CD, Netherlands \\
              \email{m.zaffar, j.f.p.kooij@tudelft.nl}           
           \and
           Sourav Garg and Michael Milford \at
              School of Electrical Engineering and Computer Science, Queensland University of Technology, Brisbane, QLD 4000, Australia \\
              \email{s.garg, michael.milford@qut.edu.au}
           \and
           David Flynn \at
              School of Engineering and Physical Sciences, Smart Systems Group, Heriot-Watt University, Edinburgh, Currie EH14 4AS, United Kingdom \\
              \email{D.Flynn@hw.ac.uk}
}

\maketitle

\begin{figure}[t]
\begin{center}
%\fbox{\rule{0pt}{2in} \rule{0.9\linewidth}{0pt}
\includegraphics[width=1\linewidth]{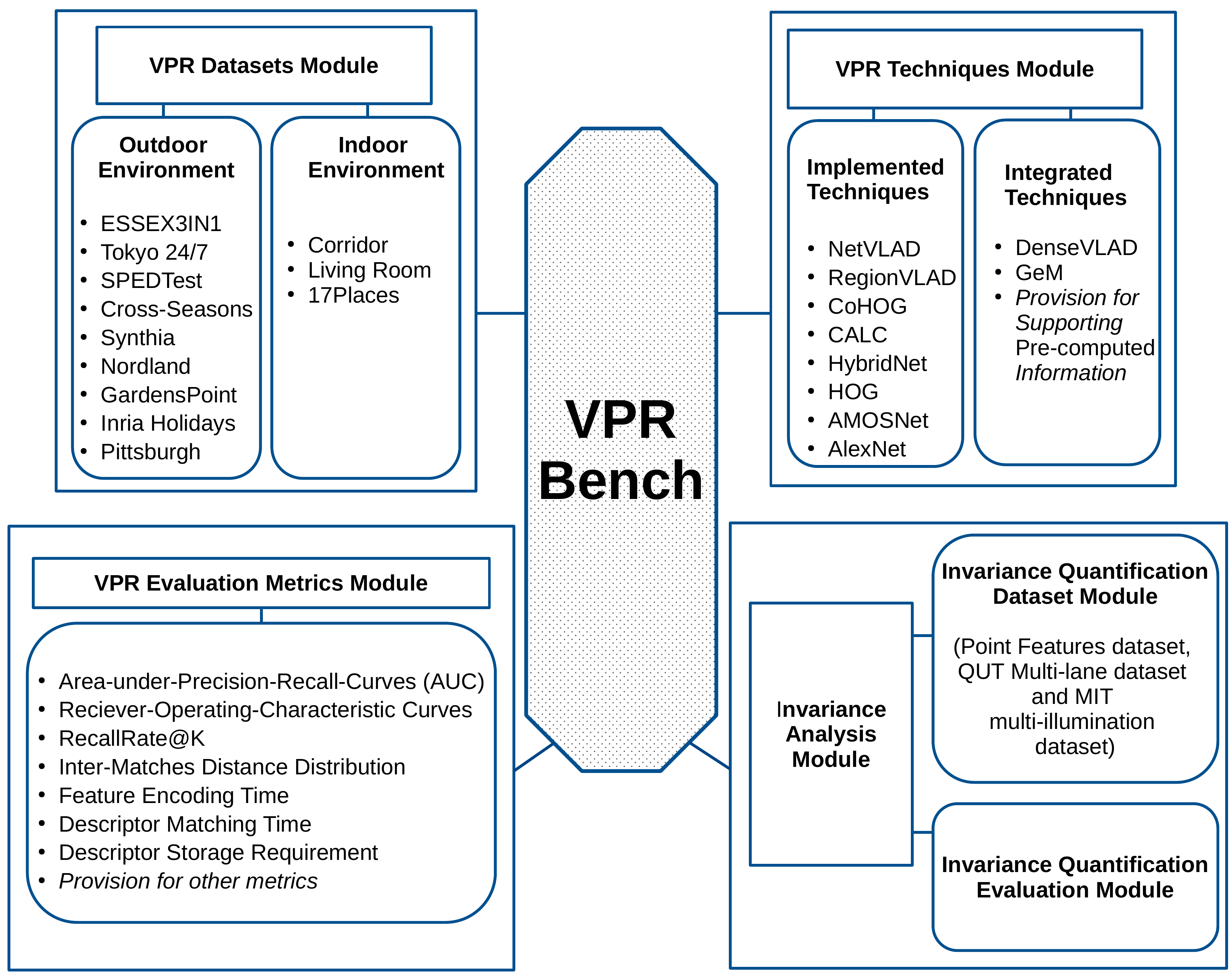}
\end{center}
\caption{A block-diagram overview of the developed VPR-Bench framework is shown here. All modules can be inter-linked within the framework and can also be independently modified for graceful updates in the future.}
\label{VPRBench_described_PDF}
\vspace{-5mm}
\end{figure}
\keywords{Visual Place Recognition \and SLAM \and Autonomous Robotics \and Robotic Vision}

\begin{abstract}
Visual Place Recognition (VPR) is the process of recognising a previously visited place using visual information, often under varying appearance conditions and viewpoint changes and with computational constraints. VPR is related to the concepts of localisation, loop closure, image retrieval and is a critical component of many autonomous navigation systems ranging from autonomous vehicles to drones and computer vision systems. While the concept of place recognition has been around for many years, VPR research has grown rapidly as a field over the past decade due to improving camera hardware and its potential for deep learning-based techniques, and has become a widely studied topic in both the computer vision and robotics communities. This growth however has led to fragmentation and a lack of standardisation in the field, especially concerning performance evaluation. Moreover, the notion of viewpoint and illumination invariance of VPR techniques has largely been assessed qualitatively and hence ambiguously in the past. In this paper, we address these gaps through a new comprehensive open-source framework for assessing the performance of VPR techniques, dubbed ``VPR-Bench''. VPR-Bench\footnote{Open-sourced at: \url{https://github.com/MubarizZaffar/VPR-Bench}} introduces two much-needed capabilities for VPR researchers: firstly, it contains a benchmark of 12 fully-integrated datasets and 10 VPR techniques, and secondly, it integrates a comprehensive variation-quantified dataset for quantifying viewpoint and illumination invariance. We apply and analyse popular evaluation metrics for VPR from both the computer vision and robotics communities, and discuss how these different metrics complement and/or replace each other, depending upon the underlying applications and system requirements. Our analysis reveals that no universal SOTA VPR technique exists, since: (a) state-of-the-art (SOTA) performance is achieved by 8 out of the 10 techniques on at least one dataset, (b) SOTA technique in one community does not necessarily yield SOTA performance in the other given the differences in datasets and metrics.
Furthermore, we identify key open challenges since: (c) all 10 techniques suffer greatly in perceptually-aliased and less-structured environments, (d) all techniques suffer from viewpoint variance where lateral change has less effect than 3D change, and (e) directional illumination change has more adverse effects on matching confidence than uniform illumination change. We also present detailed meta-analyses regarding the roles of varying ground-truths, platforms, application requirements and technique parameters. Finally, VPR-Bench provides a unified implementation to deploy these VPR techniques, metrics and datasets, and is extensible through templates.

\end{abstract}

\section{Introduction}
\label{intro}

Visual Place Recognition (VPR) is a challenging and widely investigated problem within the computer vision community (\cite{lowry2015visual}). It identifies the ability of a system to match a previously visited place using on-board computer vision prowess, with resilience to perceptual aliasing and seasonal-, illumination- and viewpoint-variations. This ability to correctly and efficiently recall previously seen places using only visual input has many important applications, such as loop-closure in SLAM (Simultaneous Localisation and Mapping) pipelines (\cite{cadena2016past}) to correct for localization drifts, image search based on visual content (\cite{tolias2016image}), location-refinement given human-machine interfaces (\cite{robertson2004image}), query-expansion (\cite{johns2011images}), improved representations (\cite{tolias2013aggregate}), vehicular navigation (\cite{fraundorfer2007topological}), asset-management using aerial imagery (\cite{odo2020towards}) and 3D-model creation (\cite{agarwal2011building}).

\begin{figure*}[t]
\begin{center}
%\fbox{\rule{0pt}{2in} \rule{0.9\linewidth}{0pt}
\includegraphics[width=1.0\linewidth]{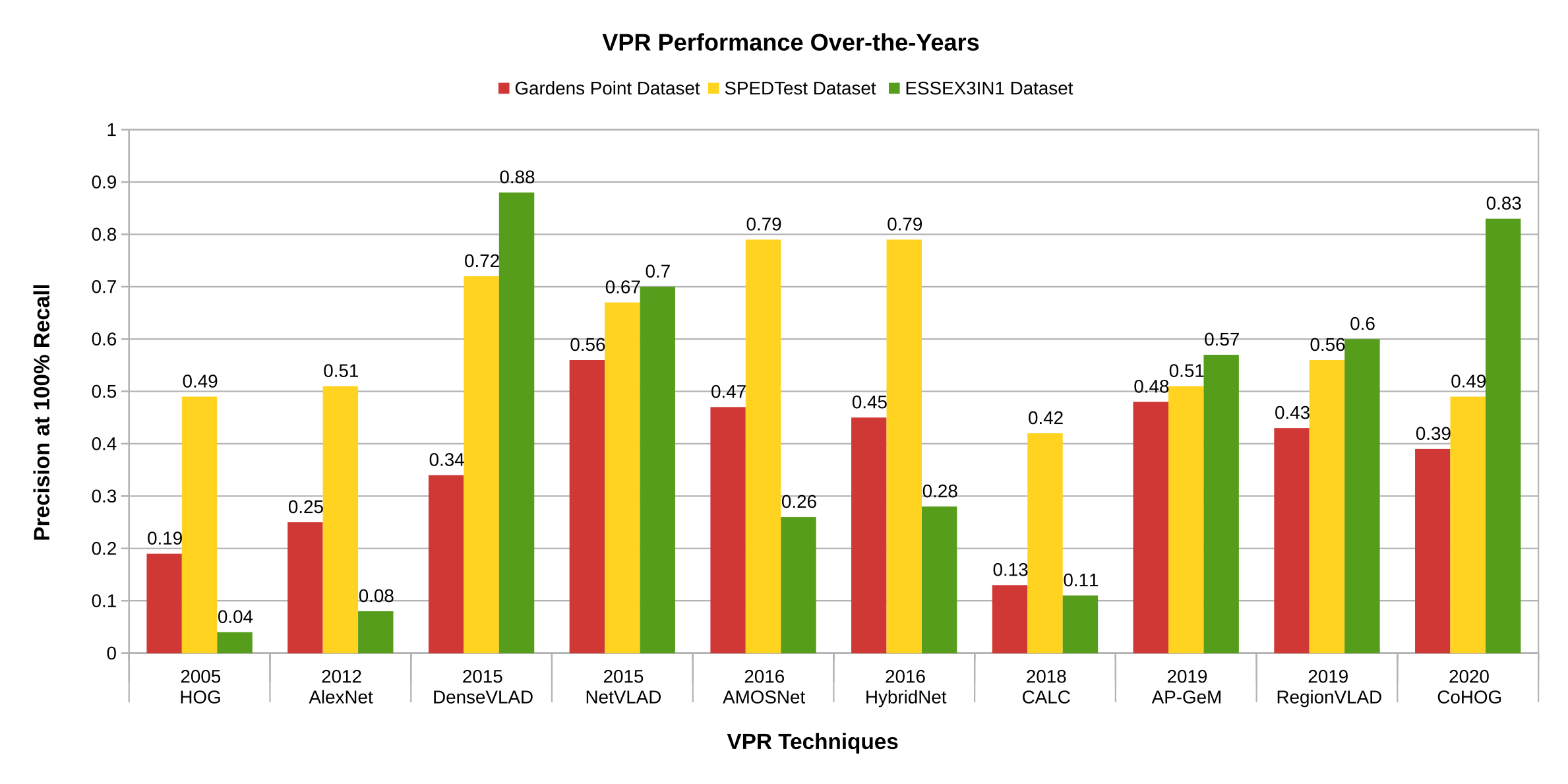}
\end{center}
\caption{Precision at 100\% Recall (equivalent to RecallRate@1) of 10 VPR techniques on Gardens Point dataset (\cite{chen2014convolutional}), SPEDTest dataset (\cite{chen2018learning}) and ESSEX3IN1 dataset (\cite{zaffar2020memorable}) is shown here in a chronological order. The trends show irregularities in between techniques and datasets, while the increase in precision is also not temporally consistent. These datasets and techniques have been discussed later in our paper. Please note that this graph is not intended to reflect the utility of these techniques, as some less-precise techniques have significantly lower computational requirements and can process more place-recognition (loop-closure) candidates.}
\label{VPR_Precision_overtheyears}
\vspace{-5mm}
\end{figure*}

Consequently, VPR researchers come from various backgrounds, as witnessed by the many workshops organised in top-tier conferences, e.g. `Long-Term Visual Localisation Workshop Series' in Computer Vision and Pattern Recognition Conference (CVPR), `Visual Place Recognition in Changing Environments Workshop Series' in IEEE International Conference on Robotics and Automation (ICRA), `Large-Scale Visual Place Recognition and Image-Based Localization Workshop' in IEEE International Conference on Computer Vision (ICCV 2019) and `Visual Localisation: Features-based vs Learning Approaches' in European Conference on Computer Vision (ECCV 2018). Thus, VPR has drawn huge interest from the computer vision and robotics research communities, leading to a large number of VPR techniques proposed over the past many years, but the communities remain separated and the state-of-the-art is not temporally consistent (see Fig. \ref{VPR_Precision_overtheyears}).

This divide is primarily due to the application requirements for both the domains: robotics researchers usually focus on having highly confident estimates predicting a revisited place to perform loop-closure, while the computer vision community prefers to retrieve as many prospective matches of a query image as possible for 3D-model creation, for example. The number of correct reference matches for the former are usually limited to a few (1-5), associated with repeated traversals under varied conditions, and thus robotics uses smaller datasets, e.g. Gardens Point dataset (\cite{arren_glover_2014_4590133}), ESSEX3IN1 (\cite{zaffar2020memorable}) dataset, Campus Loop dataset (\cite{merrill2018lightweight}) and others. For the latter, the number of correct matches (reference images) are larger ($>$ 10), corresponding to a broad collection of photos of a landmark, and thus uses substantially sized datasets, e.g. the Pittsburgh dataset (\cite{torii2013visual}), Oxford Buildings dataset (\cite{Philbin07}), Paris dataset (\cite{Philbin08}) and their revisited versions with increased 1M distractors by \cite{Radenovic_2018_CVPR}.\footnote{These remarks are only depicting the evident trends and are not absolute. Large-scale datasets (e.g. the Nordland dataset by \cite{nordlanddataset} and Oxford robot-car dataset by \cite{maddern20171}) for the robotics community, and small-scale datasets (e.g. the INRIA Holidays dataset by \cite{jegou2008hamming}) for the computer vision community do exist.}
In addition, robotics mostly focuses on high precision, usually requiring a single correct match for localisation estimates. It therefore employs evaluation metrics such as AUC-PR and F1-Score, while the computer vision community has predominantly used  Recall@N, mean-Average Precision (mAP) and/or Recall@Reduced Precision.
The divergence in datasets and metrics has limited the comparison of the techniques across the two domains to intra-domain-type evaluations, hence the state-of-the-art remains ambiguous. Therefore, one of the key contributions of our work is attempting to reduce this gap by integrating datasets, metrics and techniques from both the domains
into a novel framework called \textit{VPR-Bench}, which is carefully designed to add convenience and value for both communities.

Moreover, a significant body of VPR research has focused on proposing techniques that are invariant to viewpoint, illumination and seasonal variations, all of which are major challenges in VPR. However, these techniques have usually been assessed qualitatively in the past using a rough categorisation of invariance such as `mild', `moderate', `high' and `extreme', etc., which are subjective and ambiguous. Although seasonal variations are difficult to quantify, viewpoint and illumination variations can be modelled by quantitative metrics. Therefore, another key focus of this research is to quantify the invariance of VPR techniques to viewpoint and illumination changes. We utilise the detailed variation-quantified Point Feature dataset (\cite{aanaes2012interesting}) and integrate it into our framework to numerically and visually interpret the invariance of techniques. This quantified variation is obtained by taking images of a fixed scene from various angles and distances, under different illumination conditions, as explained later in sub-section \ref{Invariance_Quantification_Setup}. Since the Point Features dataset is a synthetically-created dataset, we also include the QUT multi-lane dataset (\cite{skinner2016high}) and MIT multi-illumination dataset (\cite{murmann19}), which each respectively represent quantified variations in viewpoint and illumination in a real-world setting.

Furthermore, we take the opportunity to present a detailed meta-analysis enabled by VPR-Bench. We have integrated Receiver-Operating-Characteristic (ROC) curves into VPR-Bench to analyse the ability of VPR techniques to find `new places', i.e. true-negatives, which are generally not available in Precision-Recall type metrics. We perform experiments and present analysis on the distribution of true-positives within a sequence in our work, which helps to understand the utility of VPR techniques based on spatial gaps between consecutive true-positives. In addition to the metric-based performance evaluations, we also discuss case-studies on ground-truth manipulation that can lead to varying state-of-the-art, and the CPU vs GPU performance differences for deep-learning-based VPR techniques. The descriptor size of VPR techniques also affects VPR performance and we analyse these effects in our work. The retrieval time of VPR techniques is compared with platform dynamics to yield insights into the relation between map-size, encoding-times, matching times and platform velocity. A sub-section is dedicated to discussing the impacts and usage of viewpoint variance instead of invariance for VPR techniques in changing application scenarios. Finally, the source-code for 
our comprehensive framework will be made fully public, and all datasets with their associated ground-truths will be re-released. An overview of our framework is shown in Fig. \ref{VPRBench_described_PDF}.

In summary, our main contributions are:

\begin{enumerate}

    \item We present a systematic analysis of VPR by employing the largest collection of techniques, datasets and evaluation metrics to date from the computer vision and the robotics VPR communities, such that we accommodate a large number of scenarios, including very-small scale datasets to large-scale datasets, indoor to outdoor and natural environments, moderate to extreme viewpoint and conditional variations and several evaluation metrics that complement each other.
     
    \item We present an open-source, fully-integrated, extensive framework for evaluating VPR performance. We re-implement a number of VPR techniques based on our unified templates and re-structure datasets and their ground-truths into consistent and compatible formats, which we will be re-releasing, thus providing a pre-established go-to strategy for employing a variety of metrics, datasets and popular VPR techniques for all new evaluations on a common-ground.

    \item We quantify the notion of viewpoint and illumination invariance of VPR techniques by employing a detailed variation-quantified Point Features dataset. We then further extend our findings to 2 real-world, variation-quantified datasets, namely QUT multi-lane dataset and MIT multi-illumination dataset.
    
    \item We present a number of different analyses within the VPR performance evaluation landscape, including the effects of acceptable ground-truth manipulation on rankings, the trade-offs between viewpoint variance vs invariance, the effects of descriptor size on the performance of a technique, the CPU vs GPU computational performance rankings and the trends of image retrieval times' variation with changing map-size on par with a platform's dynamics.
\end{enumerate}

The remainder of the paper is organized as follows. In Section \ref{literaturereview}, a comprehensive literature review regarding VPR state-of-the-art is presented. Section \ref{VPR-Bench_Framework} presents the details of the evaluation setup employed in this work. Section \ref{resultsandanalysis} puts forth the results and analysis obtained by evaluating the contemporary VPR techniques on public VPR datasets, along with insights into invariance quantification. Finally, conclusions and future directions are presented in Section \ref{conclusionsandfuturework}.  

\section{Literature Review}
\label{literaturereview}
The detailed theory behind Visual Place Recognition (VPR), its challenges, applications, proposed techniques, datasets and evaluation metrics have been thoroughly reviewed by \cite{lowry2015visual}, and more recently by~\cite{garg2021your, zhang2021visual, masone2021survey}.

Before diving deep into the core VPR literature review, it is important to co-relate and distinguish VPR research from closely related topics including visual-SLAM, visual-localisation and image matching (or correspondence problem), to set the scope of our research. A huge body of robotics research in the past few decades has been dedicated to the problem of simultaneously localising and mapping an environment, as thoroughly reviewed by \cite{cadena2016past}. Performing SLAM with only visual information is called visual-SLAM, and \cite{davison2007monoslam} were the first to fully demonstrate this. The localisation part of visual-SLAM can be broadly divided into two tasks: 1) Computing change in camera/robot pose while performing a particular motion, using inter-frame(s) co-observed information, 2) Recognising a previously seen place to perform loop-closure. The former is usually referred to as visual-localisation and  \cite{nardi2015introducing} developed an open-source framework in this context for evaluating visual-SLAM algorithms. The latter is essentially an image-retrieval problem in the computer vision community, and within the context of robotics has been referred to as Visual Place Recognition (\cite{lowry2015visual}). Image matching (also referred to as keypoint matching or correspondence problem in some literature) consists of finding repeatable, distinct and static features in images, describing them using condition-invariant descriptors and then trying to locate co-observed features in various images of the same scene. It is primarily targeted for visual-localisation, 3D-model creation, Structure-from-Motion and geometric-verification, but can also be utilised for VPR. \cite{jin2020image} developed an evaluation framework along these lines for matching images across wide baselines. It is important to note here that image matching can also be included as a sub-module of a VPR system. \cite{torii2019large} demonstrated that such a system can achieve accurate localisation without the need for large-scale 3D-models.

VPR has therefore generally been approached as a retrieval problem that focuses on retrieving a correct match (either as the best-match or among the Top-N matches) from a reference database given a query image, under varying viewpoint and conditions. However, VPR may also be combined with local-feature matching (geometric verification) to perform highly accurate localisation at increased computational cost, as shown by \cite{sattler2016large}, \cite{camarahighly} and \cite{sarlin2019coarse}. The existing literature in VPR can largely be broken down into: 1) Handcrafted feature descriptors-based VPR techniques, 2) Deep-learning-based VPR techniques, 3) Regions-of-Interest-based VPR techniques. All of these major classes have their trade-offs between matching performance, computational requirements and approach salience.

\textbf{Local Feature Descriptors-based VPR: }Handcrafted feature descriptors can be further sub-divided into two major classes: local feature descriptors and global feature descriptors. The most popular local feature descriptors developed in the vision community include Scale Invariant Feature Transform (SIFT \cite{lowe2004distinctive}) and Speeded Up Robust Features (SURF \cite{bay2006surf}). These descriptors have been used for the VPR problem by \cite{se2002mobile}, \cite{andreasson2004topological}, \cite{stumm2013probabilistic}, \cite{kovsecka2005global} and \cite{murillo2007surf}. A probabilistic visual-SLAM algorithm was presented by  \cite{cummins2011appearance}), namely Frequent Appearance-based Mapping (FAB-MAP), that used SURF as the feature detector/descriptor and represented places as visual words. Odometry information was  integrated into FAB-MAP by \cite{maddern2012cat} to achieve Continuous Appearance Trajectory-based SLAM (CAT-SLAM) using a Rao–Blackwellised particle filter. CenSurE (Center Surround Extremas by \cite{agrawal2008censure}) is another popular local feature descriptor and which has been used for VPR by \cite{konolige2008frameslam}. FAST (\cite{rosten2006machine}) is a popular high speed corner detector that has been used in combination with the SIFT descriptor for SLAM by \cite{mei2009constant}. Matching of local feature descriptors is a computationally intense process which has been addressed by the Bag of visual Words (BoW \cite{sivic2003video}) approach. BoW collects visually similar features in dedicated bins (pre-defined or learned by training a visual-dictionary) without topological consideration, enabling direct matching of BoW descriptors. Some of the techniques using BoW for VPR include the works of \cite{angeli2008incremental}, \cite{ho2007detecting},  \cite{wang2005combining} and \cite{filliat2007visual}. \cite{arandjelovic2014dislocation} present a new methodology to estimate the distinctiveness of local feature descriptors in a query image from closely related matches in reference descriptor space, thereby utilising salient features within the image. While the hand-crafted local features like SIFT and SURF had been widely used for VPR, recent advances include learnt local features, for example, LIFT~(\cite{yi2016lift}), R2D2~\cite{revaud2019r2d2}, SuperPoint (\cite{detone2018superpoint}) and D2-net (\cite{dusmanu2019d2}). \cite{noh2017large} designed a deep-learning-based local feature extractor and descriptor, namely DELF, that is used with geometric verification for large-scale image retrieval.

\textbf{Global Feature Descriptors-based VPR: }Global feature descriptors create a holistic signature for an entire image and Gist (\cite{oliva2006building}) is one of the most popular global feature descriptor. Working on panoramic images, \cite{murillo2009experiments}, \cite{singh2010visual} used Gist for VPR.  \cite{sunderhauf2011brief} combined Gist with BRIEF (\cite{calonder2011brief}) to perform large scale visual-SLAM. \cite{badino2012real} used Whole-Image SURF (WI-SURF), which is a global variant of SURF to perform place recognition. Operating on sequences of raw RGB-images, Seq-SLAM (\cite{milford2012seqslam}) uses normalized pixel-intensity matching in a global fashion to perform VPR in challenging conditionally-variant environments. The original Seq-SLAM algorithm assumes constant speed of the robotic platform, thus, \cite{pepperell2014all} extended Seq-SLAM to consider variable speed instead. \cite{mcmanus2014scene} extract scene signatures from an image by utilising some \textit{a priori} environment information and describe them using HOG-descriptors. DenseVLAD presented by \cite{torii201524} is a Vector-of-Locally-Aggregated-Descriptors-based approach using densely sampled SIFT keypoints, which has been shown to perform similar to deep-learning-based techniques in \cite{sattler2018benchmarking} \cite{torii2019large}. A more recent usage of traditional handcrafted feature descriptors for VPR was presented in CoHOG (\cite{zaffar2020cohog}) which focuses on entropy-rich regions in an image and uses HOG as the regional descriptor for convolutional-regional matching.

\textbf{Deep Learning-based VPR: }Similar to other domains of computer vision, deep-learning and especially Convolutional-Neural-Networks (CNNs) are a game-changer for the VPR problem by achieving unprecedented invariance to conditional changes. By employing off-the-shelf pre-trained neural nets, \cite{chen2014convolutional} used features from the Overfeat Network (\cite{sermanet2014overfeat}) and combined it with the spatial filtering scheme of Seq-SLAM. This work was followed up by \cite{chen2017deep}, where two neural networks (namely AMOSNet and HybridNet) were trained specifically for VPR on the Specific Places Dataset (SPED). AMOSNet was trained from scratch on SPED, while the weights for HybridNet were initialised from the top-5 convolutional layers of Caffe-Net (\cite{krizhevsky2012imagenet}). An end-to-end neural-network-based holistic descriptor NetVLAD is introduced by \cite{arandjelovic2016netvlad}, where a new VLAD (Vector-of-Locally-Aggregated-Descriptors (\cite{jegou2010aggregating})) layer is integrated into the CNN architecture achieving excellent place recognition results. A convolutional auto-encoder network is trained in an unsupervised fashion by \cite{merrill2018lightweight}, utilizing HOG-descriptors of images and synthetic viewpoint variations for training. The work of \cite{noh2017large} was extended to DELG (DEep Local and Global Features by \cite{cao2020unifying}) combining generalized mean pooling for global descriptors and attention mechanism for local features. Recently, \cite{simeoni2019local} presented that state-of-the-art image-retrieval performance can be achieved by mining local features from CNN activation tensors and by performing spatial verification on these channel-wise local features, which can be then converted into global image signatures by using Bag-of-Words description. The work of \cite{radenovic2018fine} (GeM) introduces a new trainable `Generalised Mean' layer into the deep image-retrieval architecture which has been shown to provide a performance boost. \cite{chancan2020hybrid} draw their inspiration from brain architectures of fruit flies, train a sparse two-layer neural-network and combined it with Continuous-Attractor-Networks to summarise temporal information.

\textbf{Regions-of-Interest-focused VPR: }Researchers have used Regions-of-Interest (ROIs) to introduce the concept of salience into VPR, and to ensure that static, informative and distinct regions are used for place recognition. Regions of Maximum Activated Convolutions (R-MAC) are used by \cite{tolias2015particular}, where max-pooling across cropped areas in CNN layers' features define/extract ROIs. This work on R-MAC is further advanced by \cite{gordo2017end}, where a Siamese Network is trained with a Triplet loss on the Landmarks dataset (\cite{babenko2014neural}). However, \cite{revaud2019learning} argue that ranking-based loss functions (image-pairs, triplet-loss, n-tuples, etc.) are not optimal for the final task of achieving higher mAP and therefore propose a new ranking-loss that directly optimizes mAP. This mAP-based ranking loss function which in combination with GeM achieves state-of-the-art retrieval performance. High-level features encoded in earlier neural-network layers are used for region-extraction and the following low-level features in later layers are used for describing these regions in the work of \cite{chen2017only}. This work is then followed-up with a flexible attention-based model for region extraction by \cite{chen2018learning}. \cite{khaliq2019holistic} draw their inspiration from NetVLAD and R-MAC, thereby combining VLAD description with ROI-extraction to show significant robustness to appearance- and viewpoint-variation. Photometric-normalisation using both handcrafted and learning-based methodology is investigated by \cite{jenicek2019no} to achieve illumination-invariance for place recognition.

\textbf{Other Interesting Approaches to VPR: } Other interesting approaches to place recognition include semantic-segmentation-based VPR (as in \cite{arandjelovic2014visual}, \cite{mousavian2015semantically}, \cite{stenborg2018long}, \cite{schonberger2018semantic}, \cite{naseer2017semantics}) and object-proposals-based VPR (\cite{hou2018evaluation}), as recently reviewed by~\cite{garg2020semantics}. For images containing repetitive structures, \cite{torii2013visual} proposed a robust mechanism for collecting visual words into descriptors. Synthetic views are utilized for enhanced illumination-invariant VPR in \cite{torii201524}, which shows that highly condition-variant images can still be matched, if they are from the same viewpoint. In addition to image retrieval, significant research has been performed in semantic mapping to select images for insertion into a metric, topological or topometric map as nodes/places. Semantic mapping techniques are usually annexed with VPR image retrieval techniques for real-world Visual-SLAM, see the survey by \cite{kostavelis2015semantic}. Most of these semantic mapping techniques are based on Bayesian-surprise (\cite{ranganathan2013detecting}, \cite{girdhar2010online}), coresets (\cite{paul2014visual}), region proposals (\cite{demir2018automated}), change-point detection (\cite{topp2008detecting}, \cite{ranganathan2013detecting}) and salience-computation (\cite{zaffar2020memorable}).

While the VPR literature consists of a large number of VPR techniques, we have currently implemented $8$ state-of-the-art techniques into the VPR-Bench framework. We have also added the provision to integrate results (image descriptors) from other techniques, which has been demonstrated by integrating DenseVLAD and GeM into the benchmark. We plan to increase this number over time due to the modular nature of our framework with the help of the VPR community. 

\textbf{Benchmarks for Visual-localisation: }
Within the performance evaluation landscape, if we broaden our scope, it is evident that ours is not the first attempt at benchmarking visual-localisation at scale and previous attempts exist, which have led to the rapid development in this domain. From the computer vision perspective, the well-established visual-localisation benchmark\footnote{www.visuallocalization.net} has been hosted for the past few years as workshops in top computer vision conferences. This benchmark was initially focused on 6-DOF pose estimates, but has recently also included VPR (image-retrieval) benchmarking by combining with the Mapillary Street Level Sequences (MSLS) dataset (\cite{warburg2020mapillary}) in ECCV 2020, although MSLS is mainly focused on sequences. The benchmarks have usually been organised as challenges (which have their own dedicated utility), where relevant evaluation papers also exist, e.g. the recent detailed works from \cite{torii2019large} and \cite{sattler2018benchmarking}. Google also proposed the Landmarks dataset with focus on both place/instance-level recognition and retrieval: Google Landmark V1 dataset (\cite{noh2017large}) and Google Landmark V2 dataset (\cite{weyand2020google}). These benchmark datasets (and other similar datasets like Oxford Buildings, Paris Buildings etc.) and their associated evaluation metrics serve great value to the landmark recognition/retrieval problem, but focus on a particular category of datasets containing distinctive architectures, which may not be the primary focus of the robotics-centered VPR community requiring localisation-estimates throughout a continuous traversal that may be indoor, outdoor, natural and any/all others. Here, another divide is that of direct vs indirect evaluation of image retrieval, where the former directly quantifies the performance of a VPR system's output, while the latter assesses the performance of a larger system using end-task metrics such that VPR is only a module of this system's pipeline. The scope of VPR-Bench is limited to the direct evaluation of VPR.

\textbf{Direct and Indirect Evaluation Metrics for VPR: }
With the extensive applications of VPR and therefore the correspondingly large number of relevant evaluation metrics, a higher-level breakdown can consist of two categories: direct and indirect evaluation metrics. Direct evaluation metrics are those metrics that directly measure the performance of a VPR system based on the images retrieved by the system from a given reference database for a set of query images. This direct evaluation of VPR systems is the scope of our work and discussed at length in the following paragraph. On the other hand, indirect evaluation metrics for VPR are those metrics where VPR is only a part of the particular system's pipeline. In such cases, the evaluation metric is measuring the performance of the complete pipeline, where indirectly a good performing VPR module contributes to but is not the only determinant of achieving higher overall system performance. Some key examples of such indirect metrics within the Visual-SLAM paradigm are Absolute-Trajectory-Error (ATE) and Relative-Pose-Error (RPE), as presented in the RGB-D Visual-SLAM benchmark by \cite{sturm2012benchmark}. Another commonly observed pipeline for 6-DOF camera-pose estimation with respect to a given scene is VPR followed by local feature matching, where the VPR module provides the initial coarse location estimate, which is then refined by local feature matching to yield 6-DOF camera pose. In such a case, the overall pipeline evaluation indirectly estimates VPR performance, as done by \cite{sattler2018benchmarking}.

Within direct performance evaluation, the most dominant VPR evaluation metric in robotics literature (\cite{lowry2015visual}) has been Area-under-the-Precision-Recall curves (denoted usually as AUC-PR or simply AUC), which tries to summarise the Precision-Recall curves in a single quantified value. AUC-PR favours techniques that can retrieve the correct match as the top ranked image, thus favouring applications that require highly precise localisation estimates. The reasons for more common use of PR-curves instead of Receiver Operating Characteristics curves (ROC-curves) in VPR are the imbalanced nature of the datasets and the usual lack of true-negatives in datasets/evaluations. There is extensive VPR literature employing AUC-PR, for example, \cite{lategahn2013learn}, \cite{cieslewski2017efficient}, \cite{ye2017place}, \cite{camara2019spatio}, \cite{khaliq2019holistic} and \cite{tomitua2021sequence}. Other than AUC-PR, F1-score has also been used in VPR evaluations predominantly by the robotics-focused VPR community, for example by \cite{mishkin2015place}, \cite{sunderhauf2015performance}, \cite{talbot2018openseqslam2}, \cite{garg2018lost} and \cite{hausler2019multi}, to list a few. However, metrics like AUC-PR and F1-score quantify the performance of a VPR technique without considering the geometric distribution of true-positives within the trajectory. But since robotics is mostly concerned with achieving localisation every few meters, \cite{porav2018adversarial} present a new metric/analysis to compute the VPR performance, using the maximum distance traversed by a robot without achieving a true-positive/localisation/loop-closure. Recently, \cite{ferrarini2020exploring} presented a new metric Extended Precision (EP) for VPR evaluation that is based on Precision@100\% Recall and Recall@100\% Precision. In our previous work (\cite{zaffar2020cohog}), we had presented PCU (Performance-per-Compute-Unit) as an evaluation metric for VPR, which combines place recognition precision with feature encoding time.

Recall@N (or RecallRate@N) is a dominant evaluation metric in the computer vision VPR community, which considers a retrieval to be true-positive for a given query, if the correct ground-truth image is within the Top-N retrieved images. Recall@N has been used by e.g. \cite{perronnin2010large}, \cite{torii2013visual}, \cite{arandjelovic2014dislocation}, \cite{torii201524}, \cite{arandjelovic2016netvlad} and \cite{uy2018pointnetvlad}. For multiple correct matches in the database, Recall@N does not consider how many of the correct matches for a given query were retrieved by a VPR technique, therefore mean-Average-Precision (mAP) has also been extensively used by the computer vision VPR/image-retrieval community. Some of the literature that has employed mAP as an evaluation metric for VPR includes \cite{jegou2008hamming}, \cite{gordo2016deep}, \cite{sattler2016large}, \cite{gordo2017end}, \cite{revaud2019learning} and \cite{weyand2020google}. Other than these metrics, Recall@Reduced Precision has also been used as an evaluation metric (\cite{tipaldi2013geometrical}) for place recognition. For computational analysis, feature encoding time, descriptor matching time and descriptor size have been the key metrics for both the communities.

It is evident that a large number of evaluation metrics can be employed for assessing the performance a VPR system and the selection is usually dependent upon the underlying application. However, it is also possible for the metrics from one community to be of value to the other community, such that the the above discussed distribution of metrics is not depicting absoluteness but only dominant trends/applications. For example, Recall@N and Recall@Reduced Precision are also useful for robotic systems that can discard a small number of false-positives, e.g. by using outlier rejection in SLAM, false-positive prediction, ensemble-based approaches and geometric verification. Similarly, mAP-based evaluations can support the creation of additional constraints for map optimisation in SLAM. The discussion and analysis on evaluation metrics scales quickly in the dimension of the number of metrics discussed. To limit the scope of this work, we have only used AUC-PR, RecallRate@N, true-positive trajectory distribution, feature encoding time, descriptor matching time and feature descriptor size as our evaluation metrics in this work. We discuss these metrics systematically and at length later in sub-section \ref{Evaluation_Metrics}.

\textbf{Invariance Evaluation of VPR: }The effect of viewpoint and appearance variations on visual place recognition has been well studied in the past, aiming to understand the limitations of different approaches. \cite{chen2014convolutional} and \cite{sunderhauf2015performance} evaluated different convolutional layers of off-the-shelf CNNs for their performance on VPR and concluded that mid-level and higher-level layers were respectively more robust to appearance and viewpoint variations. \cite{garg2018don't} validated this trend on a more challenging scenario of opposing viewpoints while also showcasing catastrophic failure of viewpoint-dependent representations due to 180 degrees shift in camera viewpoint. In a subsequent work, \cite{garg2018lost} presented an empirical study on the amount of translational offset needed to match places from opposing viewpoints in city-like environments. \cite{pepperell2015automatic} studied the effect of scale on VPR performance when using side-view imagery and travelling in different lanes within city suburbs and on a highway. \cite{cheron2018evaluation} evaluated the performance of local features for recognition using `\textit{free viewpoint} videos' and concluded that traditional hand-crafted features demonstrated more viewpoint-robustness than their learnt counterparts. \cite{kopitkov2018bayesian} characterized the viewpoint-dependency of CNN feature descriptors and used it to improve probabilistic inference of a robot's location. In this work, we present a more formal treatment to the effect of viewpoint and appearance variations on VPR by utilizing the Points Features dataset~(\cite{aanaes2012interesting}) for performance quantification. We then extend this analysis to real-world scenarios using the QUT Multi-Lane dataset~(\cite{skinner2016high}) and MIT Multi-Illumination dataset~(\cite{murmann19}).

\section{VPR-Bench Framework}
\label{VPR-Bench_Framework}
This section introduces the details of our novel VPR-Bench framework, including the task formulation, datasets, techniques, evaluation metrics and the invariance quantification module, respectively.

\subsection{VPR Task Formulation}
\label{Problem_Formulation}
Here, we formally define what a VPR system represents throughout this paper.

Let $Q$ be a query image and $M_R$ be a list/map of $R$ reference images. The feature descriptor(s) of a query image $Q$ and reference map $M_R$ can be denoted as $F_Q$ and $F_M$, respectively. If a technique uses ROI-extraction, $F_Q$ will hold within it all the required information in this regards, including location of regions, their descriptors and corresponding salience. The input $Q$ can also be a sequence of Query images and any other pre/post-processed form of a query candidate. For a query image $Q$, given a reference map $M_R$, let us denote the best matched image/place by a VPR technique as $P$ (where, $P \in M_R$) with a matching score $S$. The matching score $S$ can be defined as $S \in [0,1]$. The confusion matrix (matching scores with all reference images) can be denoted as $C$. Based on these notations, the following algorithm represent a VPR system.

\begin{algorithm} [H]
\small
\renewcommand\thealgorithm{}
\caption{A Generic VPR System}
\begin{algorithmic}

\STATE $\textbf{Given:} \; \; Q, M\textsubscript{R}$
\STATE $\textbf{Required:} \; \; P, S, C$

\STATE $\textbf{def} \; compute\_query\_desc \; (Q)$
\STATE $\; \; \; Preprocessing \; Steps$
\STATE $\; \; \; Function \; Body$
\STATE $\; \; \; Postprocessing \; Steps$
\STATE $\; \; \; return \; F_Q$ \\

\STATE $\textbf{def} \; compute\_map\_features \; (M_R)$
\STATE $\; \; \; Preprocessing \; Steps$
\STATE $\; \; \; Function \; Body$
\STATE $\; \; \; Postprocessing \; Steps$
\STATE $\; \; \; return \; F_M$ \\

\STATE $\textbf{def} \; perform \_VPR \; (F_Q, F_M)$
\STATE $\; \; \; Preprocessing \; Steps$
\STATE $\; \; \; Function \; Body$
\STATE $\; \; \; Postprocessing \; Steps$
\STATE $\; \; \; return \; P, \; S, \; C$ 

\STATE $\textbf{def} \; main \; ()$
\STATE $\; \; \; F_M \; = \; compute\_map\_features \; (M_R)$
\STATE $\; \; \; F_Q \; = \; compute\_query\_desc \; (Q)$
\STATE $\; \; \; P, \; S, \; C \; = \; perform \_VPR \; (F_Q, F_M)$
\STATE $\; \; \; store \; P, \; S, \; C$ 

\end{algorithmic}
\addtocounter{algorithm}{-1}
\end{algorithm}

\subsection{Evaluation Datasets}
\label{evaluation_datasets}
In this section, we present the existing patterns and features of datasets in VPR and then  discuss each of the datasets that have been used in this work by dividing them into outdoor and indoor datasets categories.

\begin{figure*}[t]
\begin{center}
%\fbox{\rule{0pt}{2in} \rule{0.9\linewidth}{0pt}
\includegraphics[width=1\linewidth]{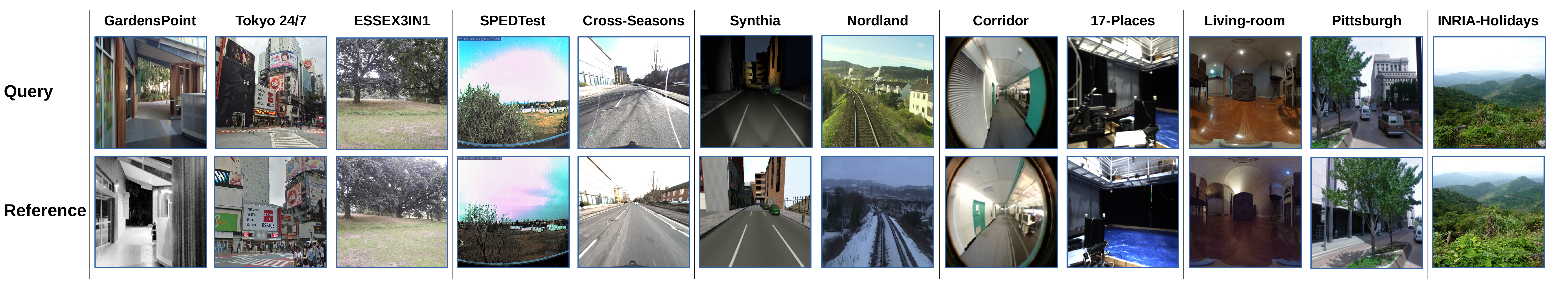}
\end{center}
\caption{Sample images from all $12$ VPR datasets employed in this work are presented here. These datasets span many different environments, including cities, natural scenery, train-lines, rooms, offices, corridors, buildings, busy-streets and such.}
\label{VPR_datasets_samples}
\vspace{-0mm}
\end{figure*}

\subsubsection{Dataset Considerations in VPR-Bench}
\label{Datasets_in_VPR}
All the datasets that have been employed to date for VPR evaluation comprise of multiple views of the same environment that may have been extracted under different seasonal, viewpoint and/or illumination conditions. These views are mostly available in the form of monocular images and are structured as separate folders representing query and reference images. However, these views may have been extracted from a traversal or a non-traversal-based mechanism. For the former, consecutive images within a folder (query/reference) usually have overlapping visual content, while for the latter, images within a folder are independent.  Accompanying these folders is usually some level of ground-truth information, which has been represented in various ways (e.g, CSV, numpy arrays, pickle files containing frame-level correspondence, GPS, pose information etc.) for different datasets. In some cases, the ground-truth is not explicitly provided, as images with the same index/name represent the same place. 

For most traversal-based datasets, there is no single correct match for a query image, because images which are geographically close-by can be considered as the same place, leading to a range requirement for ground-truth matches instead of a single match/value. For such datasets and viewpoint-invariance in general, defining a correct ground-truth is `tricky' because depending upon the acceptable level of viewpoint invariance for a VPR technique, the underlying ground-truth can be manipulated to change the performance ranking, as shown later in sub-section \ref{acceptable_groundtruth_manipulation}. Another key challenge is the relation between visual-overlap, scene-depth and physical distance. In an outdoor environment (e.g. highway), frames that are 5 meters apart may have significant visual overlap due to high scene depth, while frames that are 5 meters apart in an indoor environment may be visually very different due to low scene-depth and therefore frame-range-based ground-truth for most VPR datasets includes manual adjustment of ground-truth frame-range given visual overlap sanity checks.

Generally, there is a trade-off between pose accuracy and viewpoint invariance, where none of these can explicitly define a hard requirement from a VPR system. If a VPR system is being used as the primary localisation system (robotics perspective), higher pose accuracy is desired and the system should have \textit{viewpoint-variance}, while for retrieving maximum matches of a place from the reference database (computer vision perspective), \textit{viewpoint-invariance} is the key requirement. For the robotics perspective, pose inaccuracy can be reduced at increased computational cost by using image-matching as a subsequent pose refinement stage. Therefore, some viewpoint invariance (usually defined by a few meters) has always been required from a VPR system in both the communities. To address this `loose' nature of viewpoint-invariance definition of a VPR system, we have taken the following steps:

\begin{enumerate}
    \item We have integrated datasets that contain a large variation in the acceptable ground-truth viewpoint variance: ranging from the minimally acceptable viewpoint variation in the Corridor dataset to the large acceptable viewpoint variations of the Tokyo 24/7 dataset, thus to cover a broader audience.
    
    \item We have provided an extensive analysis on the effects of changing acceptable levels of viewpoint invariance in sub-section \ref{acceptable_groundtruth_manipulation}.
    
    \item As for consistency in VPR research and performance reporting, it is essential to affix a unified template for all of these VPR datasets, we will be re-releasing all datasets in a VPR-Bench compatible mode with their associated ground-truth information.
    
\end{enumerate}

Despite the extensive collection of datasets in this work, there are still scenarios which are not represented in these datasets, e.g. extreme weather conditions, aerial and underwater platforms, opposing views and motion-blur resulting from high-speed platforms. We have designed VPR-Bench as per unified templates to allow integration of new datasets. Further details of the datasets template are provided in the appendix of this paper.

\begin{table*}
\footnotesize
\centering
\caption{The 12 VPR-Bench datasets integrated into VPR-Bench and used in this study are enlisted here. The sign \~{} for image resolutions (pixels $\times$ pixels) indicates datasets where image resolution varies in-between different images of the dataset and we have therefore specified the common resolution observed in that dataset.}
\label{tab:1}       % Give a unique label
\setlength\tabcolsep{1.0pt}
\begin{tabular}{cccccccc}
\hline\noalign{\smallskip}
Dataset & Environment & Queries & References & Viewpoint Change & Conditional Change & Query Res. & Ref Res.\\
\noalign{\smallskip}\hline\noalign{\smallskip}
GardensPoint & University Campus & 200 & 200 & Lateral & Day-Night & 960 $\times$ 540 & 640 $\times$ 360\\
Tokyo 24/7 & Outdoor & 315 & 75984 & 3D & Day-Night & 3264 $\times$ 2448 & 640 $\times$ 480\\
ESSEX3IN1 & University Campus & 210 & 210 & 3D & Illumination  & 720 $\times$ 720 & 1080 $\times$ 1080\\
SPEDTest & Outdoor & 607 & 607 & None & Seasonal and Weather  & \~320 $\times$ \~240 & \~320 $\times$ \~240\\
Cross-Seasons & City-like & 191 & 191 & Lateral (Occasional) & Dawn-Dusk  & 1024 $\times$ 1024 & 1024 $\times$ 1024\\
Synthia & City-like (Synthetic) & 813 & 911 & Lateral & Time and Season & 300 $\times$ 200 & 300 $\times$ 200\\
Nordland & Train Journey & 2760 & 27592 & None & Seasonal & 640 $\times$ 360 & 640 $\times$ 360\\
Corridor & Indoor & 111 & 111 & Lateral & None & 160 $\times$ 120 & 160 $\times$ 120\\
17-Places & Indoor & 406 & 406 & Lateral & Day-Night & 640 $\times$ 480 & 640 $\times$ 480\\
Living-room & Indoor & 32 & 32 & Lateral & Day-Night & 1792 $\times$ 896 & 1792 $\times$ 896\\
Pittsburgh & Outdoor & 1000 & 23000 & 3D & None & 640 $\times$ 480 & 640 $\times$ 480\\
INRIA Holidays & Outdoor & 300 & 512 & Lateral/3D & None & \~250 $\times$ \~185 & \~250 $\times$ \~185\\
\noalign{\smallskip}\hline
\end{tabular}
\end{table*}

\subsubsection{Outdoor Environment}
We have integrated multiple outdoor datasets in our framework representing different types and levels of viewpoint-, illumination- and seasonal-variations. Details of these datasets have been summarised in Table 1 and sample images are shown in Fig. \ref{VPR_datasets_samples}. Each of these datasets has a particular attribute to offer, that lead to its selection and they are briefly discussed below. 

The GardensPoint dataset was created by \cite{arren_glover_2014_4590133} and first used for VPR by \cite{chen2014convolutional}, where two repeated traversals of the Gardens Point Campus of Queensland University of Technology, Brisbane, Australia were performed with varying viewpoints in day and night times. A huge body of robotics-focused VPR research has used this dataset for reporting their VPR matching performance, as it depicts outdoor, indoor and natural environments, collectively. We have only used the day and night sequences in our work because they contain both the viewpoint and conditional change. The Tokyo 24/7 dataset was proposed by \cite{torii201524}, which consists of 3D viewpoint-variations and time-of-day variations. We use version 2 of the query images, as suggested by the authors of \cite{torii201524} and \cite{arandjelovic2016netvlad} to maintain comparability. It is one of the most challenging datasets for VPR due to the sheer amount of viewpoint- and conditional-variation, and has been used by both the robotics and vision communities. The ESSEX3IN1 dataset was proposed by \cite{zaffar2020memorable} and is the only dataset designed with focus on perceptual aliasing and confusing places/frames for VPR techniques. The SPEDTest dataset was introduced by \cite{chen2018learning} and consists of low-quality, high scene-depth frames extracted from CCTV cameras across the world. This dataset has the unique attribute of covering a huge variety of scenes from all across the world under many different weather, seasonal and illumination conditions. The Synthia dataset was introduced in \cite{ros2016synthia} and represents a simulated city-like environment in various weather, seasonal and time of day conditions. In this paper, we have used the night images from Synthia Video Sequence 4 (old European town) as query and the fog images as reference from the same sequence. The Cross-Seasons dataset employed in our work represents a traversal from \cite{larsson2019cross}, which is a subset of the Oxford RobotCar dataset (\cite{maddern20171}). This dataset represents a challenging real-world car traversal from dawn and dusk conditions. One of the widely employed datasets for VPR is the Nordland dataset, developed by \cite{nordlanddataset} and introduced to VPR evaluation by \cite{sunderhauf2013we}, which represents a 728 kilometers of train journey in Norway during Summer and Winter seasons. As Nordland dataset represents natural (non-urban), outdoor environment, which is unexplored in any other dataset, we have integrated it into VPR-Bench. From the computer vision community, in addition to Tokyo 24/7, we have used the Pittsburgh dataset (\cite{torii2013visual}) and the INRIA Holidays dataset (\cite{jegou2008hamming}) to bridge the important gap between the two communities. We use only the query images of Pittsburgh dataset because this represents the only large-scale dataset in our framework that has 3D viewpoint-variation without any conditional variation. The INRIA Holidays dataset, similar to the SPEDTest dataset, explores a very large variety of scenes but also includes indoor scenes as well, and uses the highly relevant egocentric viewpoint unlike the CCTV-based SPEDTest. These datasets are still only a subset from an apparent zoo of datasets available for VPR evaluation. Despite the large number of outdoor datasets used in this work, there are still scenarios that are not covered here, including extreme weather conditions, opposing views, motion-blur, aerial and underwater datasets.

\subsubsection{Indoor Environment}
A significant focus in recent research in VPR has primarily been on evaluation on outdoor datasets, so we also incorporate indoor environments into VPR-Bench, which are usually a key area of study within robot autonomy. While indoor datasets, usually do not represent the seasonal variation challenges as outdoor datasets and the level of viewpoint-variation is relatively lesser than outdoor datasets, they do contain dynamic objects like humans, animals or changing setup/environment configurations, less-informative content and perceptual-aliasing. The details of these datasets have been summarised in Table 1 and sample images are shown in Fig. \ref{VPR_datasets_samples}. We have briefly discussed the currently available indoor datasets in VPR-Bench, in the following paragraph.

We have integrated the 17-Places dataset introduced by \cite{sahdev2016indoor} into VPR-Bench, which consists of a number of different indoor scenes, ranging from office environment to labs, hallways, seminar rooms, bedrooms and many other. This dataset exhibits both viewpoint- and conditional-variations. We also use the viewpoint-variant Corridor dataset, introduced by \cite{milford2013vision}, which represents the challenge of low-resolution and feature-less images ($160 \times 120$ pixels) for vision-based place recognition. \cite{mount20162d} introduced the living-room dataset for home-service robots, which represents indoor environment from a highly relevant and challenging viewpoint of cameras mounted close-to-ground level. This dataset only contain 32 queries and 32 references, we deliberately use such a small-scale dataset to see the ordering of VPR techniques on very small-scale datasets. 

\subsubsection{Ground-truth Information}
Because we have utilised a variety of different datasets from both the robotics and the computer vision communities, which are also from both indoor and outdoor environments, the underlying ground-truth information is varying. We have used the ground-truth information provided by the original contributors of these datasets (or in some cases the modified ground-truths used in recent evaluations) and reformatted these into ground-truth compatible to the templates developed for VPR-Bench. All the datasets and their ground-truths will be re-released and therefore we have only briefly presented this ground-truth information in Table \ref{tab:groundtruth}. The ground-truth tolerance for some of the robotics-focused VPR datasets is strict in comparison to the computer vision datasets when it comes to viewpoint variance/invariance, i.e. the reference images that are geographically far apart but have some visual overlap are not considered as correct matches for the robotics datasets. Instead of relaxing the viewpoint variance for the robotics datasets and/or restricting the viewpoint variance for the computer vision datasets, we have used the original levels being used by their respective communities. 

\begin{table}
\centering
\caption{The ground-truth tolerance for the 12 VPR-Bench datasets integrated into VPR-Bench is provided here. The $\dagger$ next to Pittsburgh dataset indicates that 23 ground-truth images are available for every query image, taken at different pitch and yaw angles without any translational movement of the camera.}
\label{tab:groundtruth}       % Give a unique label
\begin{tabular}{cc}
\hline
Dataset & Ground-truth Tolerance\\
\hline
GardensPoint & $\pm$ 2 frames\\
Tokyo 24/7 & $\pm$ 25 meters \\
ESSEX3IN1 & Frame-to-frame\\
SPEDTest & Frame-to-frame\\
Cross-Seasons & $\pm$ 5 meters \\
Synthia & $\pm$ 7 meters \\
Nordland & $\pm$ 1 frames \\
Corridor & $\pm$ 2 frames \\
17-Places & $\pm$ 3 frames \\
Living-room & $\pm$ 2 frames \\
Pittsburgh & 23 frames $\dagger$ \\
INRIA Holidays & Frame-to-frame\\
\hline
\end{tabular}
\end{table}

\subsection{VPR Techniques}
\label{VPR_techniques}
In this section, we introduce the $10$ VPR techniques that have been evaluated in this work, while also providing important implementation details of these techniques that are needed to understand the experiments and results in the next Section \ref{resultsandanalysis}.

\paragraph{\textbf{HOG-Descriptor:}} Histogram-of-oriented-gradients (HOG) is one of the most widely used handcrafted feature descriptors, which actually performs very well for VPR compared to other handcrafted feature descriptors. It is a good choice for a traditional handcrafted feature descriptor in our framework, based upon its performance as shown by \cite{mcmanus2014scene} and the value it presents as an underlying feature descriptor for training a convolutional auto-encoder in \cite{merrill2018lightweight}. We use a cell size of $16 \times 16$ and a block size of $32 \times 32$ for an image-size of $512 \times 512$. The total number of histogram bins are set equal to $9$. We use cosine-matching between HOG-descriptors of various images to find the best place match.

\paragraph{\textbf{AlexNet:}} The use of AlexNet for VPR was studied by~\cite{sunderhauf2015performance}, who suggest that \textit{conv3} is the most robust to conditional variations. Gaussian random projections are used to encode the activation-maps from \textit{conv3} into feature descriptors and cosine distance is used for matching. Our implementation of AlexNet is similar to the one employed by~\cite{merrill2018lightweight}, while the code has been restructured as per the designed template. Note that AlexNet resizes input image to 227 $\times$ 227 before it is input to the neural network.  

\paragraph{\textbf{DenseVLAD:}} DenseVLAD has been proposed by \cite{torii201524}, where they densely-sample local SIFT keypoints from images, corresponding to regional widths. These keypoints are extracted at 4 different scales. The local keypoints are then converted into a global descriptor using a Vector-of-Locally-Aggregated-Descriptors (VLAD) dictionary consisting of 128 visual-words extracted by K-means clustering on a dictionary of 25M randomly-sampled descriptors. PCA-compression and whitening is performed on the final descriptor to down-sample it into a 4096 dimensional descriptor. In this work, we have formatted (as per our template) and integrated the descriptor matching data computed by the DenseVLAD code open-sourced by~\cite{torii201524} into VPR-Bench to demonstrate the utility of our framework for cases where code conversion may not be required/desired. All input images are resized to 640 $\times$ 480, similar to~\cite{torii201524}.

\paragraph{\textbf{AP-GeM:}} GeM was originally proposed by \cite{radenovic2018fine}, where they presented a new generalised-mean layer to replace the typical max-pooling and sum-pooling for feature descriptor mining from a CNN tensor. This was then upgraded by \cite{revaud2019learning}, where they have designed a new ranking-loss based on mean-Average-Precision. We have used the GeM code open-sourced by \cite{revaud2019learning} based on the ResNet101 model (namely ResNet101-AP-GeM) with an output descriptor size of 2048 dimensions. Similar to DenseVLAD, we have used the descriptor matching data computed by the original code of the respective authors and integrated that with our framework for a seamlessly straightforward integration process. ~\cite{revaud2019learning} used 800 $\times$ 800 resolution for training but performed no resizing during testing. Thus, for a fair comparison against other input resolution-independent methods such as NetVLAD and DenseVLAD, we resized input images to 640 $\times$ 480.

\paragraph{\textbf{NetVLAD:}} The original implementation of NetVLAD was in MATLAB, as released by \cite{arandjelovic2016netvlad}. The Python port of this code was open-sourced by \cite{cieslewski2018data}. The model selected for evaluation is VGG-16, which has been trained in an end-to-end manner on Pittsburgh 30K dataset (\cite{arandjelovic2016netvlad}) with a dictionary size of $64$ while performing whitening on the final descriptors. The code has been modified as per our template. The authors of NetVLAD have suggested an image resolution of 640 $\times$ 480 at inference time and we have therefore used this image resolution for all experiments.

\paragraph{\textbf{AMOSNet:}} This technique was proposed by \cite{chen2017deep}, where a CNN has been trained from scratch on the SPED dataset. The authors have presented results from different convolutional layers by implementing spatial-pyramidal pooling on the respective layers. While the original implementation is not fully open-sourced, the trained model weights have been shared by the authors. We have implemented AMOSNet as per our template using \textit{conv5} of the shared model. L1-match has been originally proposed by the authors, which is normalised for a score between $0-1$. The default implementation of AMOSNet resizes input images to 227 $\times$ 227.

\paragraph{\textbf{HybridNet:}} While AMOSNet was trained from scratch, \cite{chen2017deep} took inspiration from transfer learning for HybridNet and re-trained the weights initialised from Top-5 convolutional layers of CaffeNet (\cite{krizhevsky2012imagenet}) on SPED dataset. We have implemented HybridNet as per our template using \textit{conv5} of the HybridNet model. L1-match has been originally proposed by the authors, which is normalised for a score between $0-1$. The default implementation of HybridNet resizes input images to 227 $\times$ 227.

\paragraph{\textbf{RegionVLAD:}} Region-VLAD has been introduced and open-sourced by \cite{khaliq2019holistic}. We have modified it as per our template and have used AlexNet trained as Places365 dataset as the underlying CNN. The total number of ROIs has been set to $400$ and we have used `conv3' for feature extraction. The dictionary size is set to $256$ visual words for VLAD retrieval. Cosine similarity is subsequently used for matching descriptors of query and reference images. The default implementation of RegionVLAD resizes input images to 227 $\times$ 227.

\paragraph{\textbf{CALC:}} The use of convolutional auto-encoders for VPR was proposed by \cite{merrill2018lightweight}, where an auto-encoder network was trained in a weakly-supervised manner to re-create similar HOG-descriptors for viewpoint-variant (cropped) images of the same place. We use model parameters from $100,000$ training iteration and adapt the open-source technique as per our template. Cosine-matching is used for descriptor comparison. This is the only semi-supervised learning technique in our framework and therefore has its own particular utility. The default implementation of CALC resizes input images to 120 $\times$ 160.

\paragraph{\textbf{CoHOG:}} CoHOG is a recently proposed (\cite{zaffar2020cohog}) handcrafted feature-descriptor-based technique, which uses image-entropy for ROI extraction. The regions are subsequently described by dedicated HOG-descriptors and these regional descriptors are convolutionally matched to achieve lateral viewpoint-invariance. It is an open-source technique, which has been modified as per our template. We have used an image-size of $512 \times 512$, cell-size of $16 \times 16$, bin-size of $8$ and an entropy-threshold (ET) of $0.4$.  CoHOG also uses cosine-matching for descriptor comparison.

\subsection{Evaluation Metrics}
\label{Evaluation_Metrics}
A trend within current VPR research has shown that a single, universal metric to evaluate VPR techniques that could simultaneously extend to all applications, platforms and user-requirements does not exist. For example, a technique which has a very high-precision, but a significantly higher image-retrieval time (few seconds), may not extend to a VPR-based, real-time topological navigation system, as the localisation module will be much slower (in frames-per-second processed) than the platform dynamics. However, for situations where real-time place matching may not be required, for example, offline loop-closures for map correction, improved-representations and structure-from-motion, high precision at the cost of higher retrieval time may be acceptable. Therefore, reporting performance on a single metric may not fully present the utility of a VPR technique to the entire academic, industrial and research audience, and the application-specific communities within them. We have integrated into VPR-Bench, a variety of different metrics that evaluate a VPR technique on the fronts of matching performance, computational needs and storage requirements.

We have collated the taxonomy of various metrics used in VPR by both the computer vision and the robotics communities in Table \ref{VPR Evaluation Metrics} for the reader's reference, which are also discussed later in the paper. The primary usage and audience of the techniques do not represent the limitations of the respective metrics to particular use-cases/communities, but instead identify the best/most-suitable use-cases for the respective metric. We have broadly classified the usage into 3 areas: primary-localisation, loop-closure and image-retrieval. Each of these classes can then contain various applications, e.g. image-retrieval (which intends to retrieve as many correct matches for a query as possible from the database) could be used for query-expansion, structure-from-motion (3D-model creation), content-based search engines and many others. Primary-localisation (a vision-only localisation system that uses VPR for position estimates) and loop-closure (error drift correction in a SLAM pipeline) do not require the retrieval of all the existing matches of a query from the database, but instead a single (or few) correct match(es) to have a location estimate at a high frame-rate. A primary-localisation system may or may not have a false-positive rejection scheme within its localisation pipeline and therefore the respective application and the suited metric would change accordingly. Loop-closure represents an important VPR application within a visual-SLAM system. Because, the objective of having loop-closure is to correct the existing uncertainty of the visual-SLAM system, it is usually preferred that a highly precise VPR technique be used for loop-closure. The \textit{kidnapped robot problem} can also be considered as a particular case of loop-closure. In the following, we discuss each of the metrics that have been used for evaluations in this work, their motivation and limitations.

\begin{table*}
\centering
\caption{A taxonomy of VPR evaluation metrics is given here.
Where PL: Primary Localisation, LC: Loop-closure, IR: Image Retrieval, FP: False-Positives, RC: Robotics Community, CV: Computer Vision Community, MB: Matching-based and CB: Computational-intensity-based. * identifies a sub-class of PL and LC, where the underlying system is not robust to false-positives,. This robustness normally arises from geometric-verification, visual-inertial odometry, re-ranking schemes, false-positive predictors, weak-prior and/or other similar modules.}
\label{VPR Evaluation Metrics}       % Give a unique label
\begin{tabular}{cccccc}
\hline\noalign{\smallskip}
Metric & Primary Usage & Output & FP Allowed? & Primary Audience & Nature\\
\noalign{\smallskip}\hline\noalign{\smallskip}
AUC-PR & PL+LC+IR & Single-value & Yes & RC+CV & MB\\
Extended Precision & PL*+LC* & Single-value & No & RC & MB\\
Recall@100\%Precision  & PL*+LC* & Single-value & No & RC & MB\\
RecallRate@N & PL+LC+IR & N-values & Yes & RC+CV & MB\\
Recall@ReducedPrecision  & PL+LC+IR & Single-value & Yes & RC+CV & MB\\
mean-Average-Precision & IR & Single-value & Yes & CV & MB\\
F1-Score  & PL+LC & Multiple-values & Yes & RC+CV & MB\\
Encoding Time  & PL+LC & Single-value & Yes & RC & CB\\
Matching Time  & PL+LC+IR & Single-value & Yes & RC+CV & CB\\
PCU & PL+LC & Single-value & Yes & RC & MB+CB \\
RMF  & PL+LC & Single/Multiple values & Yes & RC & MB+CB\\
\noalign{\smallskip}\hline
\end{tabular}
\end{table*}

\subsubsection{AUC and PR-Curves}
\label{AUC-PR_subsection}
\textbf{Motivation:} AUC-PR is one of the most used evaluation metrics in the robotics VPR community. It presents a good overview of the precision and recall performance of a VPR technique, where only a single correct match, which should be the best matched reference image, is required for a given query image. Therefore, it is usually suitable for applications that require high precision, high recall, single correct match, and that only consider the best matched image for their operation, e.g. loop-closure and topological-localisation.

\textbf{Limitations:} AUC-PR may not be relevant for applications that intend to retrieve as many correct ground-truth matches as possible from the reference database. It is not affected if the second-best (or third-best and so on) match is actually a correctly retrieved image. Thus, it has two major limitations: in cases where many correct ground-truth matches exist in the database and the system application (3D-modelling, constraint-creation) requires the correct retrieval of all of these images, AUC-PR may not present significant value, as it only considers a single retrieved image per query in its computations. Secondly, AUC-PR may not be relevant in cases where false-positive rejection is possible (e.g. weak GPS prior, geometric verification, robust optimization back-ends) and the VPR system is mainly used to retrieve a correct match within a list of top matching candidates.

\textbf{Metric Design:} AUC-PR is computed from Precision-Recall curves which are aimed at understanding the loss of precision with increasing recall at different confidence score thresholds. Generally, in VPR the image similarity scores are considered as confidence scores and are varied within the maximum range to plot PR-curves. Precision and Recall are computed for each threshold in a range of thresholds as
\begin{equation}
Precision= \frac{True \; Positives}{True \; Positives + False \; Positives},
\end{equation}
\begin{equation}
Recall= \frac{True \; Positives}{True \; Positives + False \; Negatives}.
\end{equation}

Where in terms of VPR, given a query image and a chosen confidence score threshold, a True-Positive (TP) represents a correctly retrieved image of a place based on ground-truth information. A False-Positive (FP) represents an incorrectly retrieved image based on ground-truth information. A False-Negative (FN) is a correctly retrieved image based on ground-truth, the matching score for which is lower than the chosen confidence score threshold. Please note that in most VPR datasets, all correctly matched images that are rejected due to the matching scores being lower than the chosen threshold are classified as false-negatives, because ground-truth matches exist for all images in the datasets. There are no True-Negatives (TN) usually in the datasets, i.e. query images that do not have a correct match in the reference database (we also discuss this later in the paper for ROC curves). By selecting different values of the matching threshold, varying between the highest matching score and the lowest matching score, different values of Precision and Recall can be computed. The Precision values are plotted against the Recall, and the area under this curve is computed, which is termed AUC-PR. The ideal value of AUC-PR is 1 and Precision=1 for all recall values represents an ideal PR-curve.

\subsubsection{RecallRate@N}
\textbf{Motivation:} One of the most commonly used evaluation metrics from the computer vision VPR community is RecallRate@N (also termed as Recall@N). This metric tries to model the fact that a correctly retrieved reference image (as per the ground-truth) does not necessarily has to be the top-most retrieved image, but only needs to be among the Top-N retrieved images. The primary motivation behind this is that subsequent filtering steps, e.g. geometric consistency or weak GPS-prior, can be used to re-arrange the ranking of the retrieved images and avoid false-positives. As this provision is not modelled by AUC-PR and presents an important case study, we have included this metric into our framework.

\textbf{Limitations:} There may be cases where false-positive rejection is not possible, e.g. geometric-verification may fail in dark, unstructured environments and in extreme conditions (rain, fog etc) and therefore in such cases it may be relevant to use VPR systems (and metrics like AUC-PR) that are highly precise and where the best matched image should not be a false-positive. On the other hand, similar to AUC-PR, RecallRate@N also rewards a VPR system only for retrieving a single correct match per query from the reference database. Both the metrics neither penalize nor reward retrieval of more than one correct match per query, which is a particular use-case for the mean-Average-Precision (mAP) metric.

\textbf{Metric Design:} The requirement for RecallRate@N is that the correct reference image for a query only needs to be among the Top-N retrieved images. Let the total number of query images with a correct match among the Top-N retrieved images be $M_Q$, and the total number of query images be $N_Q$, then the RecallRate@N can be computed as

\begin{equation} \label{eq:RAtN}
RecallRate@N= \frac{M_Q}{N_Q}.
\end{equation}

Please note that RecallRate@1 is actually equal to the Precision at maximum Recall $P_{Rmax}$. The ideal value of RecallRate@N is equal to 1. RecallRate@N does not consider false-negatives (incorrectly discarded correct matches) and true-negatives (new places) and is therefore not a replacement for AUC-PR and AUC-ROC, respectively. An ideal RecallRate@N graph should represent a straight line on y-axis=1 (RecallRate=1) for all values of N on the x-axis.

\subsubsection{ROC Curves}
\textbf{Motivation:} AUC-PR and RecallRate@N do not consider true-negatives within them. In VPR, true-negatives are those query images for which the ground-truth correct reference match does not exist. These true-negatives can also be thought of as `new places', i.e. places which haven't been seen before by the vision system. It is important for a VPR system to identify these true-negatives for their usage within a topological SLAM system for an exploration task. Previous metrics like AUC-PR and RecallRate@N are designed for tasks where a map is already available and the primary task of the VPR system is only accurate localisation. AUC-ROC therefore complements the analysis provided by AUC-PR and/or RecallRate, but does not replace them.

\textbf{Limitations:} ROC curves are useful for balanced class problems and therefore in datasets where true-negatives and true-positives are not balanced, ROC curves may not present value. ROC curves are also not useful for applications that already have a fixed map of the environment available, because in this case identification of new places is not a requirement.

\textbf{Metric Design:} In order to assess the true-negative classification performance of a VPR system, we utilise the well-established Receiver-Operating-Characteristic (ROC) curve. Because VPR datasets in general do not contain any true-negatives, they represent an imbalanced class problem, i.e. true-positives and true-negatives classes are not balanced. This is another reason due to which ROC curves have not been used for VPR evaluation, as the focus has always been on achieving very high-precision, i.e. retrieving as many correct place matches as possible. We therefore manually add true-negatives to the Gardens Point dataset for our ROC evaluation, where true-negatives are images taken from the Nordland dataset as a case-study. The modified Gardens Point dataset contain the 200 original true-positives and the added 200 true-negatives from Nordland dataset. The reference database remains the same, while the ground-truth is modified such that for the 200 true-negative query images, it identifies that a correct match does not exist. This modified dataset and associated ground-truth is available separately in our framework to avoid confusion with the original datasets. It is easily possible to extend this analysis on other datasets and is supported by our framework.

The definitions of true-positives, false-positives and false-negatives for ROC curves remain the same as PR curves, with only the extra addition of true-negatives as defined above. An ROC curve is a plot between the true-positive rate (TPR) on the vertical axis and the false-positive rate (FPR) on the horizontal axis.
The TPR signifies how many of the total query images that have a correct reference match have been retrieved by a VPR technique. The FPR identifies how many of the total query images that do not have a correct reference match were labelled as false-positives. These metrics are computed as
\begin{equation}
TPR= \frac{True \; Positives}{True \; Positives + False \; Negatives},
\end{equation}
\begin{equation}
FPR= \frac{False \; Positives}{False \; Positives + True \; Negatives}.
\end{equation}

Similar to PR-curves, the true-positive rate and the false-positive rate are computed for a range of different matching confidence thresholds. Area under this ROC curve (AUC-ROC) is used to model the classification quality of a VPR technique. A perfect AUC-ROC is equal to 1 and an ideal ROC curve is identified by TPR=1 for all values of FPR. An AUC-ROC of 0.5 identifies that a technique has no separation capacity between the true-class (queries with existing matches in reference database) and the false-class (new places). An AUC-ROC below 0.5 means that a technique is yielding opposite labels for most of the candidates, i.e. true-positives are classified as true-negatives and vice-versa.

\subsubsection{Image Retrieval Time}
\textbf{Motivation:} From a computational perspective, the most important factors to consider are the feature encoding time and the descriptor matching time of VPR techniques, which have been usually reported by works from both the VPR communities. These computational metrics only complement the metrics related to place matching precision. In applications where the reference database is significantly large\footnote{The quantified meaning of `large' is usually dependent upon the computational platform, system's implementation and the ratio of feature encoding time to descriptor matching time.}, descriptor matching time may be more relevant than feature encoding time and vice-versa.

\textbf{Limitations}: Unlike other precision-related metrics, computational performance is greatly dependent on the underlying platform and can change significantly from one system to another.

\textbf{Metric Design:} Feature encoding time and descriptor matching time can be combined together to model the image retrieval time of a given VPR technique. Let the total number of images in the map (reference database) be $Z$. Let $t_e$ represent the feature encoding time and $t_m$ represents the time required to match feature descriptors of two images. Also, let the retrieval-time of a VPR technique be denoted as $t_R$, where this $t_R$ represents the time taken (in seconds) by a VPR technique to encode an input query image and match it with the total number of images ($Z$) in the reference map to output a potential place matching candidate. We model this $t_R$ as
\begin{equation} \label{t_R}
t_R=t_e+O(Z) \times t_m .
\end{equation}
Here $O(Z)$ represents the complexity of search mechanism for image matching and could be linear, logarithmic or other depending upon the employed neighbourhood selection mechanism (e.g., linear search, nearest-neighbour search, approximate nearest neighbour search etc.). While implementing this framework, we ensured that $t_e$ and $t_m$ are computed in a fashion where all subsequent dependencies, input/output data transfer, pre-processing and preparations of a VPR technique are included in these timings for a fair comparison. The descriptor matching time is related to the descriptor size, computational platform, descriptor dimensions and descriptor data-type, which have all been reported in this work for completeness. 

Additional to the metrics discussed previously, we also compute and report the feature descriptor size of all VPR techniques to reflect the storage requirements, which are highly relevant for large-scale maps.  

\subsubsection{True-Positives Distribution Analysis}
\label{True_Positives_Distribution}
\textbf{Motivation} Some robotics applications may require that a loop-closure candidate (a correct VPR match) must be obtained at least every Y meters over a traversed trajectory. For a robot localisation system (visual-inertial-based, visual-SLAM-based, dead-reckoning-based and similar), a VPR technique that is moderately precise but has a uniform true-positive distribution over the robot's trajectory has more value than a highly-precise technique with a non-uniform distribution.. We have therefore included true-positives distribution over trajectory analysis in our benchmark.

\textbf{Limitations}: This metric is application-specific and does not provide insights for the non-traversal datasets usually employed by the computer vision VPR community.

\textbf{Metric Design:} This metric was presented by \cite{porav2018adversarial}. They analyse the distribution of loop-closure candidates (true-positives) by creating histograms identifying inter-loop-closure distances, such that the height of the histogram bar specifies the number of loop-closures performed in the dataset with that particular inter-frame distance constraint. We use the same analysis schema in this work.

\subsubsection{Other VPR Metrics}
The metrics discussed previously in this paper have their specific utilities, and in some cases these metrics complement each other (e.g. AUC-PR and RecallRate@N), and in other cases provide dedicated value (e.g. AUC-ROC for true-negatives, retrieval time for computational analysis). Still, even more metrics have  been used for VPR, including mAP (\cite{revaud2019learning}), Performance-per-Compute-Unit (\cite{zaffar2020cohog}, \cite{tomitua2020convsequential}), Recall@0.95 Precision (\cite{chen2011city}, Extended Precision (\cite{ferrarini2020exploring}), F1-score (\cite{hausler2019multi}), error-rate (\cite{chen2014multi}) and Recall@100\% Precision (\cite{chen2014convolutional}). To limit the scope of the analysis performed in this paper, and because there is a high corelation between some of these metrics (e.g. between RecallRate@N, Recall@100\% Precision and Recall@95\% Precision), we have implemented many of these other metrics in the implementation of VPR-Bench, but did not include them in this paper.

\subsection{Invariance Quantification Setup}
\label{Invariance_Quantification_Setup}
In this sub-section (and its respective results/analysis in sub-section \ref{Invariance Analysis}) we propose a thorough sweep over a wide range of quantified viewpoint and illumination variations and study the effect on VPR techniques.

\cite{aanaes2012interesting} proposed a well-designed and highly-detailed dataset, namely Point Features dataset, where a synthetically-created scene is captured from $119$ different viewpoints, under $19$ different illumination conditions. While the original dataset consists of different synthetic scenes, some of which are irrelevant to VPR, we utilise a subset of the dataset that represents scenes of synthetically-created `Places', and we use 2 of these scenes/places in our work. We have integrated this subset of the Point Features dataset in our framework and sub-section \ref{Point_Features_Dataset} is dedicated to explaining the details of this dataset.

An obvious limitation of the Point Features dataset is that it depicts synthetic scenes (toy-houses, toy-cars etc) instead of a real-world scene. This limitation is a challenge to address, because in real-world scenes it is significantly difficult to control the illumination of a scene. However, we do make an effort in this paper to present the analysis of viewpoint and illumination variation effects on VPR performance for real-world variation-quantified (semi-quantified) datasets as well. The level of quantification available in these datasets is not as detailed as the Point Features dataset, but they serve to bridge the sim-to-real gap in our evaluation to some degree. Therefore, in this reference, we have used the QUT multi-lane dataset (\cite{skinner2016high}) for viewpoint variations and the MIT multi-illumination dataset (\cite{murmann19}) for illumination variations. Details of both of these datasets are available in their respective sub-sections below.

We have also dedicated a sub-section (sub-section \ref{Evaluation_Mechanism}) to present the details of our evaluation mechanism on these 3 datasets. The evaluation mechanism in this paper (and in the proposed framework) is kept the same for all 3 datasets (Point-features, QUT multi-lane, MIT multi-illumination datasets) to ensure consistency. Please note that throughout this section the term `same-but-varied place' refers to the images of a place from different viewpoints or under different illumination conditions, while the term `different place' refers to a place that is geographically not the same as the `same-but-varied' place. For each of the 3 datasets in this section, there are only 2 actual places in total, i.e. `the same-but-varied' place and the `different place'.

\subsubsection{Point Features Dataset}
\label{Point_Features_Dataset}
\begin{figure}[t]
\begin{center}
%\fbox{\rule{0pt}{2in} \rule{0.9\linewidth}{0pt}
\includegraphics[width=0.8\linewidth]{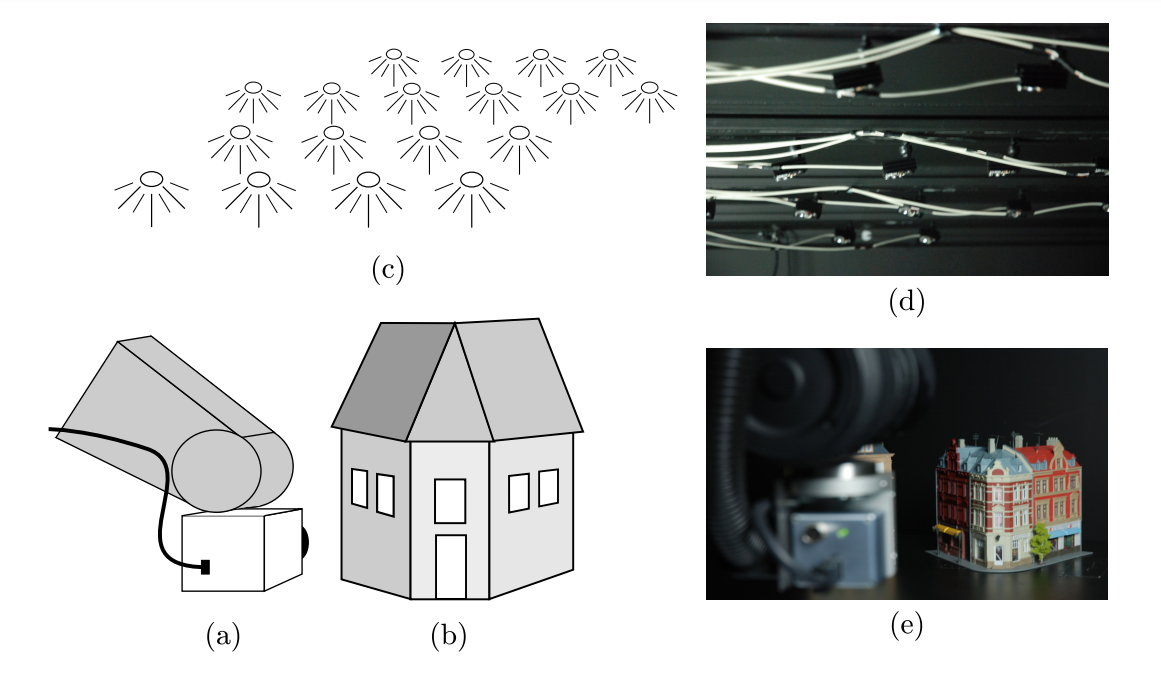}
\end{center}
\caption{The schematic setup of the Point Features dataset has been reproduced here with permission from \cite{aanaes2012interesting}. The dataset primarily consists of (a) A camera mounted on a robot-arm, (b) Synthetic Scene, (c) LED arrays for illumination, (d) (e) Snapshots of the actual setup.}
\label{Pointfeaturedataset_setup}
\vspace{-0mm}
\end{figure}

\begin{figure}[t]
\begin{center}
%\fbox{\rule{0pt}{2in} \rule{0.9\linewidth}{0pt}
\includegraphics[width=1\linewidth]{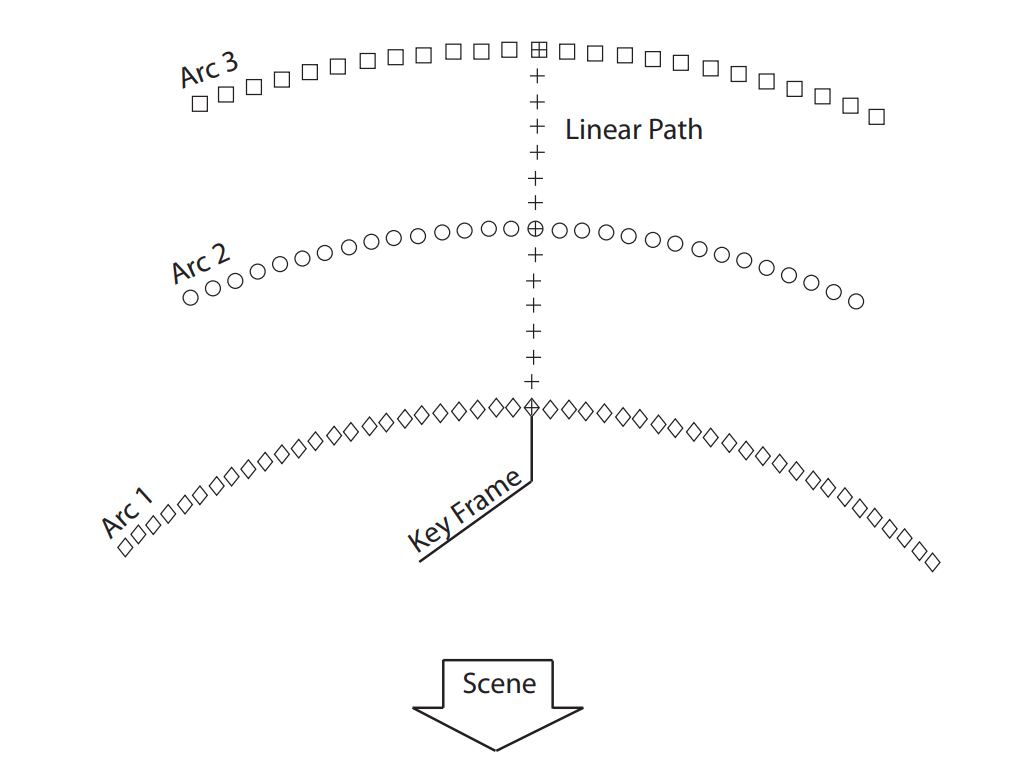}
\end{center}
\caption{The 119 different viewpoints in the Point Features dataset have been reproduced here with permission from \cite{aanaes2012interesting}. Camera is directed towards the scene from all viewpoints. Arc 1, 2 and 3 span 40, 25 and 20 degrees, respectively, while the radii are 0.5, 0.65 and 0.8 meters.}
\label{viewpoint_variation_pointfeaturedataset}
\vspace{-0mm}
\end{figure}

\begin{figure}[t]
\begin{center}
%\fbox{\rule{0pt}{2in} \rule{0.9\linewidth}{0pt}
\includegraphics[width=1\linewidth]{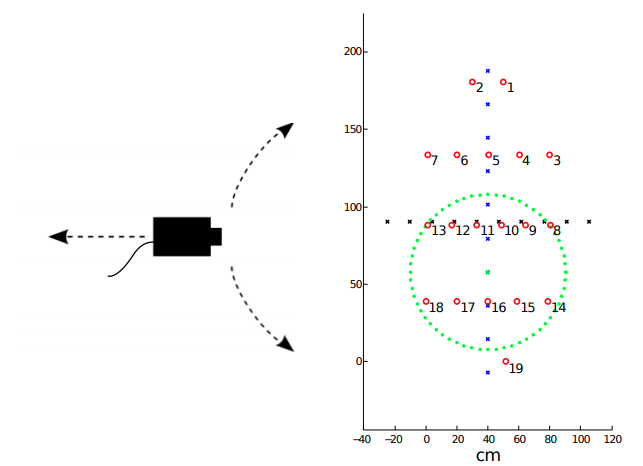}
\end{center}
\caption{The distribution of LEDs across physical space is shown as seen from above. Each red circle represents an LED and only a single LED is illuminated at a point in time, yielding $19$ different illumination conditions. In the original work, \cite{aanaes2012interesting}, used artificial linear relighting from left-to-right (blue) and front-to-back (black) based on a Gaussian-weighting, as depicted with the green-circle, but in our work we have only used the original $19$ single-LED illuminated cases. These $19$ cases (red-circles) need to be seen in correspondence with Table \ref{azimuthandelevationangles}.}
\label{illumination_variation_pointfeaturesdataset}
\vspace{-0mm}
\end{figure}

\begin{figure*}[t]
\begin{center}
%\fbox{\rule{0pt}{2in} \rule{0.9\linewidth}{0pt}
\includegraphics[width=0.8\linewidth]{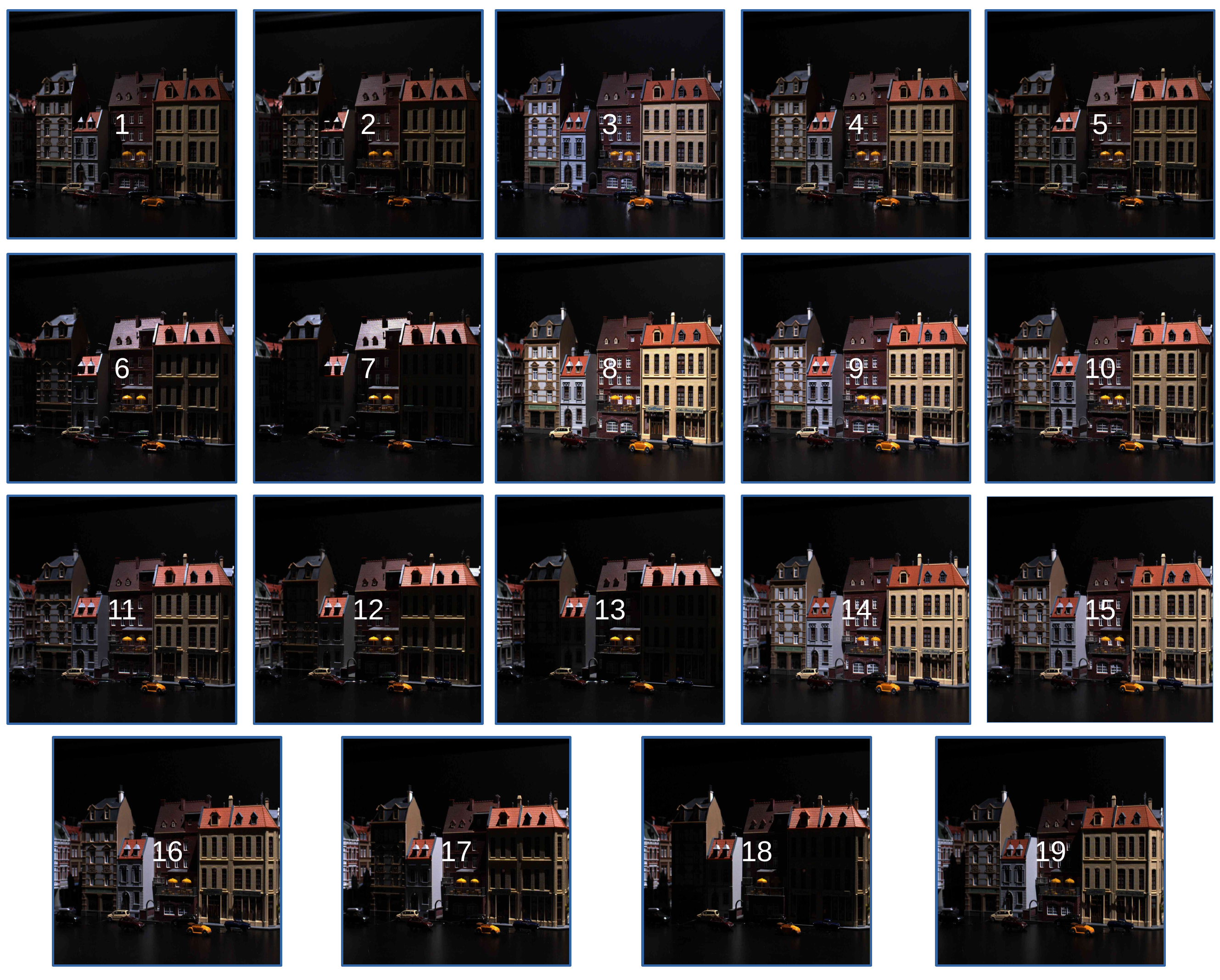}
\end{center}
\caption{The change in appearance of a scene for $19$ different illumination levels is shown here from the Point Features dataset.}
\label{illumination_variation_pointfeaturesdataset_qualitative}
\vspace{-0mm}
\end{figure*}

The Point Features dataset can be broadly classified to have $3$ variations: 1) Viewpoint, 2) Illumination and 3) Scene. We fully use the former two variations in our work, while only two relevant scenes (representing two different places) are utilised from the latter. The authors (\cite{aanaes2012interesting}) achieve viewpoint-variation by mounting the scene facing camera on a highly-precise robot arm, where this robot arm is configured to move across and in-between $3$ different arcs, that amount to a total of $119$ different viewpoints, as depicted in Fig. \ref{viewpoint_variation_pointfeaturedataset}. Their setup used $19$ LEDs that varied from left-to-right and front-to-back to depict a varying directional light source. This directional illumination setup has been reproduced in Fig. \ref{illumination_variation_pointfeaturesdataset}, while the azimuth ($\phi$) and elevation angle ($\theta$) of each LED is listed in Table \ref{azimuthandelevationangles}. Fig. \ref{Pointfeaturedataset_setup} shows various components of the dataset, while in Fig. \ref{illumination_variation_pointfeaturesdataset_qualitative} we qualitatively show all the $19$ different illumination cases on one of the scenes. 

\begin{table}
\caption{The azimuth ($\phi$) and elevation angles ($\theta$) of each LED are listed here (in degrees) with respect to the physical table surface that acts as the center of coordinate system.}
\label{azimuthandelevationangles}       % Give a unique label
\begin{tabular}{cccccc}
\hline\noalign{\smallskip}
LED Number & $\theta$ & $\phi$ & LED Number & $\theta$ & $\phi$  \\
\noalign{\smallskip}\hline\noalign{\smallskip}
1 & 264 & 57 & 11 & 28 & 86 \\
2 & 277 & 57 & 12 & 10 & 80 \\
3 & 227 & 68 & 13 & 6 & 74 \\
4 & 245 & 72 & 14 & 125 & 65 \\
5 & 270 & 73 & 15 & 109 & 68 \\
6 & 297 & 72 & 16 & 89 & 69 \\
7 & 314 & 68 & 17 & 69 & 68 \\
8 & 174 & 74 & 18 & 53 & 64 \\
9 & 170 & 80 & 19 & 97 & 56 \\
10 & 152 & 86  \\
\noalign{\smallskip}\hline
\end{tabular}
\end{table}

\subsubsection{QUT Multi-lane Dataset}
\label{QUT_Multilane_Dataset}
The QUT multi-lane dataset is a small-scale dataset depicting a traversal through an outdoor environment (\cite{skinner2016high}) performed at 5 different laterally-shifted viewpoints under similar illumination and seasonal conditions. This traversal has been performed at a near-constant velocity by a human from an ego-centric viewpoint. The dataset contains 2 types of viewpoint changes: (a) Forward and Backward movement, i.e. Zoom-in and Zoom-out effect similar to the inter-arc viewpoint change of the Point Features dataset, (b) Lateral viewpoint change, which is close to the viewpoint change across the arcs of the Point Features dataset.

\begin{figure*}[t]
\begin{center}
%\fbox{\rule{0pt}{2in} \rule{0.9\linewidth}{0pt}
\includegraphics[width=1\linewidth]{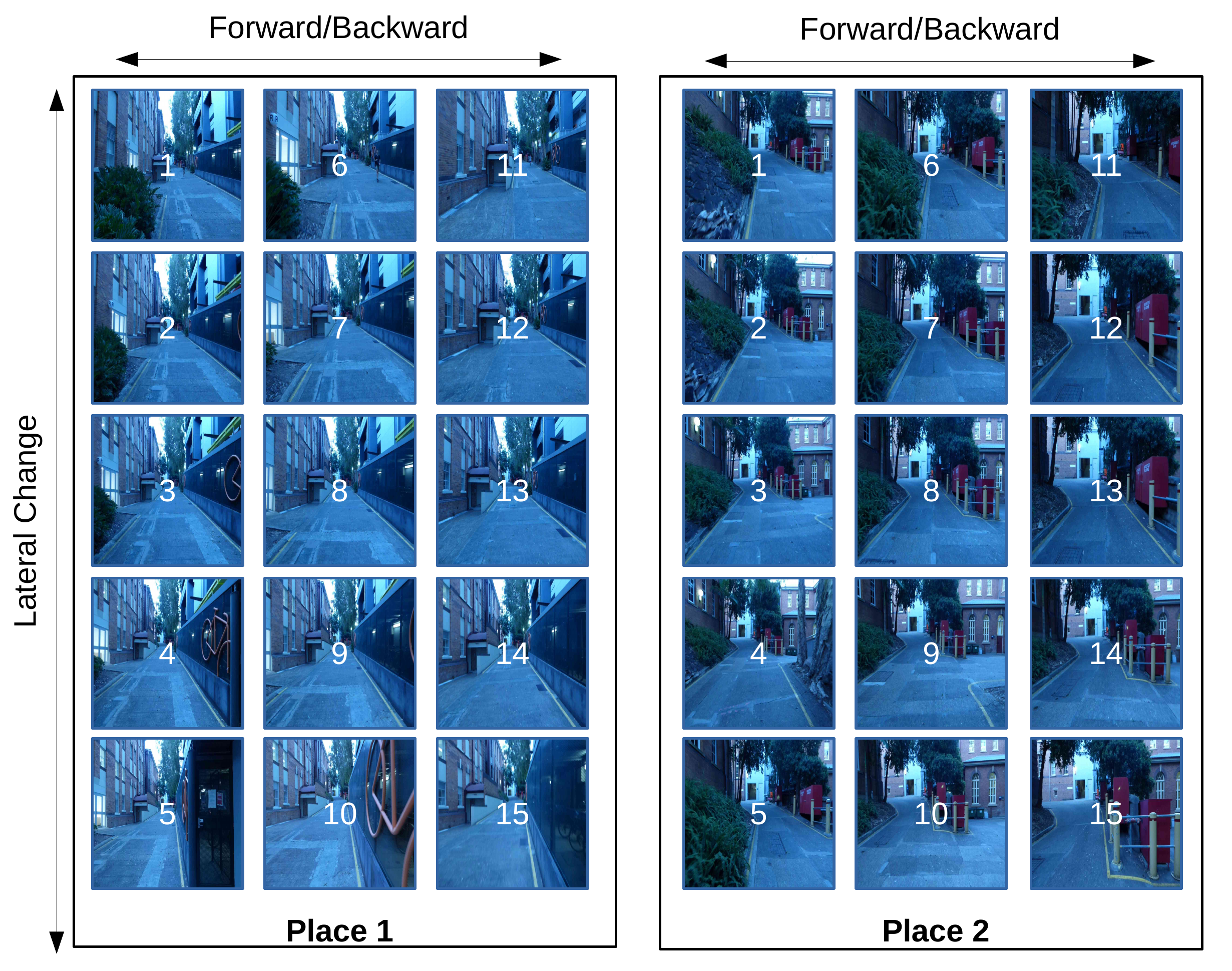}
\end{center}
\caption{The 15 different viewpoint cases in the QUT multi-lane dataset for both the scenes/places have been presented here.}
\label{QUT_multilane_dataset_exemplar}
\vspace{-0mm}
\end{figure*}

We use in total 2 different scenes (representing 2 different places) from their traversal and for each scene use $15$ viewpoints. These $15$ viewpoints represent $5$ lateral viewpoint changes for $3$ consecutive (forward/backward movement) viewpoints of each scene/place. The lateral viewpoint change is almost $1.2$ meters, while the forward/backward viewpoint change is around $3.5$ meters. Examples of these viewpoint changes have been shown in Fig. \ref{QUT_multilane_dataset_exemplar} for both the scenes/places.

\subsubsection{MIT Multi-illumination Dataset}
\label{MIT_Multiillumination_Dataset}
The MIT multi-illumination dataset was recently proposed by \cite{murmann19}. This dataset represents a variety of indoor scenes captured under 25 different illumination conditions. Most of the scenes represented in this dataset may not actually be classified as `Places', however because we only require 2 scenes/places, we have manually mined scenes that represent an indoor appearance of a place and are feature-full.
\footnote{The authors acknowledge that even the multi-illumination dataset may not fully represent a real-world `landmark' and multiple illumination sources etc, however to the best of authors' knowledge, this is the most relevant real-world illumination quantified dataset for the problem at hand. Controlled illumination, especially in outdoor scenes is notoriously difficult as identified by \cite{murmann19}.}

\begin{figure*}[t]
\begin{center}
%\fbox{\rule{0pt}{2in} \rule{0.9\linewidth}{0pt}
\includegraphics[width=1\linewidth]{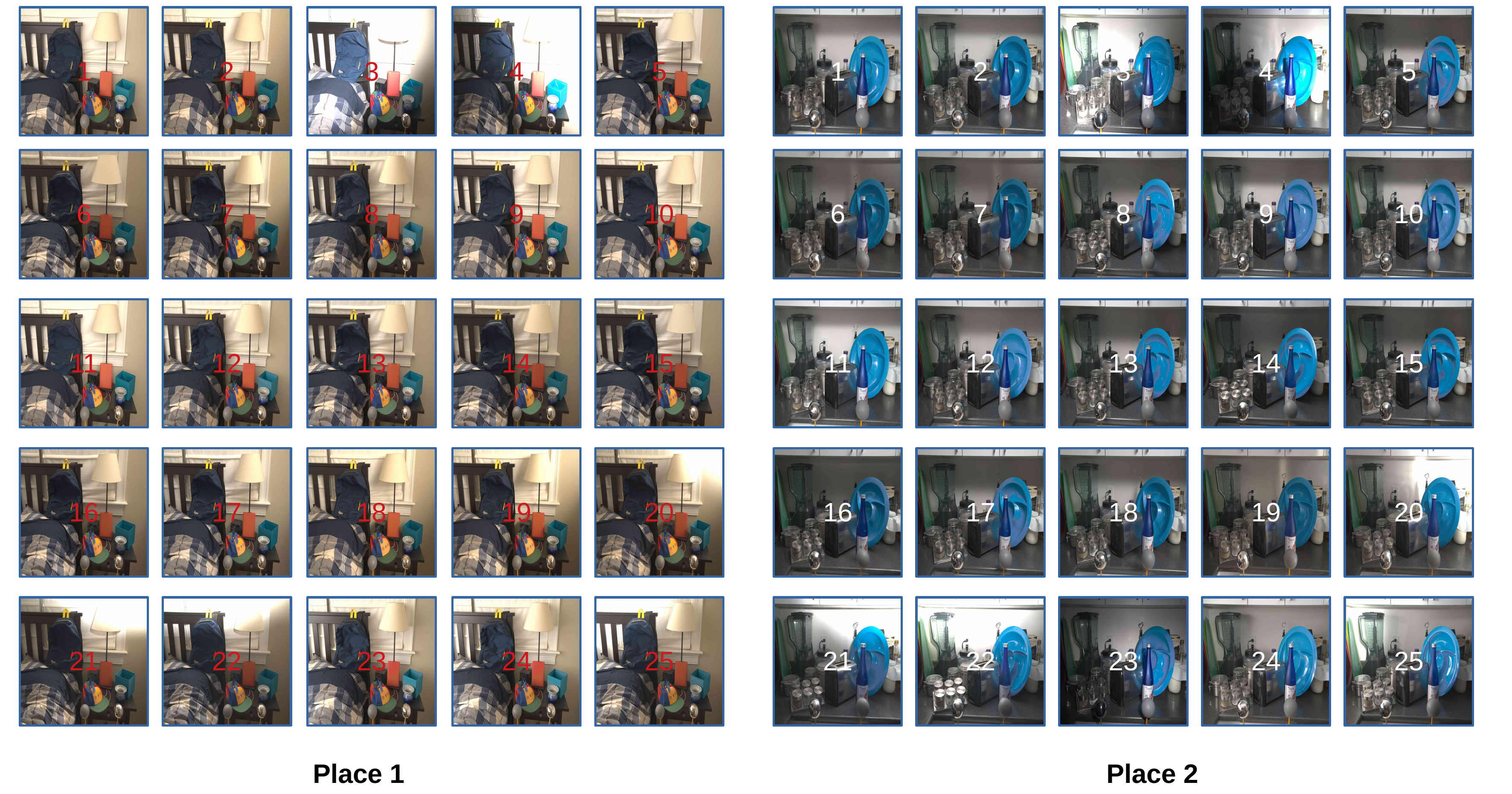}
\end{center}
\caption{The 25 different illumination cases for both the scenes/places from the MIT multi-illumination dataset have been presented here.}
\label{MIT_multillumiation_dataset_exemplar}
\vspace{-0mm}
\end{figure*}

The dataset consists of a total of 1016 interior scenes, each photographed under 25 predetermined lighting directions, sampled over the upper hemisphere relative to the camera. All of these
scenes depict common domestic and office environments. The scenes are also populated with various objects, some of which represent shiny surfaces and are therefore interesting for our evaluation.
The lighting variations are achieved by directing a concentrated flash beam towards the walls and ceiling of the room, which is similar to the works of \cite{mohan2007tabletop} and \cite{murmann2016computational}. The bright spot of light that bounces off the wall becomes a virtual light source that is the dominant source of illumination for the scene in front of the camera. The approximate position of the bounce light is controlled by rotating the flash head over a standardized set of directions. The authors propose that their camera and flash system is more portable than dedicated light sources, which simplifies its deployment `in the wild'.
Because the precise intensity, sharpness and direction of the illumination resulting from the bounced flash depends on the room geometry and its materials, these lighting conditions have been recorded by inserting a pair of light probes, a reflective chrome sphere and a plastic gray sphere, at the bottom edge of every image. For further specification details, we would refer the reader to the original paper of \cite{murmann19} for avoiding textual redundancies. Examples of the 2 different places under the varying illumination conditions have been shown in Fig. \ref{MIT_multillumiation_dataset_exemplar}, where Place 1 is chosen due to its closest-possible depiction of an indoor VPR-relevant scene, while Place 2 is chosen due to the shiny objects in that scene. Both the scenes/places are feature-full.

\subsubsection{Evaluation Mechanism}
\label{Evaluation_Mechanism}
In order to utilise the densely-sampled viewpoint and illumination conditions in the Point Feature dataset (and the less-detailed QUT multi-lane dataset and the MIT multi-illumination dataset), we had to devise an analysis scheme where VPR performance variation could be quantified and analysed. This quantification is not possible with the traditional place matching evaluation, where there are only two possible outcomes for a given query image, i.e. a correct match or a false match. This is because the mismatch cannot be guaranteed to have resulted from that particular variation and may have resulted from perceptual-aliasing or a smaller map-size. Also, even if an image is matched, it is not guaranteed that increasing the map-size (i.e. the no. of reference images) would not affect the outcome, as the greater the no. of reference images, the greater the chances of mismatch. However, each VPR technique does yield a confidence-score for the similarity of two images/places. Ideally, if two images represent the same place, then the confidence-score should remain the same, if one of the image of that place is varied with respect to viewpoint or illumination, while keeping the other constant. However, in practical cases, VPR techniques are not fully-immune to such variations and a useful analysis would be to see this effect on the confidence-score.

Therefore, our analysis on the 3 datasets in this section and the VPR-Bench framework are developed based on the effect of viewpoint- and illumination-variation on the confidence score. This confidence score usually refers to the matching score (L1-matching, L2-matching, cosine-matching etc.) in VPR research and for two exactly similar images (i.e. two copies of an image), this confidence/matching score is always equal to $1$. However, when the image of the same place/scene is varied with respect to viewpoint or illumination, the confidence score decreases. This decrease in matching score by varying images of the same place/scene along the pre-known, numerically-quantified viewpoint- and illumination-levels of the 3 datasets presents analytically and visually the limits of invariance of a VPR technique. However, the trends of these variations in-between different VPR techniques cannot be compared solely based on the decrease of confidence scores, due to different matching methodologies. Therefore, for each VPR technique, we draw the confidence score variation trend for the same place along with the trend for a different place/scene. The point at which the matching score for the same place (but viewpoint or illumination varied) approaches near (or below) the matching score for a different place, identifies the numeric value of viewpoint/illumination change that a VPR technique cannot prospectively handle.

\textbf{Evaluation Mechanism Point Features Dataset:}
There are a total of $119$ different viewpoint positions and $19$ different illumination levels. We consider the illumination case 1 in Fig. \ref{illumination_variation_pointfeaturesdataset} and the left-most point on Arc 1 of Fig. \ref{viewpoint_variation_pointfeaturedataset} as our keyframe(s) for viewpoint- and illumination-invariance analysis, respectively. The $119$ viewpoint positions are numerically labelled in consecutive ascending order from the keyframe (labelled as `1') to the right-most point on Arc 1, followed by the leftmost point on Arc 2 to the rightmost point on Arc 2, which is then followed by the left-most point on Arc 3 and the last (labelled as `119') position is the right-most point on Arc 3. For each analysis and each VPR technique, the key-frame is matched with itself to provide an ideal matching score, i.e. $1$. For viewpoint-variation analysis, we keep the illumination type/level constant, move along Arc 1 in a clock-wise fashion and compute the matching scores between the keyframe and the viewpoint-varied (quantified) images. The same is repeated for Arcs 2 and 3, where the keyframe remains the same i.e. the left-most point on Arc 1. The matching scheme yields a total of $119$ different matching scores for each of the $119$ different viewpoint positions.

For the illumination invariance analysis, the $19$ illumination cases are identified numerically in Table \ref{azimuthandelevationangles} and qualitatively in Fig. \ref{illumination_variation_pointfeaturesdataset_qualitative}. For the illumination-invariance analysis, the viewpoint position is kept constant (left-most point on Arc 1) and the illumination levels are varied.

Because the decline in matching score itself does not provide too much insight, we draw the matching scores for the same-but-varied scene in the Point Features dataset, along with the matching scores when the reference scene is a different place (i.e. the query/keypoint frame and reference frame are different places). For computing the matching scores between the keyframe and the different scene/place, we utilise all of the $119$ viewpoint positions and the $19$ illumination levels of the different scene/place. This gives us the corresponding number (119/19 for both variations) of data-points for the confidence scores between keyframe and the different place to be drawn against the data-points for the same-but-varied place. There are further advantages to using all the (119 and 19) viewpoint and illumination cases for the different place, as explained later in sub-section \ref{Invariance Analysis}.

\textbf{Evaluation Mechanism QUT Multi-lane Dataset:}
The evaluation mechanism is the same for QUT Multi-lane Dataset as that for the Point Features dataset. In this case, however, there are a total of $15$ different viewpoint positions for the same-but-varied place and $15$ different viewpoint positions for the different place. Unlike the large number of viewpoint variations in the Point Features dataset which were difficult to qualitatively represent, the $15$ different viewpoint positions for both the scenes/places for the QUT multi-lane dataset have been shown and labelled in Fig. \ref{QUT_multilane_dataset_exemplar}. For both the scenes/places, the viewpoint positions 1-5 are left-to-right variations at the beginning of the traversal, 6-10 are left-to-right variations a few meters ahead of 1-5, and 11-15 are left-to-right variations a few meters ahead of 6-10. Image 1 of Place 1 serves as the keyframe. The matching scores between the keyframe and the same-but-varied place, and between the keyframe and the $15$ viewpoints of different place (place 2) are computed/utilised in the same fashion as that for Point Features dataset. 

\textbf{Evaluation Mechanism MIT Multi-illumination Dataset:}
The evaluation mechanism for the MIT multi-illumination dataset is also the same as that of the Point Features dataset. In this case, however, there are a total of $25$ different illumination cases. These illumination cases for both the scenes have been identified in Fig. \ref{MIT_multillumiation_dataset_exemplar}. Image 1 of Place 1 serves as the keyframe. The matching scores between the keyframe and the same-but-illumination-varied place, and between the keyframe and the $25$ different illuminations of different place (place 2) are computed/utilised in the same fashion as that for the Point Features dataset.

\section{Results and Analysis}
\label{resultsandanalysis}
In this section, we present detailed results and analysis for the 10 VPR techniques on the 12 datasets for various evaluation metrics. We discuss the variation in performance by varying dataset ground-truths, computational platforms (CPU vs GPU), feature descriptor sizes and the retrieval timings vs platform speed. We provide an extensive analysis based on our viewpoint and illumination invariance quantification setup. Finally, we discuss the role of viewpoint variance vs invariance and the subjective requirements of these from a VPR system. The experiments were performed on a Ubuntu 20.04.1 LTS operating system running on an AMD(R) Ryzen(TM) 7-3700U CPU @ 2.30GHz.
\subsection{Place Matching Performance} \label{VPR Performance Evaluation}

\begin{figure*}[t]
\begin{center}
%\fbox{\rule{0pt}{2in} \rule{0.9\linewidth}{0pt}
\includegraphics[width=1.0\linewidth]{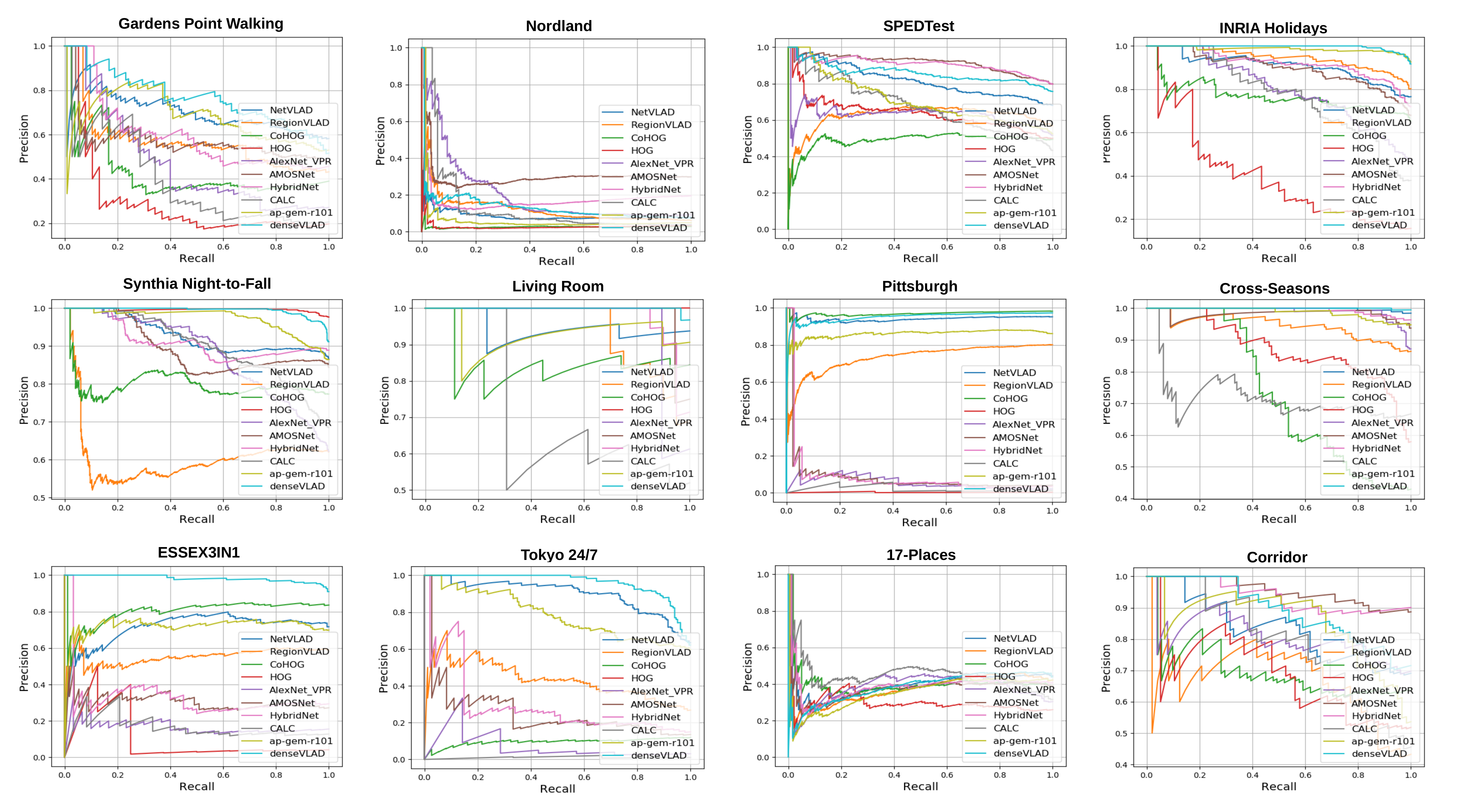}
\end{center}
\caption{The Precision-Recall curves for all $10$ VPR techniques generated on the $12$ datasets by VPR-Bench framework are presented here.}
\label{PR_Curves_All_VPRBench}
\vspace{-0mm}
\end{figure*}

\begin{table*}
% \notsotiny
\caption{The values of AUC-PR are listed here for all the techniques on the $12$ datasets. The bold values in each row represent the state-of-the-art technique for each dataset for the corresponding metric.}
\label{auc,ep}       % Give a unique label
\setlength\tabcolsep{2.5pt}
\begin{tabular}{ccccccccccc}
\hline\noalign{\smallskip}
Dataset Name & NetVLAD & RegionVLAD & CoHOG & HOG & AlexNet & AMOSNet & HybridNet & CALC  & AP-GeM & DenseVLAD\\
\noalign{\smallskip}\hline\noalign{\smallskip}
Gardens Point & 0.70 & 0.56 & 0.42 & 0.28 & 0.47 & 0.57 & 0.59 & 0.38 & 0.67 & \textbf{0.77}\\
SPEDTest & 0.81 & 0.61 & 0.48 & 0.63 & 0.63 & \textbf{0.91} & 0.90 & 0.67 & 0.71 & 0.85\\
Nordland & 0.08 & 0.12 & 0.02 & 0.02 & 0.20 & \textbf{0.30} & 0.17 & 0.12 & 0.06 & 0.13\\
Living Room & 0.94 & 0.94 & 0.85 & \textbf{1.00} & 0.95 & 0.98 & 0.97 & 0.70 & 0.93 & 0.99\\
Synthia & 0.92 & 0.60 & 0.79 & \textbf{0.99} & 0.88 & 0.89 & 0.91 & 0.90 & 0.97 & \textbf{0.99}\\
17Places & 0.39 & 0.38 & 0.40 & 0.29 & 0.39 & 0.37 & 0.39 & \textbf{0.45} & 0.36 & 0.38\\
Cross-Seasons & \textbf{0.99} & 0.94 & 0.72 & 0.87 & \textbf{0.99} & 0.98 & \textbf{0.99} & 0.71 & 0.98 & \textbf{0.99}\\
Corridor & 0.83 & 0.66 & 0.69 & 0.68 & 0.80 & \textbf{0.95} & 0.93 & 0.78 & 0.85 & 0.89\\
Tokyo 24/7 & 0.89 & 0.42 & 0.09 & 0.00 & 0.06 & 0.25 & 0.28 & 0.01 & 0.78 & \textbf{0.95}\\
ESSEX3IN1 & 0.71 & 0.55 & 0.80 & 0.09 & 0.16 & 0.30 & 0.32 & 0.16 & 0.72 & \textbf{0.98}\\
Pittsburgh & 0.94 & 0.73 & \textbf{0.97} & 0.01 & 0.05 & 0.08 & 0.08 & 0.02 & 0.86  & 0.95 \\
INRIA Holidays & 0.90 & 0.94 & 0.76 & 0.39 & 0.79 & 0.89 & 0.92 & 0.77 & 0.98 & \textbf{0.99}\\
\noalign{\smallskip}\hline
\end{tabular}
\end{table*}

\begin{table*}
\caption{The values of feature encoding time $t_e$ (sec), descriptor matching time $t_m$ (msec) are listed here for $8$ VPR techniques. Encoding time is dependent upon the image resolution, however in this work we have used the recommended image resolutions by the authors of the respective VPR techniques and therefore $t_e$ is independent of the underlying dataset. The second row reports $t_m$ for the techniques' default data-types as given in the 6th row, while the values of $t_m$ in the third row are for fixed float-64 data-type of descriptors for all techniques. Please see accompanying text regarding trends of the descriptor matching time. The 4th row shows feature descriptor sizes of all $8$ VPR techniques in Kilo-Bytes (KBs) for a single image, along with the descriptor dimensions and default data-types in the following rows. The bold values in each row represent the state-of-the-art technique for the corresponding metric. Because DenseVLAD and GeM results have been computed using a different computational platform, the values for these techniques have not been included here to keep the comparison fair.}
\label{tm,desc_size}       % Give a unique label
\begin{tabular}{ccccccccc}
\hline\noalign{\smallskip}
Metric & NetVLAD & RegionVLAD & CoHOG & HOG & AlexNet & AMOSNet & HybridNet & CALC  \\
\noalign{\smallskip}\hline\noalign{\smallskip}
$t_e$ \; & 3.71 & 1.29 & 0.06 & \textbf{0.007} & 1.14 & 0.80 & 0.81 & 0.04\\
$t_m$ \; (default) & 0.06 & 0.17 & 2.64 & 0.07 & 0.03 & 0.13 & 0.13 & \textbf{0.02}\\
$t_m$ \; (float-64) & 0.08 & 0.17 & 6.91 & 0.49 & \textbf{0.04} & 0.13 & 0.13 & \textbf{0.04}\\
Desc. Size (KBs) & 16.38 & 786 & 123 & 138.38 & 8.51 & 61.4 & 61.4 & \textbf{4.25}\\
Desc. Dimensions & $1 \times 4096$ & $256 \times 384$ & $32 \times 961$ & $1 \times 34596$ & $1 \times 1064$ & $256 \times 30$ & $256 \times 30$ & $1 \times 1064$\\
Data Type & float-32 & float-64 & float-32 & float-32 & float-64 & float-64 & float-64 & float-32\\

\noalign{\smallskip}\hline
\end{tabular}
\end{table*}

We now present the results obtained by executing the VPR-Bench framework given the attributes presented in Section \ref{VPR-Bench_Framework}. 

\textbf{PR-Curves:} Firstly, the precision-recall curves for all $10$ VPR techniques on the $12$ indoor and outdoor datasets are presented in Fig. \ref{PR_Curves_All_VPRBench}. The values of AUC-PR for all techniques have been listed in Table \ref{auc,ep}. From the perspective of place matching precision, VPR-specific deep-learning techniques generally perform better than non-deep-learning techniques, with the exception of CoHOG and DenseVLAD, which always performs better than AlexNet and CALC. While CoHOG can handle lateral viewpoint-variation, it cannot handle 3D viewpoint-variation as present in the Tokyo 24/7 dataset. NetVLAD and DenseVLAD can handle 3D viewpoint-variation better than any other technique, because the training dataset for these contained 3D viewpoint-variations. HybridNet and AMOSNet can handle only moderate viewpoint-variations, but perform well under conditional variations due to training on highly conditionally-variant SPED dataset. Please note that the SPED dataset and SPEDTest dataset do not contain the same images, therefore the state-of-the-art performance of HybridNet and AMOSNet on SPEDTest dataset advocates for the utility of deep-learning techniques in environments similar to training environments (which in this case is the world from a CCTV's point-of-view). 

All techniques suffer on the Nordland dataset which contains significant perceptual aliasing and a large reference database. HOG and AlexNet usually lie on the lower-end of matching capabilities for all viewpoint-variant datasets, but perform acceptably on moderately condition-variant datasets that have no viewpoint variation. A notable exception here is the state-of-the-art performance of HOG compared to all other techniques on the Living Room dataset, which consists of high-quality images of places under indoor illumination variations. This suggests that on very small-scale datasets (and therefore for such small-scale indoor robotics applications), simple handcrafted techniques can yield good matching performance even under moderate variations in viewpoint and illumination. CALC cannot handle conditional variations to the same level as other deep-learning-based techniques, as the auto-encoder in CALC is only trained to handle moderate and uniform illumination changes. Region-VLAD also performs in the same spectrum as NetVLAD, but cannot surpass it on most datasets. All techniques perform poorly on the 17 Places dataset that represents a challenging indoor environment with strict viewpoint variance, suggesting that the outdoor performance success of techniques cannot be extended to an indoor environment. The perceptual-aliasing of datasets like Cross-Seasons and Synthia also presents significant challenges to VPR techniques. The AUC-PR of HOG comes out as $1$ for the Living Room dataset, because a threshold exists above which all images are correct matches (17 out of 32) and below which (15 out of 32) all images are incorrect matches. The results on Pittsburgh dataset and Tokyo 24/7 dataset identify two very separable clusters of VPR techniques: those (e.g. AMOSNet, HybridNet, CALC) that cannot handle large reference databases which essentially have many distractors and those (e.g. NetVLAD, DenseVLAD, CoHOG) which can handle such large reference databases.

\begin{figure*}[t]
\begin{center}
%\fbox{\rule{0pt}{2in} \rule{0.9\linewidth}{0pt}
\includegraphics[width=1.0\linewidth]{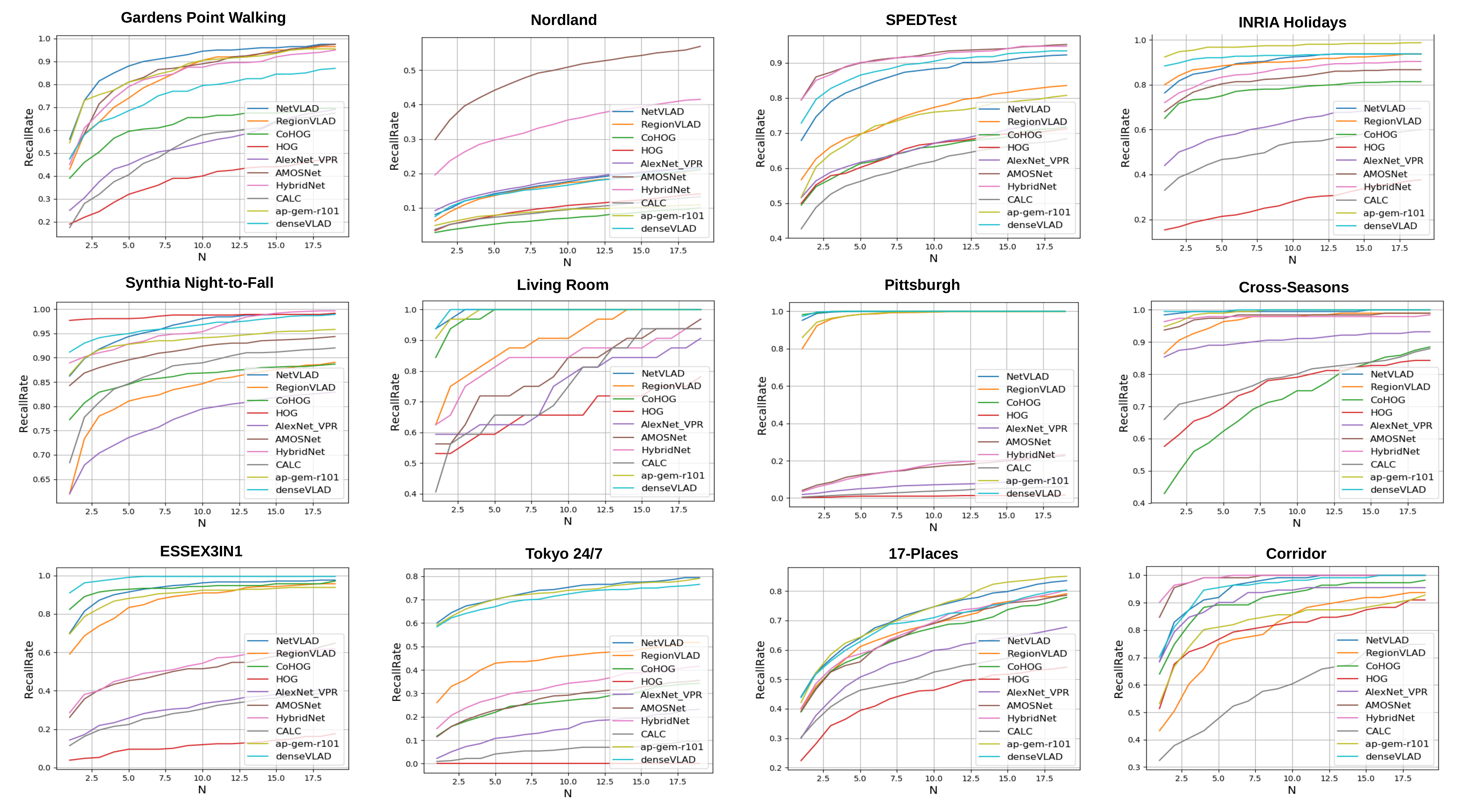}
\end{center}
\caption{The RecallRate@N curves for all $10$ VPR techniques generated on the $12$ datasets by VPR-Bench framework are presented here. The range of N used here is 1 to 20 with a step-size of 1. The values of RecallRate@1 represent the Precision@100\% Recall of a VPR technique.}
\label{RecallRate_Curves_All_VPRBench}
\vspace{-0mm}
\end{figure*}

\begin{figure*}[t]
\begin{center}
%\fbox{\rule{0pt}{2in} \rule{0.9\linewidth}{0pt}
\includegraphics[width=1\linewidth]{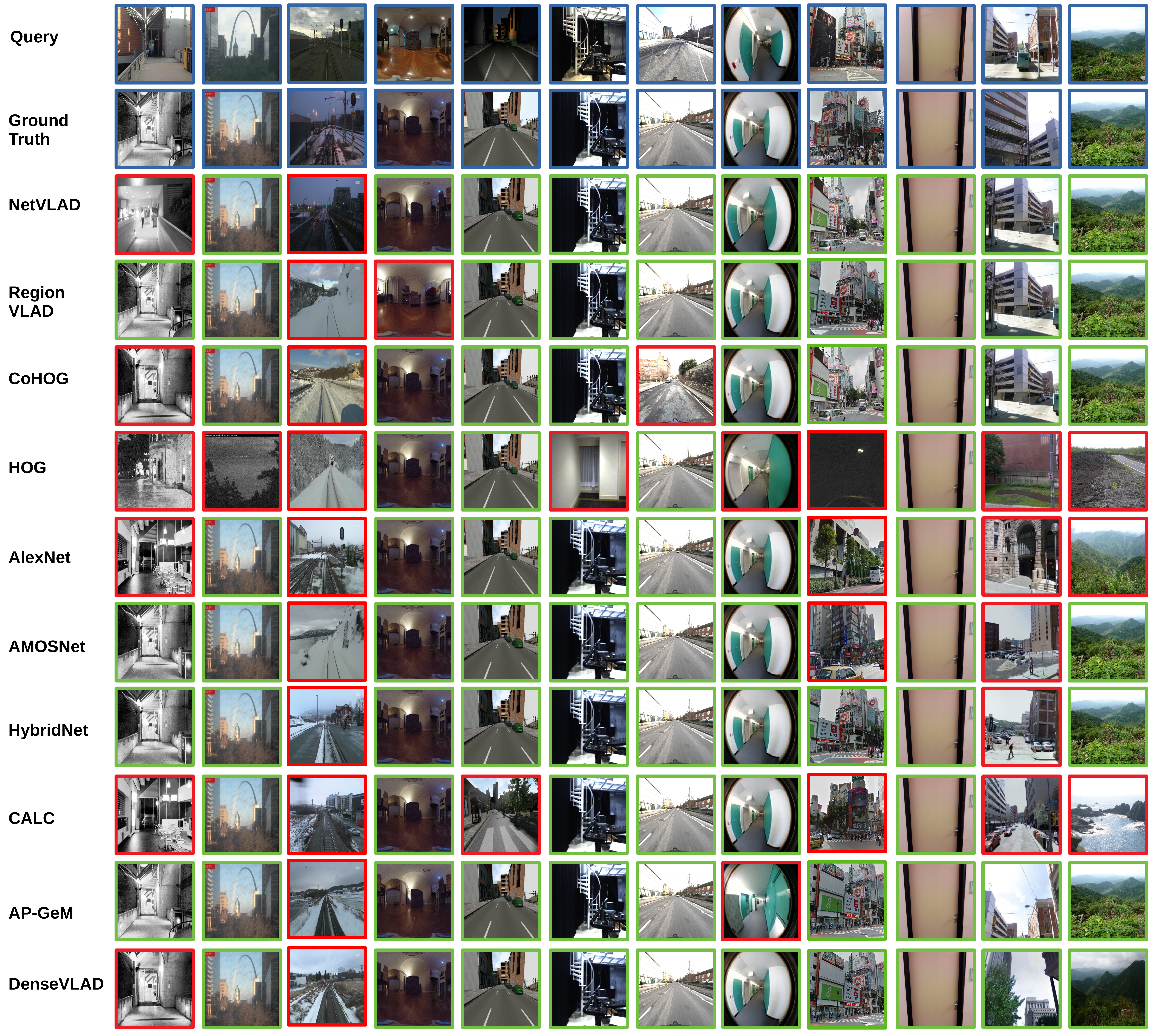}
\end{center}
\caption{Exemplar images matched/mismatched by VPR techniques are shown here for a qualitative insight. Red bounded images are incorrect matches (false positives) and green-bounded images are correct matches (true positives). An image is taken from each of the $12$ datasets, where the order of datasets from left to right follows the same sequence as top to bottom in Table \ref{auc,ep} first column. An important insight here is that some images are matched by all of the techniques, irrespective of the technique's complexities and abilities. This figure also suggests that because almost all of the images are matched by at least $1$ technique, an ensemble-based approach can significantly improve matching performance of a VPR-system.}
\label{Examplar_Matches}
\vspace{-0mm}
\end{figure*}

\textbf{RecallRate@N:} While for AUC-PR, the results have been listed in Table \ref{auc,ep}, RecallRate@N is usually represented as a trend and not as a single value. Therefore, for RecallRate@N, we plot the variations in RecallRate for values of N in the range of 1 to 20. These plots have been created for all the $10$ VPR techniques on the $12$ datasets and are shown in Fig. \ref{RecallRate_Curves_All_VPRBench}. Clearly, increasing/relaxing the value of N leads to an increase in RecallRate for all $10$ techniques and thus systems/applications that have a subsequent verification stage to re-rank the output of a VPR system would benefit from the trends presented in Fig. \ref{RecallRate_Curves_All_VPRBench}. An interesting insight is depicted by the values of N on which the ordering of techniques changes, which re-affirms the utility of this metric, for example see results on Gardens Point, ESSEX3IN1, Cross-Seasons and Corridor datasets. CALC starts from the bottom for RecallRate@1 on the Living Room dataset and sharply rises for later values of N.  It is important to note the changing state-of-the-art for RecallRate in comparison to AUC-PR, for example, DenseVLAD is the state-of-the-art on Tokyo 24/7 dataset for AUC-PR but for most values of RecallRate, NetVLAD and AP-GeM outperform DenseVLAD. Examples of images matched/mismatched by all VPR techniques on the $12$ datasets are shown in Fig. \ref{Examplar_Matches} for a qualitative insight.

\textbf{Computational Performance:} The values of feature encoding time, descriptor matching time and descriptor size have been listed in Table \ref{tm,desc_size} for our fixed platform. For all experiments in this work, we have used the default data-types of descriptors as specified in Table \ref{tm,desc_size} last row, however for the sake of complete comparison of matching time $t_m$, we affixed data-type of all techniques to float-64 for the values of $t_m$ in Table \ref{tm,desc_size} third row. The encoding time is usually higher for deep-learning-based techniques, while the matching time is generally higher for larger feature descriptors. Evidently, there are four factors affecting descriptor matching time: distance/similarity function, number of descriptor dimensions, length of each dimension and the descriptor data-type. For the reported 64-bit platform, cosine-distance as a similarity function and float-32 data-type, the change of size of a descriptor dimension (e.g. NetVLAD vs HOG in Table \ref{tm,desc_size} second row) has less effect on the matching time than a change in the total number of dimensions of a descriptor (e.g. NetVLAD vs CoHOG in Table \ref{tm,desc_size} second row). On the other hand, for float-64 data-type and fixed similarity function, the increase in matching time is almost linear with increasing size of a descriptor dimension (e.g. NetVLAD vs HOG in Table \ref{tm,desc_size} third row). AMOSNet has half the descriptor size than CoHOG, both descriptors are 2-dimensional, but the matching time for CoHOG is significantly higher than AMOSNet due to different distance functions, i.e. L1-matching for AMOSNet and cosine-distance for CoHOG.

Some of the key findings from the analysis in this sub-section can be summarised as follows:

\begin{enumerate}
    \item Unlike previous evaluations (\cite{zaffar2019levelling}, \cite{zaffar2019state}), where state-of-the-art AUC-PR performance was almost always achieved by NetVLAD, this paper shows that state-of-the-art AUC-PR performance is widely distributed among all the techniques across the $12$ datasets.
    
    \item The state-of-the-art technique for a particular dataset is metric-dependent and therefore, application-specific. A computationally-restricted application may find metrics like descriptor-size or retrieval-time important, while computationally-powerful platforms may only utilise AUC-PR and RecallRate.
    
    \item Interestingly, hand-crafted and non-deep-learning place recognition techniques can also achieve state-of-the-art performance. For DenseVLAD, this had been previously reported by \cite{sattler2018benchmarking} and \cite{torii2019large}, and we re-affirm their findings here. In our work, we also show how HOG and CoHOG have achieved state-of-the-art performance for all metrics on at least one dataset (see results on Synthia Night-to-Fall dataset and Pittsburgh dataset in Table \ref{auc,ep}). 
    
    \item Applications where the explored environment is small (e.g, a home service robot as in the Living Room dataset) and the variations are moderate, it is better to use a handcrafted computationally-efficient technique, as suggested by results in Table \ref{auc,ep} for Living Room dataset.
    
    \item Learning-based techniques that are trained on feature-full datasets do not extend well to non-salient, perceptually-aliased and feature-less environments. See for example the matching results on the Nordland dataset and Corridor dataset in Fig. \ref{RecallRate_Curves_All_VPRBench} and Table \ref{auc,ep}.
    
    \item Because state-of-the-art performance is distributed across the entire set of VPR techniques, an ensemble-based approach presents more value to VPR than a single-technique-based VPR, provided that the high computational and storage requirements of an ensemble can be afforded.
    
    \item A perfect AUC-PR score (i.e. equal to one) may be misinterpreted as a technique retrieving correct matches for all the query images in the dataset. However, a perfect AUC-PR in fact only means that when the query images and their retrieved matches are collectively arranged in a descending order based on confidence scores, all the true-positives lie above all the false-positives. Thus, it is important that the RecallRate@N (for some value of N) of VPR techniques is also reported in addition to AUC-PR. See for example the AUC-PR and RecallRate@1 of HOG on the Living Room dataset, where the former proposes perfect VPR performance while the latter shows a significant room for improvement.
    
    \item The descriptor size of techniques is also a key evaluation metric to be considered. A large descriptor size not only translates into excessive storage needs for the respective reference maps, but also affects the descriptor matching time and leads to higher run-time memory (RAM) consumption/needs. We further present analysis on this in sub-section \ref{descriptor_size_analysis}.
    
\end{enumerate}

\subsection{ROC Curves: Finding New Places}
\begin{figure}[t]
\begin{center}
%\fbox{\rule{0pt}{2in} \rule{0.9\linewidth}{0pt}
\includegraphics[width=1\linewidth]{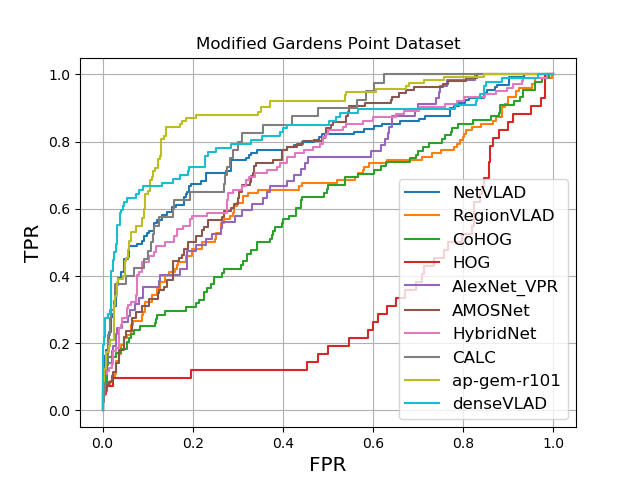}
\end{center}
\caption{The ROC performance of $10$ VPR techniques is shown here on a modified (true-negative added) version of Gardens Point dataset that contains $200$ true-negatives and $200$ true-positives.}
\label{ROC_curves}
\vspace{-0mm}
\end{figure}

Next, we show the ROC curves for all techniques on a modified version of the Gardens Point dataset. We have modified the Gardens Point dataset to contain $200$ queries as true-negatives in addition to its existing $200$ true-positives. The number of true-positives and true-negatives is kept equal, because ROC curves work well for balanced classification problems. These curves have been shown in Fig. \ref{ROC_curves}. We note that unlike the PR-curves for the techniques on Gardens Point dataset, where most techniques perform very well, the class separation capacity (ROC performance) of these techniques is not as good. However, among the techniques, learning-based techniques clearly outperform handcrafted VPR techniques. Although CALC cannot perform well among learning-based techniques for PR curves, the ROC curves show that it has a better class separation capacity than most of the other learning-based techniques. The AUC-ROC for all the techniques has also been listed in Table \ref{auc_roc} and all techniques generally achieve a lower AUC-ROC than ideal. The AUC-ROC of HOG is less than 0.5, because it yields opposite labels for true-positives and true-negatives (i.e. existing places are classified as new places and vice versa).

\begin{table*}
\caption{The values of AUC-ROC achieved by $10$ VPR techniques on the modified (true-negative added) version of the Gardens Point dataset have been reported here.}
\label{auc_roc}       % Give a unique label
\footnotesize
\begin{tabular}{cccccccccc}
\hline\noalign{\smallskip}
 NetVLAD & RegionVLAD & CoHOG & HOG & AlexNet & AMOSNet & HybridNet & CALC & AP-GeM & DenseVLAD  \\
\noalign{\smallskip}\hline\noalign{\smallskip}
 0.77 & 0.64 & 0.60 & 0.31 & 0.70 & 0.74 & 0.74 & 0.82 & 0.87 & 0.82\\
\noalign{\smallskip}\hline
\end{tabular}
\end{table*}

\subsection{Computational Performance: CPU vs GPU}
While the previous sub-sections have shown the performance of 10 VPR techniques on the fronts of place matching precision and computational requirements, the underlying hardware has been a CPU-only platform. Generally, CPU represents the common computational hardware for resource-constrained platforms, but learning-based techniques are favored well by GPU-based platforms. Thus, depending on the underlying platform characteristics (CPU vs GPU), it may or may not be fair to compare handcrafted VPR techniques with deep-learning-based VPR techniques on computational front.

We here report the feature encoding time $t_e$ and the descriptor matching time $t_m$ of the 7 deep-learning-based techniques in our suite when implemented on a GPU-based platform. The GPU-based evaluation was performed using an Nvidia GeForce GTX 1080 Ti with 12GB memory using a batch size of 1. The mechanism for computation of the timings is the same as that for CPU (i.e. averaged over the entire dataset) and the same codes/parameters were used as those for CPU. We have reported these timings in Table \ref{gpu_performance} for the Gardens Point dataset.

It can be observed that the GPU-based ordering of methods is mostly similar as their CPU-based ordering (see Table \ref{tm,desc_size}), with notable exception of RegionVLAD vs NetVLAD for $t_e$, because of the former's compute-intensive CPU-based region-extraction and VLAD description. In general, the computation times between CPU and GPU vary noticeably for all the methods. This cross-analysis highlights the varying utility of VPR techniques across different platforms.

\begin{table}
\caption{The values of encoding times and matching times for $7$ VPR techniques on the Gardens Point dataset for a GPU-based platform have been reported here.}
\label{gpu_performance}       % Give a unique label
\footnotesize
\begin{tabular}{ccc}
\hline\noalign{\smallskip}
 VPR Technique & $t_e$ (seconds) & $t_m$ (milliseconds)  \\
\noalign{\smallskip}\hline\noalign{\smallskip}
 NetVLAD & 0.075 & 0.002 \\
 RegionVLAD & 0.451 & 0.061 \\
 AMOSNet & 0.032 & 0.038 \\
 HybridNet & 0.032 & 0.035 \\
 CALC & \textbf{0.001} & \textbf{0.001} \\
 AP-GeM & 0.027 & 0.045 \\
 AlexNet & 0.203 & \textbf{0.001} \\
\noalign{\smallskip}\hline
\end{tabular}
\end{table}

\subsection{Descriptor Size Analysis}
\label{descriptor_size_analysis}
In this sub-section, we further extend upon the descriptor size analysis and show that changing the descriptor size affects various performance-related aspects of a VPR technique, in particular memory footprint, place matching precision and descriptor matching time. To perform this analysis, we use the Gardens Point dataset and change various descriptor-related parameters of 5 VPR techniques, namely CoHOG, HOG, NetVLAD, DenseVLAD and AP-GeM, that directly affect the descriptor size.

For HOG and CoHOG, we have changed the cell-size of the HOG-computation scheme, where the block-size remained twice of the cell-size and all the other parameters like image-size and bin-size were kept constant. For NetVLAD, DenseVLAD and AP-GeM, we changed the PCA output dimensions while all other parameters were kept constant. The effect of these descriptor size changes on the memory footprint (descriptor size), AUC-PR and descriptor matching time is reported in Table \ref{descriptor_size_analysis_table}. The absolute and relative variation of these different performance indicators by changing descriptor size is dependent upon the underlying matching scheme and descriptor dimensions, and this variation is therefore not constant between the different VPR techniques. However, there is a general trend where increasing the descriptor dimension leads to increased descriptor matching time and memory footprint, while AUC-PR also varies for VPR techniques.

The descriptor matching time usually decreases by varying parameters that lead to the decrease of descriptor size. The change in AUC-PR by varying descriptor dimensions is subject to the intrinsics of the individual VPR techniques and the role of their corresponding parameters. For deep-learning-based techniques followed by PCA (see NetVLAD and AP-GeM in Table \ref{descriptor_size_analysis_table})), decrease of descriptor size may or may not lead to decrease of AUC-PR, because a decreased descriptor size can lead to either the decrease of confusing/non-salient features (e.g. those coming from vegetation, dynamic objects etc) or distinguishable/salient features and/or a combination of both. The AUC-PR variation for NetVLAD and AP-GeM generally follows a descending trend with decreasing PCA dimensions, but does remain constant for some immediate steps/levels of PCA.
The learning-based DenseVLAD (albeit not deep-learning-based) suffers significantly from the decreased descriptor size. For CoHOG, the AUC-PR variation is similar to the original findings in \cite{zaffar2020cohog}, where increasing cell-size leads to reduced viewpoint invariance and lesser AUC-PR. For HOG the increased cell-size (which reduces descriptor size) actually leads to an increase of AUC-PR due to the optimal settings for the traditional fully global HOG-descriptor scheme. The AUC-PR of HOG is highest for cell-size of 64 $\times$ 64 but decreases when the cell-size in either increased or decreased from this optimal setting. Please note that this optimal setting of the cell-size may differ for different datasets depending on the amount and nature of viewpoint and conditional variations in the dataset.

\begin{table*}[]
\caption{The values of AUC-PR, descriptor size (Kilo-Bytes) and matching time (msec) are reported on the Gardens Point dataset by varying descriptor size-related parameters (cell-size and PCA-dimensions) of VPR techniques. Please note that the computations for AP-GeM and DenseVLAD were done on a platform different from that of NetVLAD, HOG and CoHOG. The maximum PCA dimensions given the AP-GeM default design are 2048.}
\label{descriptor_size_analysis_table}       % Give a unique label
\notsotiny
\setlength\tabcolsep{3.5pt}
\begin{tabular}{|cccc|cccc|cccc|cccc|cccc|}
\hline
\multicolumn{4}{|c|}{CoHOG}            & \multicolumn{4}{c|}{HOG}               & \multicolumn{4}{c|}{NetVLAD}                              & \multicolumn{4}{c|}{DenseVLAD}        & \multicolumn{4}{c|}{AP-GeM}           \\ \hline
Cell-Size & AUC & KBs & $t_m$ & Cell-Size & AUC & KBs & $t_m$ & PCA & AUC & KBs & $t_m$ & PCA & AUC & KBs & $t_m$ & PCA & AUC & KBs & $t_m$ \\ \hline
8x8       & 0.47 & 508        & 47.0 & 8x8       & 0.19 & 571     & 0.14 & 4096     & 0.69 & 16.30 & 0.06                   &  
4096 & 0.77 & 16.30 & 0.06 &      
4096    & - & - &      \\ \hline

16X16     & 0.42   & 123        & 2.64  & 16X16     & 0.29   & 138     & 0.07 & 2048     & 0.69 & 8.19       & 0.06                   &  
2048 & 0.69 & 8.19 & 0.06 &      
2048    & 0.67 & 8.19 & 0.06 \\ \hline

32X32     & 0.36   & 28.8       & 0.18 & 32X32     & 0.29 & 32.4      & 0.06 & 1024     & 0.59 & 4.09       & 0.05                   &    
1024      & 0.64       & 4.09 & 0.05 &    
1024      & 0.65 & 4.09 & 0.05 \\ \hline

64X64     & 0.30   & 6.27       & 0.06 & 64X64     & 0.35 & 7.05      & 0.05 & 512     & 0.59 & 2.04       & 0.05                   &  
512    & 0.58 & 2.04 & 0.05 &   
512       & 0.67 & 2.04 & 0.05 \\ \hline

128X128     & 0.19   & 1.15       & 0.05 & 128X128     & 0.33 & 1.29      & 0.04 & 256     & 0.52 & 1.02       & 0.04                   &      
256    & 0.52 & 1.02 & 0.04 &   
256       & 0.64 & 1.02 & 0.04 \\ \hline

256X256     & 0.12   & 0.128       & 0.03 & 256X256     & 0.16 & 0.14      & 0.02 & 128     & 0.52 & 0.51       & 0.02                   &      
128    & 0.33 & 0.51 & 0.02 &   
128       & 0.62     &  0.51          & 0.02      \\ \hline

\end{tabular}
\end{table*}

\subsection{True-Positives Trajectory Distribution}
\begin{figure*}[t]
\begin{center}
%\fbox{\rule{0pt}{2in} \rule{0.9\linewidth}{0pt}
\includegraphics[width=1.0\linewidth]{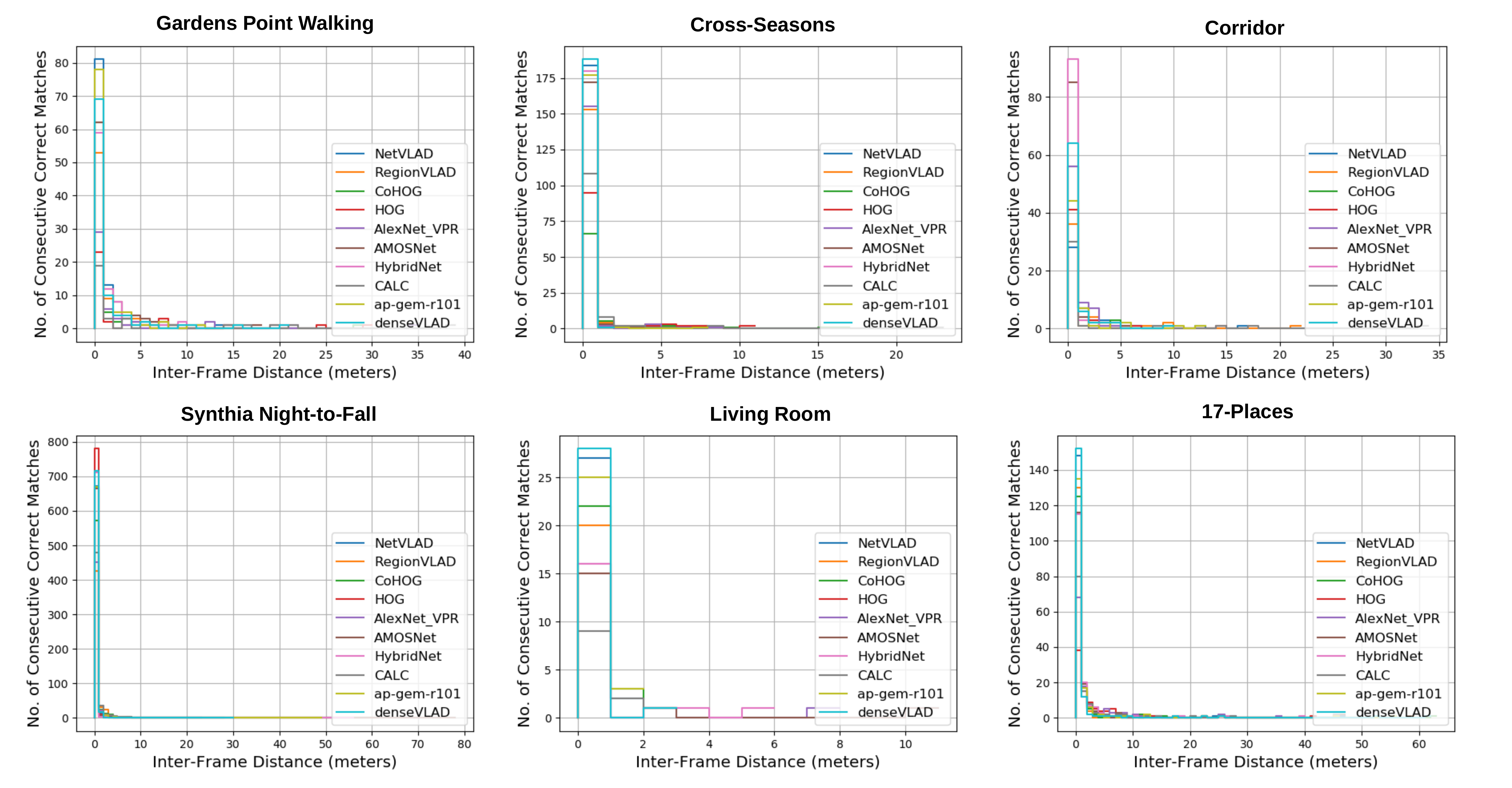}
\end{center}
\caption{The distribution of true-positives over the trajectory of a dataset are shown here. The horizontal axis represents the distance between two consecutive true-positives in a sequence and the vertical axis shows the number of true-positives that satisfy this distance constraint.}
\label{TPdistribution_Curves_All_VPRBench}
\vspace{-0mm}
\end{figure*}

In addition to the image retrieval timings, it is important to look at the distribution of true-positives (loop-closures) within a dataset sequence. Therefore, as explained in sub-section \ref{True_Positives_Distribution}, we report  in Fig. \ref{TPdistribution_Curves_All_VPRBench} the distribution of true-positives for 6 trajectory-based datasets. The distribution here refers to no. of true-positives (Y-axis) for a given distance (X-axis) between two correctly retrieved frames. For all the datasets, we have assumed an inter-frame distance of 1 meter, i.e. true-positives that are assumed to be 5 meters apart represent two correctly-matched query frames that are 5 frames apart. This assumption is required because we do not have the exact knowledge of inter-frame physical distance for all the datasets and because the X-axis can be easily scaled-up to represent a different inter-frame distance. 

Ideally, all techniques should have a single peak value equal to the total number of query images at the vertical axis in Fig. \ref{TPdistribution_Curves_All_VPRBench}. For most techniques on all the datasets, the loop-closures are distributed evenly i.e. curves in Fig. \ref{TPdistribution_Curves_All_VPRBench} peak at small values of X-axis. There is a ripple effect that starts from Y-axis and dies towards larger values of inter-frame distance. This ripple effect is more distributed for Gardens Point and Corridor datasets than the other datasets. Thus, for applications such as SLAM where VPR is used in addition to a visual-localisation system, techniques can mostly achieve periodic loop-closure and correct error-drifts. However, these ripples can be catastrophic for VPR-based topological/primary localisation systems (\cite{cummins2011appearance}) which rely solely on location estimated through VPR. We have not provided this analysis for non-trajectory-type datasets (SPEDTest, INRIA Holidays etc), because the inter-frame distance is not a valid assumption for these cases.

\subsection{Acceptable Ground-truth Manipulation}
\label{acceptable_groundtruth_manipulation}
\begin{figure}[t]
\begin{center}
%\fbox{\rule{0pt}{2in} \rule{0.9\linewidth}{0pt}
\includegraphics[width=1\linewidth]{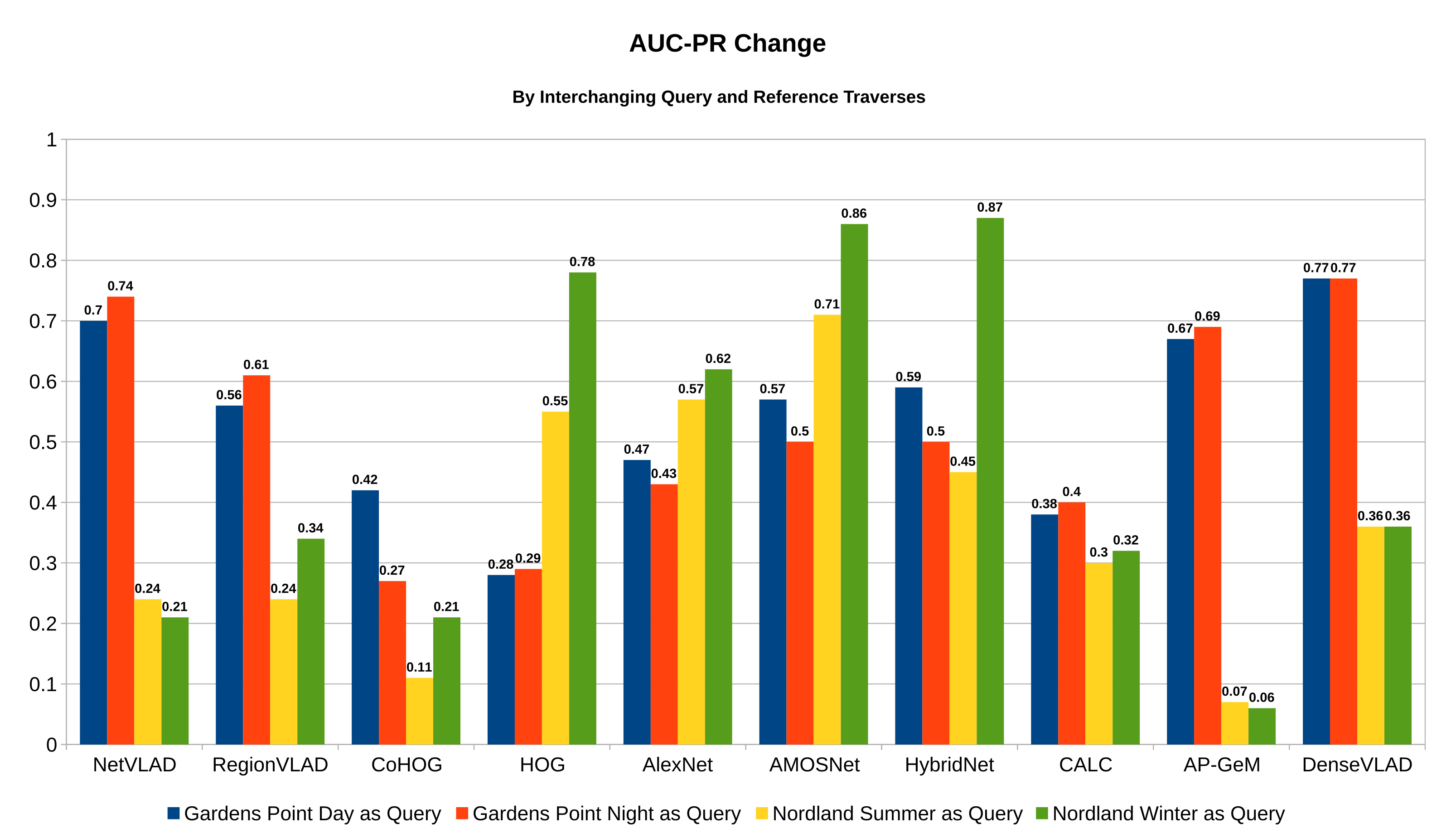}
\end{center}
\caption{The effect on AUC-PR performance of techniques by inter-changing the query and reference traverses is shown here for the Gardens Point dataset and Nordland dataset.}
\label{auceffect_queryref_interchange}
\vspace{-0mm}
\end{figure}

An important finding from the analysis performed for sub-section \ref{VPR Performance Evaluation} was that the matching performance also varies depending on the ground-truth information in a VPR dataset. It is possible that the ground-truth is slightly modified such that the new ground-truth is usually acceptable to the reviewing audience, but it also leads to a change of state-of-the-art technique on a particular dataset.
For example, the matching performance varies if the query and reference databases are inter-changed (i.e. query folder becomes the new reference folder and reference folder becomes the new query folder), especially for conditionally-variant datasets. We show this in Fig. \ref{auceffect_queryref_interchange} for the Nordland and Gardens Point dataset. Here we use a small section of the Nordland traversal (as used in \cite{merrill2018lightweight}, \cite{zaffar2019levelling}) containing 1622 query and 1622 reference images such that the effects of ground-truth manipulation are more prominent, since all the techniques have very low precision on the full traversal. Interestingly, this analysis reveals that for all the VPR techniques the rise/decline in performance is not necessarily the same in magnitude and direction. Changing ground-truth in this manner is based on the constraint that reference matches for queries are available from a particular conditional appearance (weather, seasons, time etc) and that this condition is different from that of query images. This is normally the case for most of the robotics-focused VPR datasets and for applications like teach-and-repeat. This analysis assumes the non-existence of the same appearance conditions of a place in query and reference images.

\begin{figure}[t]
\begin{center}
%\fbox{\rule{0pt}{2in} \rule{0.9\linewidth}{0pt}
\includegraphics[width=1\linewidth]{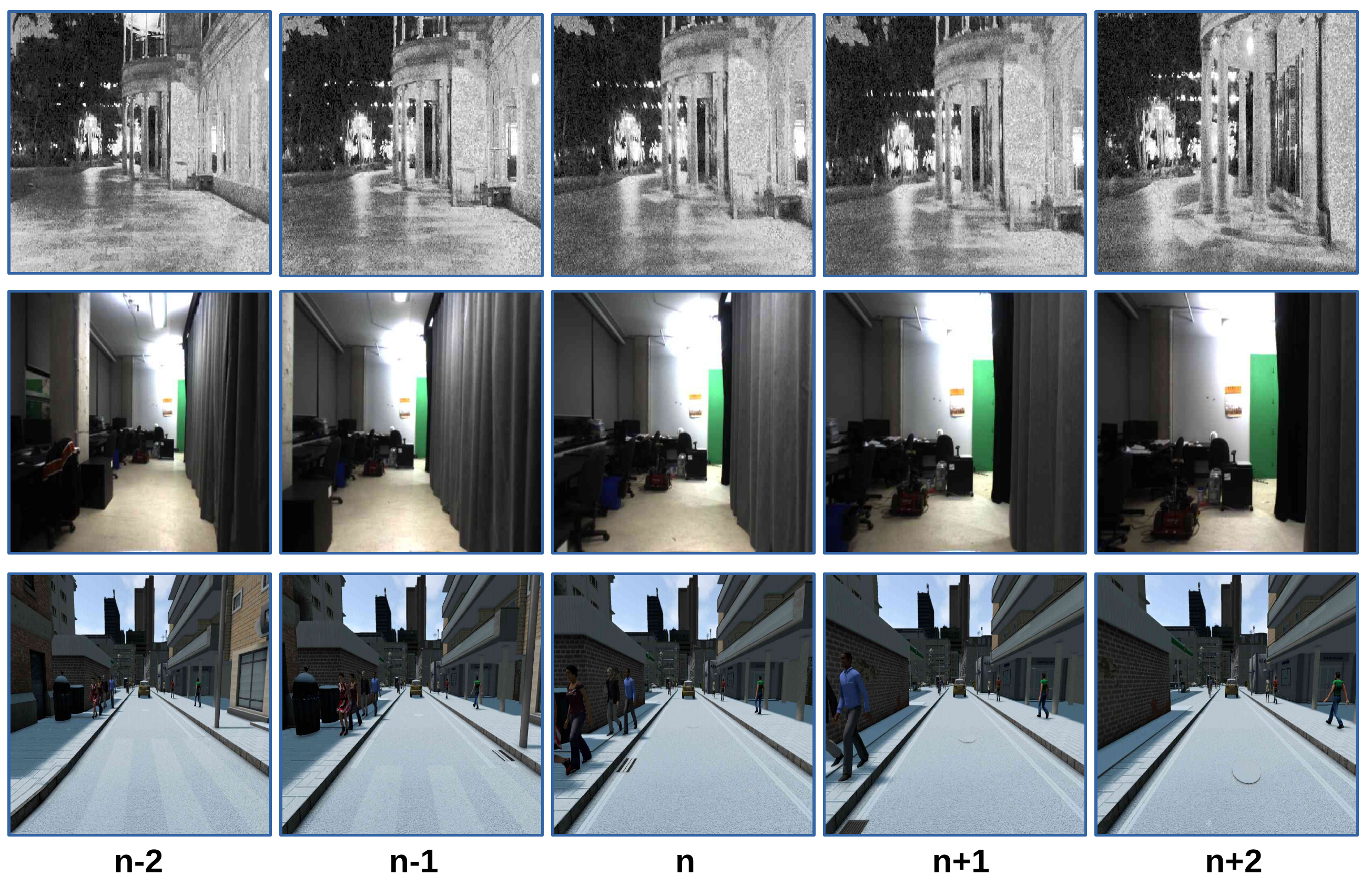}
\end{center}
\caption{The overlap between visual information among subsequent images in traversal-based datasets is shown here. Depending on what level of ground-truth true positive range is acceptable, benefits will be distributed among the techniques based on their viewpoint-invariance.}
\label{visualoverlap_samples}
\vspace{-0mm}
\end{figure}

\begin{figure*}[t]
\begin{center}
%\fbox{\rule{0pt}{2in} \rule{0.9\linewidth}{0pt}
\includegraphics[width=1\linewidth]{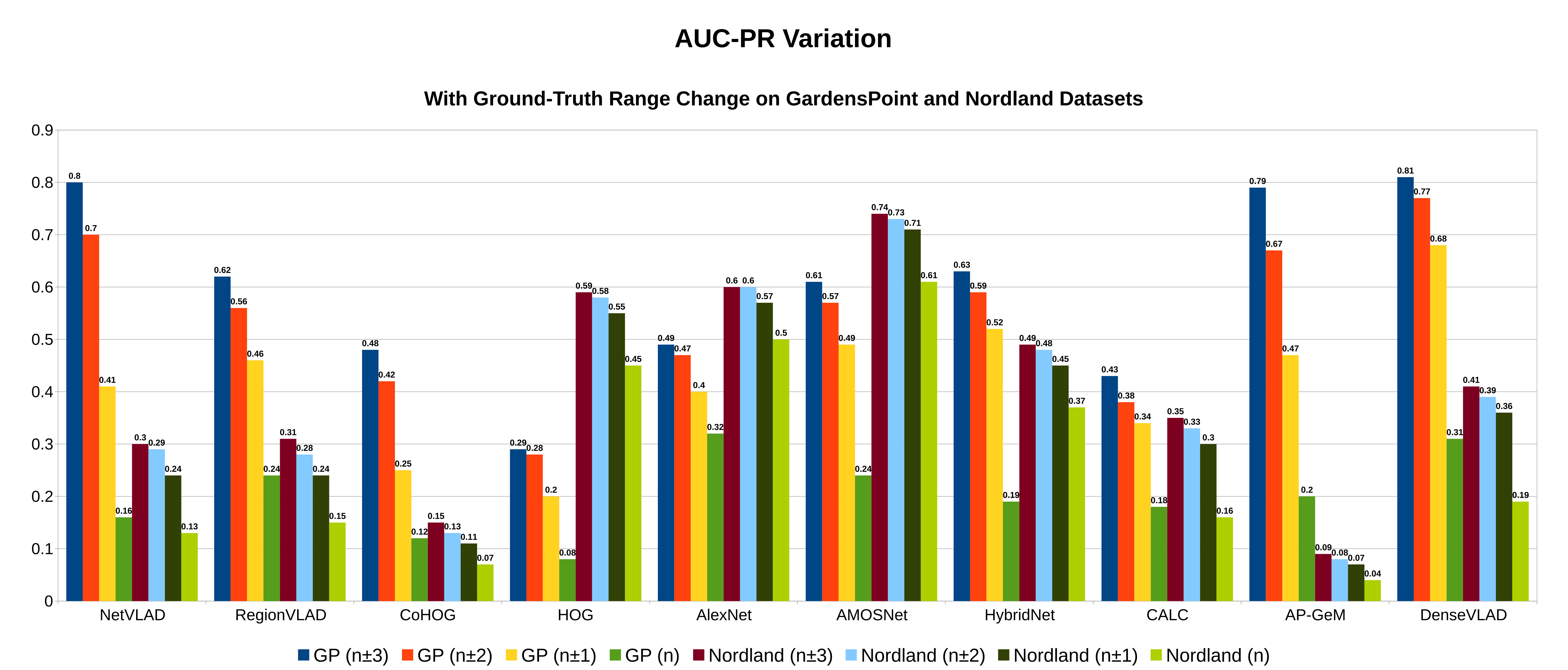}
\end{center}
\caption{The effect on AUC-PR performance of techniques by changing the range of ground-truth true positive images is shown here for the Gardens Point dataset and Nordland dataset.}
\label{auceffect_gtrangechange}
\vspace{-0mm}
\end{figure*}

Moreover, in most of the traversal-based VPR datasets, there is always some level of overlap in visual content in between consecutive frames. Thus, techniques which are viewpoint-invariant may get benefits if the ground-truth identifies such frames as correct matches. On the other hand, if the ground-truth only considers frame-to-frame matches (i.e. one query frame has only one correct matching reference frame), such viewpoint-invariant techniques may not get the same matching performance (in the form of AUC-PR, RecallRate@N, EP etc), because their viewpoint invariance will actually lead to false positives. Examples of these consecutive frames with visual overlap are shown in Fig. \ref{visualoverlap_samples}. We report this effect of changing ground-truth range on the AUC-PR of various VPR techniques for the Gardens Point dataset and Nordland dataset in Fig. \ref{auceffect_gtrangechange}. One could argue that a correct ground-truth must regard such viewpoint-variant images of the same place as true positives, however, a contrary argument exists for applications that utilise VPR as the primary (only) module for localisation, as discussed further in subsection \ref{Variance vs Invariance}. This sub-section demonstrates that different state-of-the-arts (i.e. top performing techniques) can be created on the same dataset by manipulating the ground-truth information accordingly. 

\subsection{Retrieval Time vs Platform Speed}
One of the questions that we wanted to address through this manuscript is, `What is a good image-retrieval time?'. This is important because most VPR research papers (as covered in our literature review) that claim real-time performance consider anything between 5-25 frames-per-second (FPS) as real-time. However, there are $2$ important caveats to such performance. Firstly, the retrieval performance for a VPR application depends on the size of the map. It is therefore very important that the size of the map is addressed either by presenting the limits for the map-size or by proposing methodologies to affix the map-size. Secondly, the retrieval performance is directly related to the platform speed. A real-time VPR application may require that a place-match (localisation) is achieved every few meters, while a dynamic platform traverses an environment. In such a case, the utility of a technique will depend upon the speed of the platform, as the faster the platform moves, the lower the retrieval time that is acceptable. We have modelled this as follows.

Let us assume that a particular application requires $K$ frames-per-meter (where $K$ could be fractional) and that the platform moves with a velocity $V$. Also, let the size of the map (no. of reference images) be $Z$. Then, the required FPS retrieval performance given the values of $K$ and $V$ is denoted as $FPS_{req}$ and computed as

\begin{equation} \label{FPS_req_eq}
    FPS_{req}=K \times V .
\end{equation}

\begin{figure*}[t]
\begin{center}
%\fbox{\rule{0pt}{2in} \rule{0.9\linewidth}{0pt}
\includegraphics[width=1\linewidth]{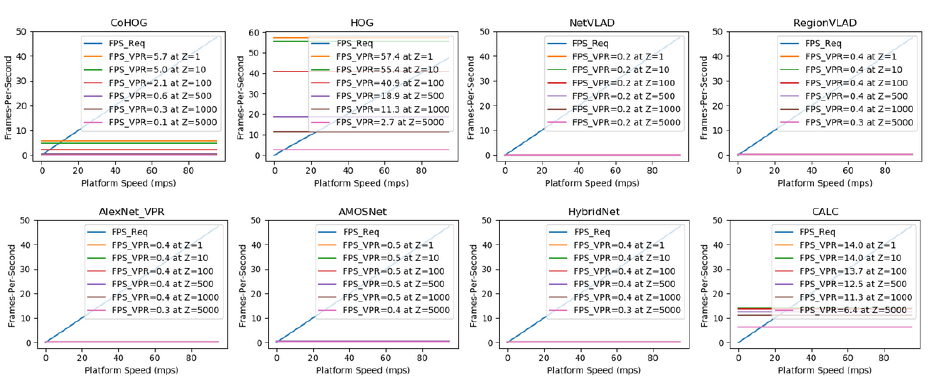}
\end{center}
\caption{The retrieval performance of techniques is drawn for different map-sizes ($Z$) across the platform speed. Depending upon the value of frames required per meter ($K$) for an application, these curves will scale linearly according to equation \ref{FPS_req_eq}.}
\label{retrievaltimevsplatformspeed}
\vspace{-0mm}
\end{figure*}

The retrieval performance of a VPR technique will depend on the number of reference images and can be denoted as $FPS_{VPR}$. This $FPS_{VPR}$ has been modelled previously in equation \ref{t_R}, such that $FPS_{VPR}=1/t_R$. Therefore, to understand the limits of real-time performance of a VPR technique given the application requirements ($V$, $K$ and $Z$), we draw the retrieval performance of all techniques along the platform speed for different values of $Z$ in Fig. \ref{retrievaltimevsplatformspeed}, assuming $K=0.5$ frames-per-meter. The curves for $FPS_{VPR}$ are straight-lines for constant values of $Z$ and the range of horizontal-axis (Speed $V$) for which $FPS_{VPR}$ is less than or equal to $FPS_{req}$ represents the range of platform speed (for that map-size) that a technique can handle. The VPR-Bench framework enables the creation of these curves conveniently and therefore, presents value to address the subjective real-time nature of a technique's retrieval time for VPR.

\subsection{Invariance Analysis} \label{Invariance Analysis}
One of the key aspects of the VPR-Bench framework as explained in Section \ref{VPR-Bench_Framework} is the quantification of viewpoint- and illumination-invariance of a VPR technique. In sub-section \ref{VPR Performance Evaluation}, we had utilised the traditional VPR analysis schema, where datasets are usually classified based on the qualitative severity of a particular variation. However, in this section, we utilise the Point Features dataset presented in sub-section \ref{Invariance_Quantification_Setup} and utilise the quantitative information presented in Fig. \ref{viewpoint_variation_pointfeaturedataset}, Fig. \ref{illumination_variation_pointfeaturesdataset} and Table \ref{azimuthandelevationangles}. 

\begin{figure*}[t]
\begin{center}
%\fbox{\rule{0pt}{2in} \rule{0.9\linewidth}{0pt}
\includegraphics[width=1\linewidth]{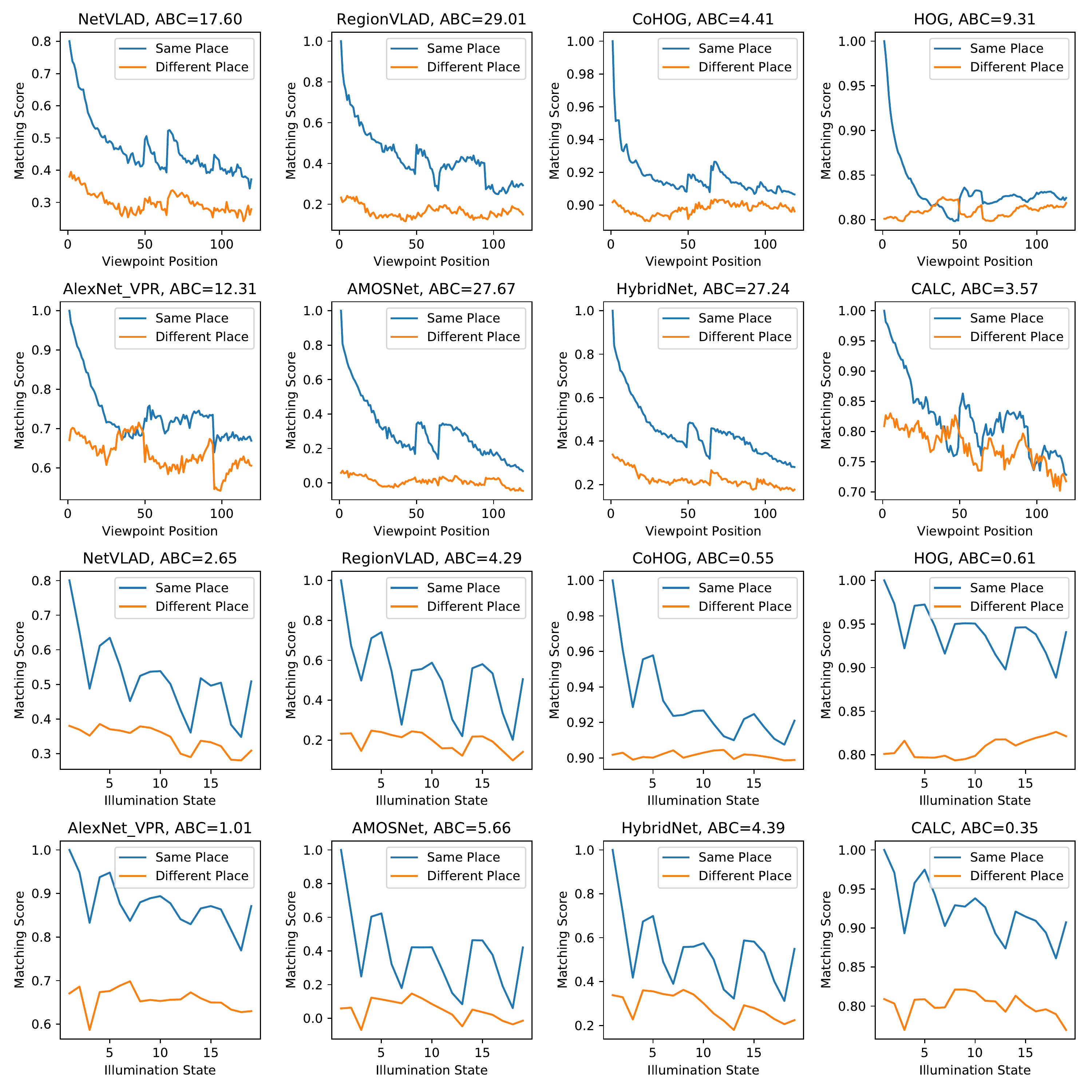}
\end{center}
\caption{The change in matching score for quantified viewpoint and illumination variations is shown here on the Point Features dataset. The first two rows contain changes for all techniques with $119$ viewpoint positions, while the bottom two row show these changes for $19$ different illumination levels. Please see accompanying text for analysis.}
\label{matchingscore_variation_pointfeaturesdataset}
\vspace{-0mm}
\end{figure*}

The change in matching score along these arcs is shown in Fig. \ref{matchingscore_variation_pointfeaturesdataset} for all the techniques. There is clear decline in matching scores as the viewpoint is varied both along the arcs and in-between the arcs. A key insight is that moving along the arcs has more effect (negative) on the matching score than jumping between the arcs (i.e. moving towards or away from the scene). From a computer vision perspective, this means that a change in the scale of the world (zooming-in, zooming-out) has lesser effect on matching scores than the change in 3D-appearance of the scene. 

Ideally, the matching scores for the same scene/place should be equal to $1$ for the range of variation a technique can handle and the matching score for a different scene/place should be $0$. However, in practice, all techniques give lower than $1$ matching scores, when two images of a scene have a particular variation in-between them, while giving higher than $0$ scores to places that are different. The point at which the matching score for the same-but-varied place is equal to or lower than `any' of the matching scores for different place, represents the absolute limits for that VPR technique. Please note, that the two curves (same-but-varied place and different place) should not be compared point-to-point, but instead point-to-curve, because the matching score for the same-but-varied place should not be less than `any' of the matching scores for different place. Thus, while it may appear that the two curves for NetVLAD do not intersect under any viewpoint positions, the matching score for the same-but-varied place for positions $110-119$ is almost equal to the matching score for different place at position $0$, which will lead to false positives. A conclusive remark from this viewpoint-variation analysis is that none of the $8$ VPR techniques in this work is immune to all levels of viewpoint-variation.  

Another benefit of having the matching scores curves for different place in contrast with the same-but-varied place is that it allows us to compute the Area-between-the-Curves (ABC) for each of the techniques. These values of ABC have been reported for all the techniques. Higher value of ABC represents that a technique can distinguish well between the same-but-varied place and a different place. The ideal value of ABC is equal to the number of variations (x-axis), as the matching score should remain $1$ along the entire x-axis in an ideal scenario. Please note that the ABC does not reflect the absolute matching performance of a VPR technique, and should not be compared with AUC-PR/EP/AUC-ROC, because the analysis in only based on two places/scenes. 

\begin{figure*}[t]
\begin{center}
%\fbox{\rule{0pt}{2in} \rule{0.9\linewidth}{0pt}
\includegraphics[width=1\linewidth]{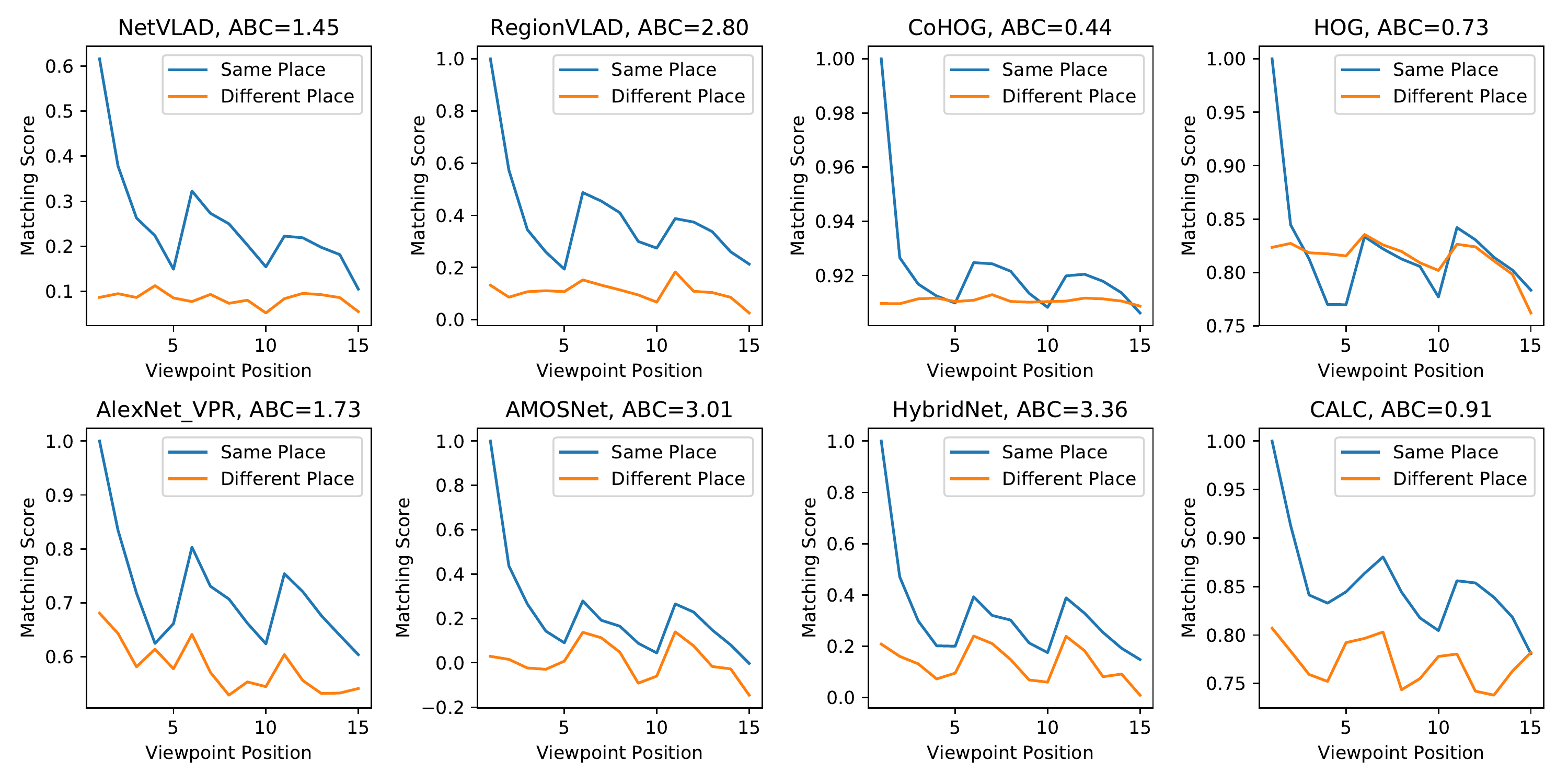}
\end{center}
\caption{The change in matching score for the quantified viewpoint variations is shown here on the QUT Multi-lane dataset. The confidence score variation is shown for all techniques against the $15$ viewpoint positions, as explained in sub-section \ref{Evaluation_Mechanism}.}
\label{matchingscore_variation_multilanedataset}
\vspace{-0mm}
\end{figure*}

We have extended the analysis of viewpoint-invariance from the synthetic Point Features dataset to the real-world QUT Multi-lane dataset. The analysis scheme is the same for both the datasets and the obtained curves are shown in Fig. \ref{matchingscore_variation_multilanedataset}. The curves on the QUT Multi-lane dataset re-affirm our findings from the Point Features dataset and the trends on both the datasets are similar. More importantly, lateral viewpoint changes have been shown to have a greater effect on the place matching confidence score than the forward/backward movement. The scale/level of this (for viewpoint variations on both Point Features dataset and QUT Multi-lane dataset) is however dependent upon the scene depth and the exact physical movement for lateral and forward/backward changes. Generally, for higher scene-depth, forward/backward movement leads to a lesser change in visual-content than lateral variations and therefore has a lesser effect. Very large forward/backward movement (definition of `very large' is dependent upon the scene depth) may lead to a greater reduction in confidence score than a small change in lateral viewpoint. 

A similar analysis is performed for the $19$ different matching scores given the $19$ quantified illumination variations, as shown in Fig. \ref{matchingscore_variation_pointfeaturesdataset}. While the $119$ different viewpoint positions represented in Fig. \ref{viewpoint_variation_pointfeaturedataset} are intuitive for analysis, the nature and level of illumination change in Table: \ref{azimuthandelevationangles} is not obvious. We have presented these $19$ different cases qualitatively in Fig. \ref{illumination_variation_pointfeaturesdataset_qualitative}, so that the illumination-variance curves in Fig. \ref{matchingscore_variation_pointfeaturesdataset} can be easily understood. It can be seen that uniform or close to uniform changes do not have much effect on the matching score. However, directional illumination changes that lead to the partitioning of a scene between highly-illuminated and low-illuminated portions has the most dramatic effect. An interesting insight is that some basic handcrafted VPR techniques (HOG-based) are able to distinguish between the same-but-illumination-varied places and different places, under all $19$ scenarios (i.e. no point on the same-but-varied place curve is lower than any point on the different place curve), while contemporary deep-learning-based techniques struggle with such illumination-variation.

\begin{figure*}[t]
\begin{center}
%\fbox{\rule{0pt}{2in} \rule{0.9\linewidth}{0pt}
\includegraphics[width=1\linewidth]{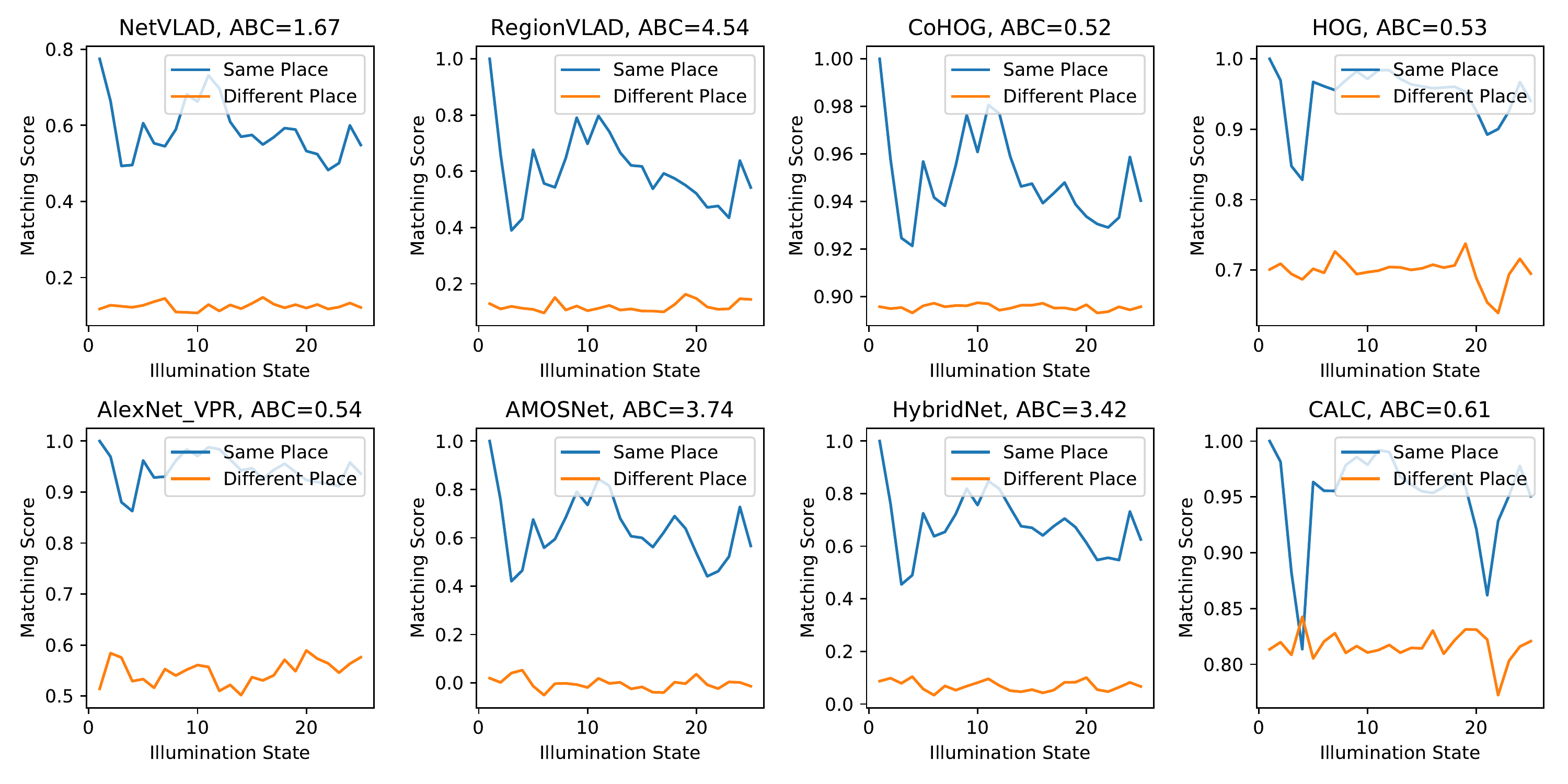}
\end{center}
\caption{The change in matching score for the illumination variations is shown here on the MIT Multi-illumination dataset. The confidence score variation is given for all techniques on the $25$ illumination positions, as explained in sub-section \ref{Evaluation_Mechanism}.}
\label{matchingscore_variation_multiilluminationdataset}
\vspace{-0mm}
\end{figure*}

We have extended our illumination-invariance analysis from the Point Features dataset to the MIT Multi-illumination dataset and the curves on Multi-illumination dataset are presented in Fig. \ref{matchingscore_variation_multiilluminationdataset}. There is a very sharp drop in place matching confidence for illumination cases 3 and 4 for all the VPR techniques, which re-affirms our finding on the Point Features dataset regarding the significantly large effect of directional illumination change (see Fig. \ref{MIT_multillumiation_dataset_exemplar}) on the place matching performance. The effect of illumination change on a handcrafted technique such as HOG is lower than that on a learning-based technique like CALC on the MIT Multi-illumination dataset, similar to prior observations on the Point Features dataset, however this does not generalise to other learning-based techniques. The reported performance decline by varying illumination cases can be potentially combined with illumination-source prediction works (\cite{gardner2017learning}, \cite{hold2017deep}) to predict when a VPR technique might fail and how different VPR techniques could complement each other in these scenarios.

\subsection{Variance vs Invariance} \label{Variance vs Invariance}
A generic perception among the VPR research community, as evident from the recent trend in developing highly viewpoint-invariant VPR techniques is that the more viewpoint-invariant a technique is, the more utility it has to offer. Through this sub-section, we take the opportunity to address that this may not always be the case. In fact, viewpoint-variance may actually be required in some applications, instead of viewpoint-invariance. A key example here are the applications where VPR techniques act as the primary localisation module and where, there is no image-to-image, epipolar-geometry-based motion estimation (location refinement) module. For example, \cite{zeng2019lookup} extend the concept of VPR for precise localisation in mining environments. Similar extensions of VPR as the only module for precise-localisation are possible in several applications, where an accurate geo-tagged image database of the environment exists, e.g, in factory/plant environments or outdoor applications which can afford to create an \textit{a priori} accurate appearance-based metric/topometric map of the environment. For such applications, VPR techniques are required to have viewpoint-variance, so that even if the $2$ images of the same place are viewpoint-varied, the VPR technique can distinguish between them to perform metrically-precise localisation. If a viewpoint-invariant technique is utilised in this scenario, the inherent viewpoint-invariance will lead to discrepancies in localisation estimates and eventually cause a system failure.          

Thus, a key area to investigate within VPR research should be controlled viewpoint-variance. In sub-section \ref{Invariance Analysis}, we presented a methodology to estimate the viewpoint-invariance of a technique, however, there is no control parameter for any technique that could govern and tune its invariance to viewpoint changes. We believe that this is an exciting research challenge and should be a topic for VPR research in the upcoming years. Nevertheless, our proposal is that both viewpoint-variance and invariance are desirable properties, depending upon the underlying application and should be regarded/investigated accordingly.

\section{Conclusions and Future Work}
\label{conclusionsandfuturework}
In this paper, we presented a comprehensive and variation-quantified evaluation framework for visual place recognition performance. Our open-source framework VPR-Bench integrates $12$ different indoor and outdoor datasets, along with $10$ contemporary VPR techniques and popular evaluation metrics from both the computer vision and robotics communities to assess the performance of techniques on various fronts. The framework design is modular and permits future integration of datasets, techniques and metrics in a convenient manner. We utilised the variation- and illumination-quantified Point Features dataset to evaluate and analyse the level and nature of variations that a VPR technique can handle. We then extended this analysis and our findings from the synthetic Point Features dataset to the QUT Multi-lane dataset and the MIT multi-illumination dataset.

Using our framework, we provide a number of useful insights about the nature of challenges that a particular technique can handle. We identify that no universal state-of-the-art technique exists for place matching and discuss the reasons behind the success/failure of these techniques from one dataset to another. In our evaluations, DenseVLAD, a learning-based but non-deep-learning technique has achieved state-of-the-art AUC-PR on 6 out of the 12 datasets, which indicates the potential for further developing the traditional specialised techniques and pipelines for VPR. We also report that 8 out of the 10 techniques have achieved state-of-the-art AUC-PR on at least one dataset and therefore ensemble-based approaches can present value towards creating a generic VPR system. Our results reveal that the utility of VPR techniques highly depends on the employed evaluation metric, and that the corresponding utility is application-dependent, e.g. the state-of-the-art for RecallRate is different from that of AUC-PR because the former assumes the availability of a false-positive rejection scheme. Our results demonstrate the utility of ROC curves for finding new places which is usually not discussed in existing VPR literature. The encoding times for deep-learning-based techniques are significantly higher than handcrafted feature descriptors, but the availability of a GPU-based platform reduces this gap for most techniques. There are exceptions to this, e.g. RegionVLAD, a deep-learning-based technique which cannot benefit much from a GPU in terms of encoding time due to its CPU-bound intense region-extraction scheme. We demonstrate that the descriptor matching time is dependent upon four factors: distance/similarity function, number of descriptor dimensions, length of each dimension, and the descriptor data-types. This identifies the need for further investigating the trade-offs between reduced matching time at reduced descriptor precision and size. Overall, our work found that there is no one-for-all evaluation metric for VPR research, and that only a combination of these metrics presents the overall utility of a technique.

Our new analysis for viewpoint and illumination-invariance quantification is developed around the Point Features dataset, and integrated within the framework for ease-of-use by other VPR researchers. Our results on this dataset identify that 3D viewpoint change has more adverse effect on matching confidence than lateral viewpoint change, but deep-learning-based techniques generally suffer less from 3D change than handcrafted feature descriptors. We further show that directional illumination change presents a bigger challenge for VPR than uniform illumination change, both for deep-learning and handcrafted techniques. We also propose that viewpoint variance instead of viewpoint invariance can also be important for VPR systems, e.g. for accurate localisation, sensitivity to viewpoint change can be a feature. Because we have employed a number of different datasets, techniques and metrics, VPR-Bench enables many more performance comparisons, and we have only discussed a few selected comparisons to limit the scope.

It remains future work to further investigate the relation between place matching performance and the bottle-necks caused by encoding times and linear scaling of matching times. The role of various parameters that determine the descriptor matching time is briefly introduced in this work, but also deserves more detailed future investigation. It would also be useful to include evaluations on more challenging environments, such as under-water or aerial, on more extreme weather conditions, on motion-blur and on opposing viewpoints. Further insights could be obtained by evaluating how different metrics yield different state-of-the-art VPR techniques on the same dataset. We hope that this work proves useful for both the computer vision and the robotics VPR communities to compare newly proposed techniques in detail to the state-of-the-art on these varied datasets using diverse evaluation metrics. We are keen on integrating more VPR techniques into VPR-Bench and encourage any feedback, collaborations and suggestions.

\begin{acknowledgements}
Our work was supported by the UK Engineering and Physical Sciences Research Council through grants EP/R02572X/1, EP/P017487/1 and EP/R026173/1 and in part by the RICE project funded by the National Centre for Nuclear Robotics Flexible Partnership Fund. This research has also been supported by the TU Delft AI Labs programme. Michael Milford was supported by ARC grants FT140101229, CE140100016 and the QUT Centre for Robotics.
\end{acknowledgements}

\appendix
\appendixpage

\section{VPR-Bench Design}
\label{Design_Structure}

\subsection{Code Structure}
\begin{figure}[t]
\begin{center}
%\fbox{\rule{0pt}{2in} \rule{0.9\linewidth}{0pt}
\includegraphics[width=1\linewidth]{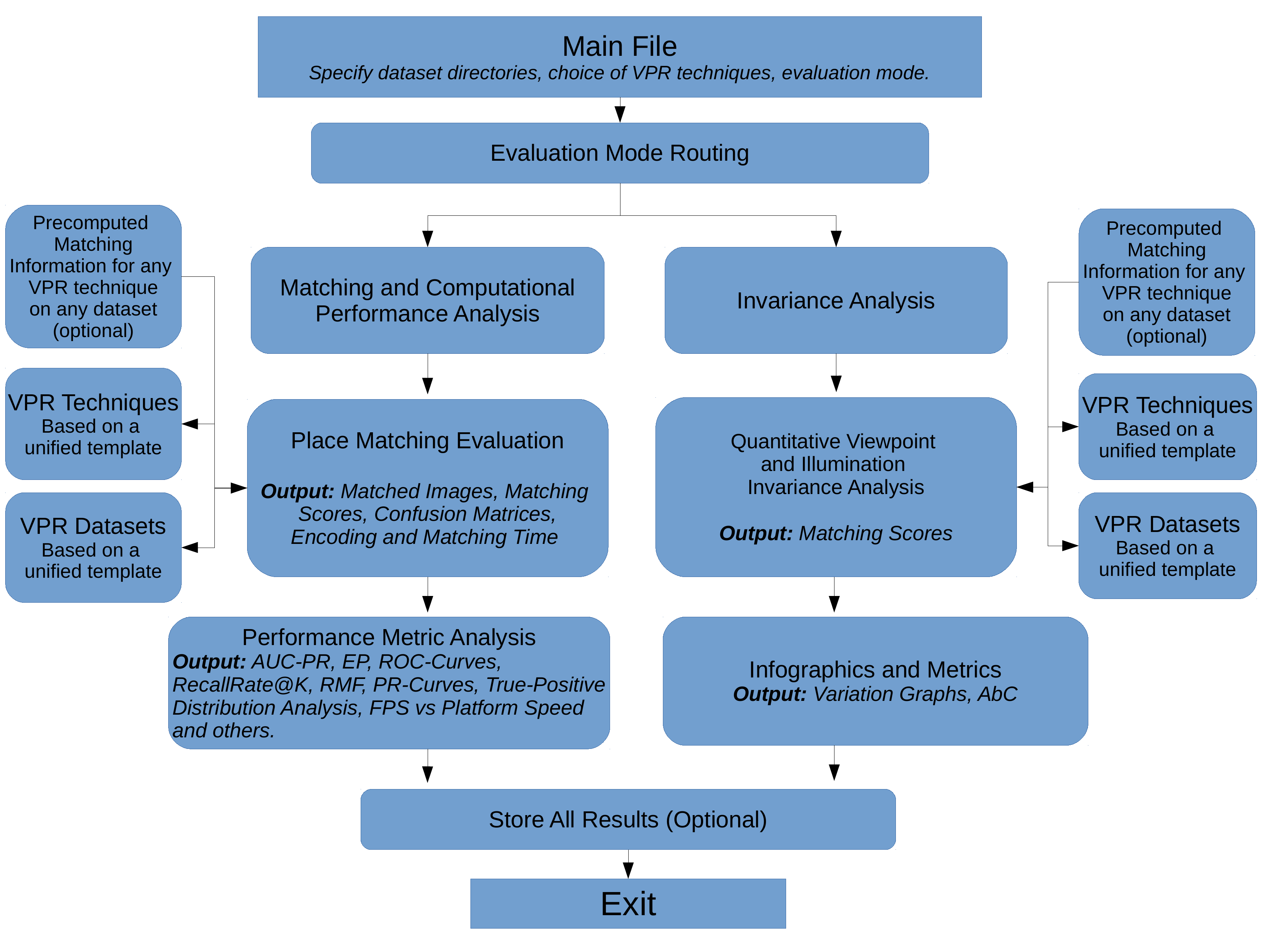}
\end{center}
\caption{The code structure of the VPR-Bench framework is shown here.}
\label{VPR-Bench_Codeflow}
\vspace{-5mm}
\end{figure}

The entire framework has been designed with $2$ key focuses: a) A holistic, fully-integrated and easy-to-use framework for VPR performance evaluation at all fronts, b) Modularity and convenient templates for regular updates and future consistency. In this respect, while the modularity, template design and available content within the modules, are explained individually for each of the modules in their respective dedicated sub-sections; this sub-section presents the overall framework structure and implementation details. The code structure of our framework has been described in Fig. \ref{VPR-Bench_Codeflow}. 

The entry to the framework is a convenient main file, where the choice of evaluation datasets, VPR techniques and evaluation mode can be specified. At present there are $2$ evaluation modes: 1) VPR Performance Evaluation and 2) Invariance Analysis. The former yields the place matching performance of different VPR techniques (implemented within the framework and/or integrated using pre-computed matching information) on a specified dataset using different metrics related to precision and computation. The latter tries to present the invariance of these techniques to quantified viewpoint- and illumination-variations. There are 12 evaluation datasets available in the framework from both indoor and outdoor environments. We have re-implemented $8$ different VPR techniques by modifying the open-source codes as per our templates or self-implementing in cases where open-source codes were not available. 
The VPR-Bench framework is written fully in Python (2.7) (working on upgrading to Python 3), which has been the most used programming/scripting language for VPR research. Our framework does not have a dedicated Graphical-User-Interface (GUI), because the framework is targeted for developers/researchers who are assumed to have basic knowledge of the domain. GUIs also make future improvements much complex and limit the flexibility of an application. The open-source code has been tested on a Ubuntu 20.04.1 LTS system. By default, the framework does not need a GPU (Graphical Processing Unit) for any of the evaluations. Therefore, a huge percentage of VPR researchers, academics and developers, from a broad range of application domains can conveniently use our framework.

\subsection{Integrating New Datasets and Techniques}
\label{Integrating_Datasets_and_Techniques}
As we are focused on providing flexibility and ease for integrating new VPR techniques and datasets into VPR-Bench (additional to the already available 12 datasets and 8 techniques), we have briefly summarised the required steps for both of these changes below:

\begin{enumerate}
    \item For integrating a new dataset into VPR-Bench, no change in the framework code is required. You need to setup the dataset as per our unified template, which has been explained in appendix \ref{datasets_template}. and then set the directory path for this dataset in the main file.
    
    \item There are two possible ways to integrate a new VPR technique into VPR-Bench: (a) Re-implement the technique as per our template within the VPR-Bench framework, (b) Use pre-computed data through an external implementation of the technique. We encourage the former, where the main file for this respective technique needs to implement $3$ functions, as per the template described in appendix \ref{techniques_template}. Once these functions have been implemented, they only need to be imported in our framework and all other modules will be implicitly integrated for this technique. The benefit of re-implementing a technique as per our template is the ease for new researchers to understand, utilize and modify the implementation of these various VPR techniques based on a fixed and compact template. Moreover, templates also make computational analysis more fair, by affixing the input and output pipelines (i.e. the time taken to input and output data to a VPR techniques' various functions). For the latter, we maintain a provision in our framework to re-use pre-computed data through an external implementation and integrate it with the features offered by our framework. The computational analysis for techniques integrated via external implementations (non-template) is still relevant (albeit will vary based upon the implementation) as long as the underlying hardware is the same. A unified template has been developed for integrating pre-computed data, that takes in the matching scores for all the query images with all the reference images, feature encoding time and descriptor matching time. We have integrated DenseVLAD and GeM using this pre-computed data in our work. The details for integrating VPR techniques in this fashion will also be provided in the files supporting the release of our open-source code.
    
\end{enumerate}

\section{VPR-Bench Datasets Template}
\label{datasets_template}
In order to have a fixed template for all the datasets that are available in (or can be integrated into) VPR-Bench, we design a simplistic, generic template that can accommodate the variations within the dataset formats. Firstly, the query and reference traverses for a dataset are represented by their dedicated sub-folders. Images within each of these folders need to be named as integers, which is motivated by a graph structure, such that for a traversal-based dataset, increments or decrements to integer values can represent the temporally and/or geographically next or previous image, respectively. The ground-truth file for each dataset is a numpy array (.npy). This multi-dimensional numpy array of ground-truth information has dimensions of $T_Q \times 2$, where $T_Q$ is the total number of query images in the dataset. For all $T_Q$ rows of query images, the first column represents the query image index and the second column contains the list of indices of all ground-truth matching reference images. We have used the simplistic image indices/names as our choice of ground-truth, because they can be parsed from a range of different modalities, like GPS information, pose-information and/or manual frame correspondences, as shown in this work by restructuring all the 12 datasets to the described common template. 

\section{VPR-Bench Techniques Template}
\label{techniques_template}
Each VPR technique has a different approach to the problem, which may include neural-network models or traditional feature descriptors. There may be added functionality, like ROI-extraction, image pre-processing, descriptor adaptation, usage of sequential and/or geometric prior etc. The designed templates for techniques have the provision to allow for such pre- and post-processing steps. We also provide a parallel path in our pipeline to seamlessly integrate pre-computed place matching information from a different technique running on the same/different platform.

Let $Q$ be a query image and $M_R$ be a list/map of $R$ reference images. The feature descriptor(s) of a query image $Q$ and reference map $M_R$ can be denoted as $F_Q$ and $F_M$, respectively. If a technique uses ROI-extraction, $F_Q$ will hold within it all the required information in this regards, including location of regions, their descriptors and corresponding salience as a multi-dimensional list. The input $Q$ can also be a sequence of Query images and any other pre/post-processed form of a query candidate. For a query image $Q$, given a reference map $M_R$, let us denote the best matched image/place by a VPR technique as $P$ (where, $P \in M_R$) with a matching score $S$. The matching score $S$ can be defined as $S \in [0,1]$. The confusion matrix (matching scores with all reference images) can be denoted as $C$. Based on these notations, the following $3$ functions need to be implemented in the main file of a VPR technique. The definitions (names) of these functions remain the same for all VPR techniques and our framework performs technique-aware selective re-imports of these functions to maintain consistency and ease-of-integration.

\begin{algorithm} [H]
\small
\renewcommand\thealgorithm{}
\caption{VPR Technique Required Template}
\begin{algorithmic}

\STATE $\textbf{def} \; compute\_query\_desc \; (Q)$
\STATE $\; \; \; Preprocessing \; Steps$
\STATE $\; \; \; Function \; Body$
\STATE $\; \; \; Postprocessing \; Steps$
\STATE $\; \; \; return \; F_Q$ \\

\STATE $\textbf{def} \; compute\_map\_features \; (M_R)$
\STATE $\; \; \; Preprocessing \; Steps$
\STATE $\; \; \; Function \; Body$
\STATE $\; \; \; Postprocessing \; Steps$
\STATE $\; \; \; return \; F_M$ \\

\STATE $\textbf{def} \; perform \_VPR \; (F_Q, F_M)$
\STATE $\; \; \; Preprocessing \; Steps$
\STATE $\; \; \; Function \; Body$
\STATE $\; \; \; Postprocessing \; Steps$
\STATE $\; \; \; return \; P, \; S, \; C$

\end{algorithmic}
\addtocounter{algorithm}{-1}
\end{algorithm}

\bibliographystyle{spbasic}      % basic style, author-year citations
\bibliography{main.bbl}   % name your BibTeX data base

\end{document}